\documentclass[final,1p,times]{elsarticle}

\usepackage[T5]{fontenc}

\usepackage{times}
\usepackage{latexsym}
\usepackage{graphicx}
\usepackage{multirow}
\usepackage{pgf-pie}   
\usepackage{soul}
\usepackage{color}
\usepackage{tablefootnote}
\usepackage{adjustbox}
\usepackage{arydshln}
\usepackage{url}
\usepackage{float}
\usepackage{graphicx,graphics}
\usepackage{caption}
\usepackage{subcaption}
\usepackage[bookmarks,bookmarksopen,bookmarksdepth=3, bookmarksnumbered]{hyperref}
\usepackage{longtable}

\usepackage{tablefootnote}
\usepackage{amsmath}

\usepackage{xcolor, color, soul}

\usepackage{CJKutf8}

\usepackage{adjustbox}
\usepackage{caption}

\usepackage{graphicx}
\usepackage{hyperref}
\usepackage{cleveref}
\usepackage{multirow}
\usepackage{multicol}
\usepackage{longtable}
\usepackage{pgfplots}

\usepackage{amsmath}

\usepackage{tgpagella}
\usepackage{threeparttable}

\usepackage{adjustbox}

\usepackage{amsfonts}

\usepackage{lipsum}

\usepackage{xcolor}

\usepackage{soul}

\definecolor{lavender}{RGB}{240,230,250}
\definecolor{lightblue}{RGB}{173, 216, 230}

\sethlcolor{yellow}

\begin{document}
% \let\WriteBookmarks\relax
% \def\floatpagepagefraction{1}
% \def\textpagefraction{.001}

\title{A New Benchmark Dataset and Mixture-of-Experts Language Models for Adversarial Natural Language Inference in Vietnamese}                      

% First author/
\author[1,2]{Tin Van Huynh}
\ead{tihv@uit.edu.vn}

\author[1,2]{Kiet Van Nguyen\corref{cor1}}%[orcid=0000-0002-8456-2742]
\ead{kietnv@uit.edu.vn}
% \cormark[1]

\author[1,2]{Ngan Luu-Thuy Nguyen}%[orcid=0000-0003-3931-849X]
\ead{ngannlt@uit.edu.vn}

\affiliation[1]{organization={Faculty of Information Science and Engineering, University of Information Technology},
    city={\\Ho Chi Minh City},
    %postcode={70000}, 
    country={Vietnam}}
    
\affiliation[2]{organization={Vietnam National University},
    city={Ho Chi Minh City},
    %postcode={70000}, 
    country={Vietnam}}

\cortext[cor1]{Corresponding author}

\journal{}

\begin{abstract}

Existing Vietnamese Natural Language Inference (NLI) datasets lack adversarial complexity, limiting their ability to evaluate model robustness against challenging linguistic phenomena. In this article, we address the gap in robust Vietnamese NLI resources by introducing ViANLI, the first adversarial NLI dataset for Vietnamese, and propose NLIMoE, a Mixture-of-Experts model to tackle its complexity. We construct ViANLI using an adversarial human-and-machine-in-the-loop approach with rigorous verification. NLIMoE integrates expert subnetworks with a learned dynamic routing mechanism on top of a shared transformer encoder. ViANLI comprises over 10,000 premise – hypothesis pairs and challenges state-of-the-art models, with XLM-R $_{Large}$ achieving only 45.5\% accuracy, while NLIMoE reaches 47.3\%. Training with ViANLI improves performance on other benchmark Vietnamese NLI datasets including ViNLI, VLSP2021-NLI, and VnNewsNLI. ViANLI is released\footnote{Our dataset is available at: \url{https://huggingface.co/datasets/uitnlp/ViANLI}} for enhancing research into model robustness and enriching resources for future Vietnamese and multilingual NLI research.
\end{abstract}

%    \item Created {\bf multi-domain} dataset with {\bf 500k+ target-aspect pairs} from social comments.
%    \item Proposed {\bf transformer-based} approach achieving SOTA on Vietnamese TASA task.
%    \item Evaluated Vietnamese {\bf multi-task sentiment analysis} including TBSA and ABSA tasks.
%    \item Investigated {\bf challenges and characteristics} in Vietnamese sentiment analysis tasks.

% Keywords

\maketitle

\section{Introduction}
\label{sect:introduction}

Natural language inference is a core task in natural language processing (NLP) that involves determining whether a hypothesis can be logically inferred from a given premise \citep{bowman2015large}. As a cornerstone of natural language understanding, NLI underpins various downstream applications such as question answering \citep{nguyen2020vietnamese, van2022new}, machine translation \citep{kann2022americasnli}, relation extraction \cite{li2025exploring}, and text summarization \citep{mishra2021looking}. With the advancement of deep learning, transformer-based models have achieved impressive results across many NLI benchmarks. However, studies have shown that state-of-the-art (SOTA) models remain highly vulnerable to adversarial examples—inputs carefully crafted to induce incorrect predictions despite being semantically coherent \citep{jia2017adversarial, nie-etal-2020-adversarial}. These adversarial failures raise critical concerns about the robustness and generalizability of NLI models, particularly when deployed in real-world or high-stakes domains \citep{agarwal2022graphnli, tandon2021use}. As a result, constructing adversarial NLI datasets has emerged as an essential direction for evaluating and improving the reliability of NLI systems. Unlike traditional datasets, which often consist of semantically transparent sentence pairs, adversarial datasets intentionally contain challenging linguistic phenomena designed to expose model weaknesses.

Despite progress in adversarial dataset construction for high-resource languages such as English \citep{nie-etal-2020-adversarial} and Chinese \citep{xu-markert-2022-chinese}, the Vietnamese language remains underrepresented. Existing Vietnamese NLI datasets achieving high performances such as ViNLI \citep{huynh-etal-2022-vinli}, VnNewsNLI \citep{nguyen2022building}, and VLSP2021-NLI \citep{JCSCE} were developed using conventional annotation pipelines and lack the adversarial design necessary for assessing model robustness. Although ViNLI has played an important role in evaluating Vietnamese NLI models, it still suffers from several weaknesses, including annotator artifacts and limited coverage of reasoning types. To address this gap, we present ViANLI, the first adversarial NLI dataset for the Vietnamese language. Inspired by the human-and-machine-in-the-loop methodology \citep{BONETJOVER2023107152,nie-etal-2020-adversarial, cao2025humanintheloopgenerationadversarialtexts}, our dataset construction process involves iterative collaboration between annotators and machine learning models. This method is also used to create other adversarial datasets such as ANLI \citep{nie-etal-2020-adversarial}, AdversarialQA \citep{bartolo2020beat}, and Adversarial VQA \cite{Li_2021_ICCV}. This design enables the generation of premise–hypothesis pairs that are difficult for models to classify, thereby enhancing both dataset complexity and linguistic coverage. ViANLI consists of over 10,000 high-quality, adversarially constructed sentence pairs and serves as a new benchmark for Vietnamese NLI evaluation.

While constructing such a dataset offers a robust evaluation benchmark, it also necessitates novel modeling strategies capable of dealing with adversarial inputs. To this end, in addition to releasing ViANLI, we propose NLIMoE, a Mixture-of-Experts language model tailored for adversarial NLI. NLIMoE integrates expert subnetworks with a learned dynamic routing mechanism built on top of a shared transformer encoder (e.g., XLM-RoBERTa \citep{conneau-etal-2020-unsupervised}), enabling the model to dynamically assign inference examples to specialized experts. Unlike a static threshold, our routing process employs a dynamic threshold that flexibly adjusts expert selection based on input complexity. This design enhances the ability of the model to handle syntactic variation, semantic ambiguity, and linguistic noise as key characteristics of adversarial data.

In this article, we have four main contributions described as follows.
\begin{itemize}
    \item \textbf{Introduction of ViANLI Dataset}: We introduce ViANLI, the first adversarial NLI dataset for Vietnamese, containing over 10,000 premise-hypothesis pairs. Created through a human-and-machine-in-the-loop approach with rigorous human-machine verification, ViANLI is designed to challenge state-of-the-art models, thereby enhancing research into model robustness and enriching resources for future Vietnamese and multilingual NLI research.

    \item \textbf{Proposal of NLIMoE Model}: We propose NLIMoE, a Mixture-of-Experts (MoE) language model tailored for high-complexity NLI tasks in adversarial settings. Integrating expert subnetworks and dynamic routing on a shared transformer encoder, NLIMoE achieves a 47.3\% accuracy on ViANLI, surpassing the best baseline model (XLM-R $_{Large}$) at 45.5\%, setting a new benchmark for NLI performance.
    
    \item \textbf{Performance Improvement on Other Benchmark NLI Datasets}: Our experimental results reveal that training models such as mBERT, PhoBERT, XLM-R, and NLIMoE with ViANLI-augmented data improves their performance on other benchmark Vietnamese NLI datasets. This validates the dual role of ViANLI as a benchmarking tool and a valuable training resource.
    
    \item \textbf{Enhanced Inference Capabilities of NLIMoE}: Our proposed model, NLIMoE, demonstrates stronger inference performance, particularly when handling long sentences and resolving entailment ambiguities. It outperforms other models in challenging scenarios such as numerical \& quantitative reasoning, reference \& naming, logical inference, lexical reasoning, and external knowledge reasoning.
    
\end{itemize}

The structure of this article is as follows: Section \ref{sect:introduction} provides an overview of the NLI research context, the trends in constructing adversarial datasets, and the rationale for developing an adversarial NLI dataset for Vietnamese. Section \ref{sect:related_works} reviews related datasets and models in the field of NLI. Section \ref{sect:dataset} outlines the process of constructing our ViANLI dataset. Section \ref{sect:dataset_analysis} presents both basic and detailed statistical analyses of the dataset. Section \ref{sect:proposed_model} describes the architecture of the Mixture-of-Experts NLI model (NLIMoE) that we designed. Section \ref{sect:experiment} discusses the experimental setup and presents the results. Section \ref{sect:result_analysis} provides an analysis and discussion of the experimental findings. Finally, Section \ref{sect:conclusion} summarizes the key conclusions of our research and suggests directions for future work.

\section{Related Works}
\label{sect:related_works}
In recent years, extensive research has been conducted to produce adversarial resources for various languages to aid in natural language inference. In this section, we discuss numerous available NLI datasets (non-adversarial and adversarial datasets) and related models for natural language inference. Additionally, we provide details of resources available for other languages. Lastly, we elaborate on the research conducted on natural language inference using various machine learning and deep learning models.

\subsection{Related Datasets}

Based on the data collection strategy of the NLI datasets, we divide them into adversarial and non-adversarial datasets. Some well-known datasets related to NLI in other languages are summarized and compared by us to the Vietnamese NLI datasets in Table \ref{tab:summary_dataset}.

\textbf{Non-adversarial datasets}: Through ongoing research efforts, researchers have released many datasets related to the field of Natural Language Inference (NLI). In particular, these datasets are generated using a variety of methodologies and languages. This has brought many positive effects on Natural Language Understanding. In the early stages of the development of NLI research, The Recognizing Textual Entailment (RTE) Challenge, introduced by \cite{dagan2005pascal}, was one of the renewed efforts aimed at formalizing the evaluation of Natural Language Inference (NLI) tasks. The challenge created a common platform and a standardized dataset that allowed researchers to assess the ability of computational models to recognize textual entailment. This research area exploded in 2015, when \cite{bowman2015large} published the SNLI dataset. This is a large-scale resource introduced by researchers at Stanford that contains nearly 600,000 sentence pairs. SNLI employs a unique data collection method that combines crowd-sourcing with careful curation. Quality control measures were in place to ensure the reliability and consistency of the annotations. This methodical approach to data collection enables the creation of a large-scale, high-quality dataset that has been crucial for benchmarking. Therefore, SNLI is quickly considered a cornerstone in the NLI field, providing a benchmark for evaluating various machine learning models' generalization and reasoning capabilities. The MultiNLI (Multi-Genre Natural Language Inference) \citep{multinli} dataset was developed to address some of the limitations of the Stanford Natural Language Inference (SNLI) dataset, including the problem of the data being relatively simple and easy for advanced machine learning models. The authors have revised the process of building MultiNLI in several ways to solve these issues, including using diverse sources, complex reasoning, multiple domains, and better annotation guidelines. MultiNLI aims to provide a more rigorous and challenging benchmark for NLI models by making these improvements. This data construction method was quickly applied to build a series of other datasets such as OCNLI \citep{hu2020ocnli}, IndoNLI \citep{mahendra-etal-2021-indonli}, MedNLI \citep{romanov-shivade-2018-lessons}, ViNLI \citep{huynh-etal-2022-vinli},VnNewsNLI \citep{nguyen2022building}, and VLSP2021-NLI \citep{JCSCE}. Also, there are some other approaches to building NLI data sets, such as using question and answer data to convert to NLI data sets such as SciTail \citep{khot2018scitail}, FarsTail \citep{amirkhani2023farstail}, using a translator to translate NLI data from one language to another such as XNLI \citep{conneau2018xnli}, Hinglish \citep{khanuja2020new}, KorNLI \citep{ham2020kornli}. Using the NLI framework to label claims based on their logical relationships with hypotheses, researchers can create a fact-checking dataset that enables the development and evaluation of fact-checking models and systems. There are famous fact-checking datasets created in such a way that have been published, such as FEVER \citep{thorne2018fever}, VITAMINC \citep{schuster2021get}, and FEVEROUS \citep{aly2021feverous}.

\textbf{Adversarial datasets}: The growing trend of building adversarial datasets in Natural Language Processing (NLP) has been propelled by the increasing realization that machine learning models, though powerful, are often susceptible to nuanced forms of data manipulation. Significant efforts are underway in the field of Natural Language Inference (NLI) to construct more challenging and representative datasets. One approach to building adversarial datasets involves generating examples specifically designed to challenge or fool existing NLI models. Adversarial examples are meant to expose the limitations and vulnerabilities of existing NLI models, thereby aiding in their subsequent fortification. According to \cite{bartolo2020beat}, there are different approaches to building adversarial datasets: i) adversarial filtering, where the adversarial model is typically implemented in an isolated phase, which comes after the data creation stage. Techniques include manipulating input text by flipping, inserting, deleting, or swapping words or characters. Perturbation data is generated to be used during adversarial training. Many studies have combined this type of data with machine learning models and have achieved positive results in making the model more robust; works include AdvEntuRe \citep{kang2018adventure}, SCPNs \citep{iyyer2018adversarial}, First-Order Logic constraints \citep{minervini2018adversarially}, ASCC \citep{dong2020towards}, FreeLB \citep{zhu2019freelb}. In addition, on the reading comprehension task, there are also datasets built according to this method such as SWAG \citep{zellers2018swag}, HotpotQA \citep{yang2018hotpotqa}; ii) adversarial model-in-the-loop annotation, where both annotators and machine learning models work iteratively in a cycle to create and refine adversarial examples. This approach usually starts with a model initially generating adversarial samples, followed by annotators reviewing, modifying, or creating new ones based on these initial examples. This method was successfully applied to build the famous ANLI \citep{nie-etal-2020-adversarial} dataset, as it exploited the vulnerabilities of the model very well. In the question-answering task, there are many datasets with adversarial properties, such as DROP \citep{dua2019drop}, CODAH \citep{chen2019codah}, Quoref \citep{dasigi2019quoref}, AdversarialQA \citep{bartolo2020beat}, Quizbowl \citep{wallace-etal-2019-trick}. The adversarial human-and-machine-in-the-loop procedure is also quickly used to create fact-checking datasets, as seen in FEVER 2.0 \citep{thorne2019fever2}, which addresses weaknesses of the original FEVER dataset.

\begin{table*}[]
\centering
\caption{Summary of Well-known Datasets Related to Natural Language Inference. NLI and FC Stand for Natural Language Inference and Fact-checking, Respectively.}
\label{tab:summary_dataset}
\resizebox{1\textwidth}{!}{\begin{tabular}{lrrrrrr}
\hline
\bf Dataset  & \bf NLP task  & \bf Language  & \bf Text genre & \bf Year  & \bf Quantity & \begin{tabular}[c]{@{}r@{}} \bf Adversarial\\ \bf strategy\end{tabular} \\ \hline
FEVER \citep{thorne2018fever}     & NLI, FC  & English   & Wikipedia & 2018 & $\sim$ 185K & No   \\
FEVER2.0 \citep{thorne2019fever2}     & NLI, FC  & English   & Wikipedia & 2019 & 1,174 & Yes    \\
VITAMINC \citep{schuster2021get}      & NLI, FC  & English   & Wikipedia & 2021 & $\sim$ 488K & No    \\
FEVEROUS \citep{aly2021feverous}      & NLI, FC  & English   & Wikipedia & 2021 & $\sim$ 87K & No    \\
SNLI \citep{bowman2015large}          & NLI  & English   & Image captions & 2015 & $\sim$ 570K & No    \\ 
MultiNLI \citep{multinli}     & NLI  & English   & Multi-genre & 2018 & $\sim$ 433K & No    \\ 
OCNLI \citep{hu2020ocnli} & NLI  & Chinese   & Multi-genre & 2020 & $\sim$ 56K  & No    \\ 
IndoNLI \citep{mahendra-etal-2021-indonli}       & NLI  & Indonesian   & Multi-genre & 2021 & $\sim$ 18K  & No   \\ 
MedNLI \citep{romanov-shivade-2018-lessons}        & NLI   & English    & Medical & 2019  & $\sim$ 14K   & No    \\ 
SciTail \citep{khot2018scitail}   & NLI   & English    & Education & 2018  & $\sim$ 27K  & No   \\ 
FarsTail \citep{amirkhani2023farstail}   & NLI  & Persian    & Education & 2021  & $\sim$ 10K  & No   \\ 
XNLI \citep{conneau2018xnli}  & NLI   & Multi-language    & Multi-genre & 2018  & $\sim$ 129K  & No   \\ 
Hinglish \citep{khanuja2020new}   & NLI   & Multi-language    & Movie scripts & 2020  &  2,240  & No   \\
KorNLI \citep{ham2020kornli}   & NLI   & Korean     & Multi-genre & 2020  & $\sim$ 950K  & No   \\
ANLI \citep{nie-etal-2020-adversarial}   & NLI   & English    & Multi-genre & 2020  & $\sim$ 169K  & Yes   \\
VLSP2021-NLI \citep{JCSCE}    & NLI   & Multi-language    & Newswire & 2022  & $\sim$ 20K  & No   \\
VnNewsNLI \citep{nguyen2022building}  & NLI   & Vietnamese    & Newswire & 2022  & $\sim$ 32K   & No   \\
ViNLI \citep{huynh-etal-2022-vinli}  & NLI   & Vietnamese    & Newswire & 2022  & $\sim$ 30K   & No   \\ \hline
\textbf{ViANLI (Ours)}  & \bf NLI   & \bf Vietnamese    & \bf Newswire & \bf 2025  & $\sim$ \bf 10K  & \bf Yes   \\
\hline 
\end{tabular}}

\end{table*}

Although recent efforts have introduced several Vietnamese NLI datasets such as ViNLI \citep{huynh-etal-2022-vinli}, VnNewsNLI \citep{nguyen2022building}, and VLSP2021-NLI \citep{JCSCE}, these resources were constructed without adversarial considerations and therefore provide limited challenge for modern models. Prior studies already report strong baselines like XLM-R $_{Large}$ achieving high performance (85.99\% accuracy on ViNLI, 94.79\% accuracy on VnNewsNLI, and 88.69\% F1 on VLSP2021-NLI), which indicates that these datasets lack adversarial complexity to evaluate model robustness. Although ViNLI has played an important role in evaluating Vietnamese natural language inference models, it still suffers from several weaknesses, such as annotator artifacts and limited reasoning diversity. To address these issues, we introduce ViANLI, a more challenging adversarial dataset designed to better evaluate the reasoning ability of NLP models. This motivates the construction of ViANLI as the first adversarial NLI dataset for Vietnamese. Our approach adopts an adversarial human-and-machine-in-the-loop paradigm, where humans and models iteratively collaborate to generate and refine examples. This design not only increases dataset difficulty but also allows annotators to expose model vulnerabilities during the creation process, yielding a more reliable benchmark for robust reasoning.

The research gaps identified in existing Vietnamese NLI datasets are as follows. First, there is a significant absence of adversarial NLI datasets specifically for the Vietnamese language, which hinders the evaluation of model robustness against challenging linguistic phenomena in this low-resource context. Second, current Vietnamese NLI datasets, such as ViNLI, VLSP2021-NLI, and VnNewsNLI, lack the iterative human-model collaboration needed to incorporate diverse adversarial elements like syntactic variations and semantic ambiguities, resulting in over-optimistic performance metrics for state-of-the-art models. Third, multilingual adversarial datasets overlook language-specific cultural and contextual inferences relevant to Vietnamese, limiting their applicability and transferability. These gaps directly align with the objectives of this article, which include introducing ViANLI as the first adversarial Vietnamese NLI dataset to challenge models and proposing NLIMoE as a specialized Mixture-of-Experts model to handle such complexities, thereby improving robustness and performance on both adversarial and benchmark NLI datasets.

\subsection{Related NLI Models}

Research topics in NLP, including NLI, have witnessed the daily development of machine learning models from simple models to more complex ones. The diverse evolution of current machine learning models has enriched the choices of models for experimentation, suitable for the characteristics of different tasks and data domains. Moreover, as more adversarial NLI datasets emerge, the performance of these models also needs to increase significantly.

In its early stages, probability-based models have historically laid the groundwork for approaches to natural language inference and have established a foundation long before the emergence of current deep learning techniques. These models operated on the principle of computing the likelihoods of hypotheses given premises, often rooted in Bayesian inference \citep{dagan2005pascal, finkel2006solving}. Latent Semantic Analysis (LSA) \citep{landauer1997solution} uses Singular Value Decomposition (SVD) to compress space and discover semantic relationships between words and documents. LSA can be used to measure the semantic similarity between a premise and a hypothesis \citep{oh-etal-2017-deep}. CBOW (Continuous Bag of Words) and Skip-gram are two architectures introduced in the Word2Vec model \citep{mikolov2013distributed} to learn vector word representations. Both of these techniques aim to learn word representations that can capture semantics based on the context in which the word appears, and these methods have also yielded promising results on the NLI task \citep{bowman2015large, multinli, hu2020ocnli, mahendra-etal-2021-indonli, huynh-etal-2022-vinli}. Although these models offered valuable information, they often struggled to capture intricate semantic relationships and nuances in language.

The challenges and shortcomings of these probabilistic approaches paved the way for neural models, especially with the advent of deep learning. Neural models, such as Recurrent Neural Networks (RNNs) \citep{elman1990finding}, Long Short-Term Memory Networks (LSTMs) \citep{hochreiter1997long}, and ESIM \citep{chen2017enhanced}, have achieved promising results on NLI tasks. Recently, transformer architectures such as BERT \citep{devlin-etal-2019-bert} and GPT \citep{radford2018improving} have demonstrated an unprecedented ability to capture complex semantic structures and relationships. Their capacity to represent words in dense vector spaces, process sequences, and utilize attention mechanisms has made them exceptionally adept at tasks like NLI. In addition to these models, a series of variant models have emerged using the BERT architecture, such as RoBERTa \citep{liu2019roberta}, XLM-R \citep{conneau-etal-2020-unsupervised}, XLNet \citep{yang2019xlnet}, ALBERT \citep{lan2019albert}, DistilBERT \citep{sanh2019distilbert}, ELECTRA \citep{clark2016electra}, and T5 \citep{raffel2020exploring}, and have achieved high performance on datasets such as SNLI \citep{bowman2015large}, MultiNLI \citep{multinli}, FEVER \citep{thorne2018fever}, and ANLI \citep{nie-etal-2020-adversarial}.

Building upon the resounding success of previous transformer-based models on English data, considerable efforts have been made to extend their capabilities to other languages, including Vietnamese. The PhoBERT model \citep{nguyen2020phobert}, for example, represents a significant advance by leveraging the foundation of BERT and fine-tuning it on Vietnamese data, yielding impressive performance across various tasks. Additionally, earlier models trained on multilingual data, including Vietnamese, such as mBERT \citep{devlin-etal-2019-bert}, XLM-R \citep{conneau-etal-2020-unsupervised}, mT5 \citep{xue-etal-2021-mt5}, CafeBERT \citep{do-etal-2024-vlue}, and mBART \citep{liu-etal-2020-multilingual-denoising}, have also demonstrated excellent results on Vietnamese Natural Language Inference (NLI) tasks \citep{huynh-etal-2022-vinli}.

%em viết lại đoạn này
Mixture-of-Experts (MoE) models have emerged as an effective approach for addressing complex NLP tasks, where single models often fail to capture the diverse aspects of data. MoE architectures consist of multiple sub-models (experts), each specializing in specific data types or tasks, along with a routing mechanism to allocate inputs to the appropriate expert. This approach enables efficient handling of complex data distributions and enhances overall performance. MoE has been successfully applied to various NLP tasks, including machine translation \citep{shen2019mixture}, text summarization \citep{ravaut2022summareranker}, and text generation \citep{du2022glam}. For instance, Switch Transformers \citep{fedus2022switch}, GShard \citep{lepikhin2020gshard}, MoCE \cite{GAO2025127422}, and M-BLS \cite{WANG2025126389} demonstrate the effectiveness of MoE architectures in processing large datasets and improving model robustness by leveraging conditional computation for efficient scaling, thus adeptly managing diverse linguistic phenomena. However, training and deploying MoE models require substantial computational resources, and designing effective routing mechanisms remains a primary research challenge.

Unlike large-scale MoE architectures such as Switch Transformers \citep{fedus2022switch}, GShard \citep{lepikhin2020gshard}, which primarily emphasize scalability for general-purpose language modeling, our work introduces NLIMoE tailored for adversarial NLI in a low-resource language. While prior studies have leveraged conditional computation mainly for efficiency, NLIMoE adopts a dynamic routing mechanism specifically aimed at enhancing robustness against adversarial linguistic phenomena. Specifically, NLIMoE integrates a shared transformer encoder (e.g., XLM-RoBERTa) with dynamically selected sub-networks, allowing different experts to specialize in handling syntactic variations, semantic ambiguity, and linguistic noise in Vietnamese. This design enables the model to achieve stronger performance on the adversarial dataset and highlights the potential of MoE architectures for complex inference tasks in low-resource languages.

\section{Dataset}
\label{sect:dataset}

\subsection{Task Definition}
%Chỗ này định nghĩa task vụ trước khi làm

The task of Adversarial Natural Language Inference in the Vietnamese language aims to rigorously evaluate and enhance the robustness of language models in discerning logical semantic relationships based on complex and adversarial linguistic constructs. This task is defined as follows.
\begin{itemize}
    \item Input: A pair of Vietnamese sentences, comprising a premise, extracted from diverse news articles, and a corresponding hypothesis, meticulously crafted to probe model reasoning capabilities.
    \item Output: A label of the logical relationship between the premise and hypothesis into one of three classes: entailment, where the hypothesis is logically derivable from the premise; contradiction, where the hypothesis negates the premise; or neutral, where the hypothesis neither supports nor contradicts the premise.
\end{itemize}

\subsection{Dataset Creation}

\begin{figure}[]
    \centering
    \includegraphics[width=0.8\linewidth]{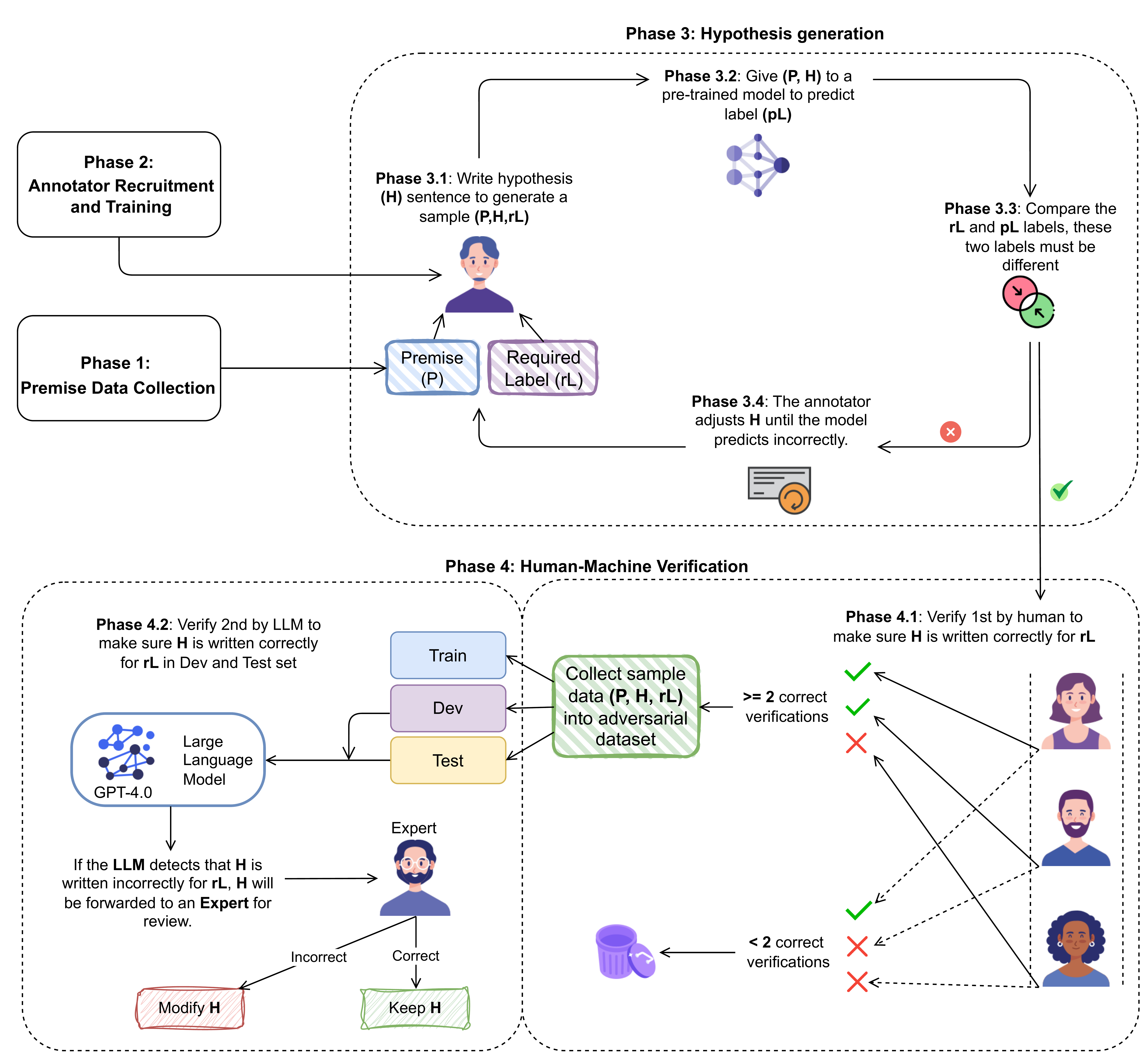}
    \caption{Overall Process for the Creation of the ViANLI Dataset.}
    \label{fig:avinli-creation}
\end{figure}

We propose a new approach using the human-and-machine-in-the-loop integrated with human-machine verification for constructing the ViANLI dataset. This approach integrates both human annotators and machine learning models to generate and verify data, addressing the challenges of current data collection methods related to the complexity and quality of the dataset. The process is illustrated in Figure \ref{fig:avinli-creation} and consists of four key phases: Premise Data Collection (see Section \ref{Premise-Data-Collection}), Annotator Recruitment and Training (see Section \ref{Annotator-Recruitment-and-Training}), Hypothesis Generation (see Section \ref{Hypothesis-Generation}), and Human-Machine Verification (see Section \ref{Human-Machine Verification}).

\subsubsection{Phase 1: Premise Data Collection}
\label{Premise-Data-Collection}

In the creation of the ViANLI dataset, the premise sentences were extracted from news content sourced from VnExpress\footnote{https://vnexpress.net/}, a popular Vietnamese online news platform that covers a wide range of topics. The news content was selected to ensure diversity in the premise sentences, with a focus on 13 domains such as business, tourism, technology, news, science, entertainment, law, sports, health, life, education, world, and vehicles. This diversity provides a solid foundation for generating hypothesis sentences that cover different topics and complexity levels.

\subsubsection{Phase 2: Annotator Recruitment and Training}
\label{Annotator-Recruitment-and-Training}

Fifteen students participated in the ViANLI dataset generation process. The students we hired were from famous universities in Vietnam. They are native speakers with good language skills, enough knowledge, and awareness to read premise sentences and write hypothesis sentences. Similar to the ViNLI dataset, we built an initial guideline to teach students how to write hypothesis sentences to create hypothesis-premise adversarial pairs of sentences. In addition, annotators were explicitly trained to use diverse inference categories (Numerical and Quantitative Reasoning; Reference and Names; Standard Logical Inferences; Lexical Reasoning; Tricky Linguistic Inferences; External Knowledge Reasoning; and Cultural and Contextual Inferences) to ensure that the generated pairs achieved a high level of adversarial difficulty. These types of inferences are clearly defined in Section \ref{sec:Inference_Types}.

Annotators went through at least five rounds of hypothesis writing to be able to participate in the development of ViANLI officially. In each round, they have to write 50 hypotheses corresponding to 50 premise sentences covered on three labels: entailment, contradiction, and neutral. The data completed at each round of annotators is manually checked by us to ask them to correct any errors, and it was also time for them to practice writing their sentences. In the process of checking the data, we also constantly update and edit the guidelines to ensure consistency between the annotators. Besides, we used the XLM-R$_{Large}$ baseline model, which achieved the highest performance on ViNLI to evaluate the data annotators generated in each round. We always required annotators to write hypotheses on a high difficulty level so that the prediction success rate on XLM-R$_{Large}$ is below 40\%. We returned the prediction results of the model to the annotators to help them study how the model encounters difficulties with the data characteristics to improve writing for the next time. Once the annotators’ data generation capabilities stabilized, they were selected to formally participate in our data-building process. At the end of this process, we selected 15 students, as mentioned above. They were paid \$0.025 for writing a hypothesis.

\subsubsection{Phase 3: Hypothesis Generation}
\label{Hypothesis-Generation}

\begin{itemize}
    \item \textbf{Phase 3.1 - Writing Hypothesis:} Annotators are provided with a premise (P) and tasked with writing a corresponding hypothesis (H) for the required label (rL), which can be one of three labels: entailment, contradiction, or neutral. To write complex hypothesis sentences that challenge machine learning models, they may incorporate various types of inferences, such as numerical and quantitative reasoning, coreference and name-based inference, basic logical relations (e.g., negation, causality, comparison), lexical inference (e.g., synonyms, antonyms), tricky inferences (e.g., wordplay, syntactic transformations), reasoning based on external knowledge, and cultural or contextual inferences grounded in Vietnamese norms.
    
    \item \textbf{Phase 3.2 - Model Prediction:} The (P, H) pair is input into a pre-trained machine learning model on a natural language inference task to predict an inference label (pL).
    
    \item \textbf{Phase 3.3 - Label Comparison:} After model prediction, the predicted label (pL) is compared with the required label (rL) to evaluate the ability of the model to infer the semantic relationship between the premise and hypothesis. If the two labels match, it means that the model has correctly identified the semantic relationship, and the sample is considered not complex enough to challenge the model. In this case, the H is moved to Phase 3.4 to adjust the hypothesis sentence and make it more complex. On the other hand, if the labels differ, it indicates that the sample is sufficiently complex to mislead the model. To ensure the accuracy of the generated sample, it is moved to Phase 4, where the validation process begins.
    
    \item \textbf{Phase 3.4 - Annotator Adjustment:} In cases where the model predicts correctly, annotators adjust the hypothesis (H) until the model produces the incorrect prediction (Return to Phase 3.1). This phase focuses on refining the hypothesis to expose the weaknesses of the model.
\end{itemize}

The hypothesis generation process is conducted over three rounds, with each round utilizing a different pre-trained NLI model in Phase 3.2 to predict inference labels for the premise–hypothesis pairs. We selected models with progressively stronger inference capabilities, as reported in the ViNLI benchmark \cite{huynh-etal-2022-vinli}, for each round. This gradual increase in model strength is intended to encourage annotators to generate more diverse and challenging sentence pairs that are difficult enough to mislead the model. Furthermore, the data generated in each round is used to fine-tune a new model, which is subsequently used to evaluate the hypotheses in the next round. The models and their corresponding training datasets for each round are summarized in Table \ref{tab:modelphase2}.

\begin{table}[H]
\centering
\caption{Fine-tuning Models and the Corresponding Training Datasets Used in Each Round.}
\label{tab:modelphase2}
\resizebox{0.8\columnwidth}{!}{%
\begin{tabular}{clr}
\hline
\textbf{Round} & \multicolumn{1}{c}{\textbf{Fine-tuning Model}} & \multicolumn{1}{c}{\textbf{Training Dataset}} \\ \hline
1 & mBERT        & ViNLI, XNLI                                  \\ 
2 & PhoBERT$_{Large}$ & ViNLI, XNLI, VnNewsNLI, ViA1                 \\ 
3 & XLM-R$_{Large}$   & ViNLI, XNLI, VnNewsNLI, VLSP2021-NLI, ViA1, ViA2 \\ \hline
\end{tabular}%
}
\end{table}

\begin{itemize}
    \item \textbf{Round 1 - ViA1 Data:} In the first round, We fine-tuned the mBERT model \citep{devlin-etal-2019-bert}, which was trained on the ViNLI dataset \citep{huynh-etal-2022-vinli}, along with 7,500 Vietnamese samples from the XNLI dataset \citep{conneau2018xnli}, to predict labels for the premise-hypothesis pairs generated by the annotators. The data generated in this round is referred to as ViA1.
    
    \item \textbf{Round 2 - ViA2 Data:} In the second round, we utilized the PhoBERT model \citep{nguyen2020phobert}, which outperformed mBERT on various tasks. The model was trained on a combination of the ViNLI dataset, 7,500 Vietnamese samples from XNLI, and the data generated in Round 1 (ViA1). To improve its understanding of Vietnamese NLI tasks, we also included the VnNewsNLI dataset \citep{nguyen2022building}. The data generated in this round is referred to as ViA2.
    
    \item \textbf{Round 3 - ViA3 Data:} In the third round, the premise sentences remained the same as in the previous rounds. In this round, we used the XLM-R model \citep{conneau-etal-2020-unsupervised} to predict the inference labels for the new premise-hypothesis pairs. To improve the robustness of the model, we trained XLM-R on a larger dataset compared to the previous rounds. The training data for XLM-R included the ViNLI dataset, 7,500 Vietnamese samples from XNLI, data generated in Rounds 1 (ViA1) and 2 (ViA2), the VnNewsNLI dataset, and over 10,000 pairs from the VLSP2021-NLI dataset \citep{JCSCE}. The data generated in this round is referred to as ViA3.
\end{itemize}

\subsubsection{Phase 4: Human-Machine Verification}
\label{Human-Machine Verification}

\begin{itemize}
    \item \textbf{Phase 4.1 - First verification by human:} The first verification ensures that the hypothesis (H) sentence is written correctly according to the required label (rL). In this phase, each premise-hypothesis pair (P, H) is relabeled by three other annotators. If two or more annotators agree with the original annotator, the sample is considered correct and added to the dataset. Conversely, if fewer than two annotators agree with the original annotator, the sample is removed.
    \item \textbf{Phase 4.2 - Second verification by machine and expert:} We perform an additional round of annotation on test and development patterns to ensure accurate labelling. The hypotheses (H) in the development and test sets are verified to ensure it is correctly written according to the required label (rL) for creating accurate premise-hypothesis pairs. A Large Language Model (LLM), specifically GPT-4 \citep{achiam2023gpt}, is employed to evaluate the correctness of H. If the LLM determines that H is written incorrectly for rL, the sample is forwarded to an expert for review. The expert then assesses H: if deemed correct, H is retained as is; if incorrect, the expert modifies H to align with rL, ensuring the quality of the dataset.
\end{itemize}

\section{Dataset Analysis}
\label{sect:dataset_analysis}

In this section, we analyze ViANLI from multiple perspectives to better understand its characteristics. We provide statistical insights into dataset size and the distribution of sentence lengths in both premises and hypotheses. We further examine word repetition and the introduction of new vocabulary in hypothesis generation. In particular, we highlight the linguistic features captured through diverse inference types, which play a crucial role in assessing vocabulary diversity and contribute significantly to the overall difficulty of the dataset.

\subsection{Initial Statistic}
After completing the data construction process, we obtained three datasets (ViA1, ViA2, ViA3) corresponding to the three rounds. Combining the data from these rounds, we created the Vietnamese adversarial dataset (ViANLI), which consists of over 10,000 premise-hypothesis pairs sourced from more than 700 online news articles. Statistical analysis reveals that the distribution of labels across the training, development, and test sets is balanced, a result of our deliberate efforts during data construction to minimize bias and enhance the learning capability of models. Additionally, we categorized the samples by various text genres, with the business genre having the highest number of instances at 947, while the vehicles genre had the lowest at 605. Nevertheless, the variation in sample counts across the 13 genres remains relatively minor overall. Detailed statistics of the dataset are presented in Table \ref{tab:Dataset_statistics}. The data is distributed in a ratio of 8:1:1 for the training, development, and test sets, respectively.

\begin{table}[H]
\centering
\caption{Dataset Statistics of the ViANLI Dataset.}
\begin{tabular}{llrrrr}
\hline
\multicolumn{2}{l}{\textbf{Dataset Statistic}} & \multicolumn{1}{c}{\textbf{Train}} & \multicolumn{1}{c}{\textbf{Dev}} & \multicolumn{1}{c}{\textbf{Test}} & \multicolumn{1}{c}{\textbf{Total}} \\ \hline
\multicolumn{1}{l}{\multirow{4}{*}{\#Sample}} & ViA1 (Round1) & 2,609 & 330 & 330 & 3,269 \\ 
\multicolumn{1}{l}{} & ViA2 (Round2) & 2,672 & 330 & 330 & 3,332 \\ 
\multicolumn{1}{l}{} & ViA3 (Round3) & 2,731 & 340 & 340 & 3,411 \\
\multicolumn{1}{l}{} & ViANLI (Total) & \textbf{8,012} & \textbf{1,000} & \textbf{1,000} & \textbf{10,012} \\ \hline
\multicolumn{1}{l}{\multirow{3}{*}{\#Label}} & Entailment & 2,615 & 334 & 334 & 3,283 \\ 
\multicolumn{1}{l}{} & Contradiction & 2,473 & 333 & 333 & 3,139 \\ 
\multicolumn{1}{l}{} & Neutral & 2,924 & 333 & 333 & 3,590 \\ \hline
\multicolumn{1}{l}{\multirow{13}{*}{\#Topic}} & Business & 783 & 74 & 90 & 947 \\ 
\multicolumn{1}{l}{} & Tourism & 717 & 89 & 101 & 907 \\ 
\multicolumn{1}{l}{} & Technology & 690 & 78 & 72 & 840 \\ 
\multicolumn{1}{l}{} & News & 674 & 92 & 83 & 849 \\ 
\multicolumn{1}{l}{} & Science & 673 & 84 & 87 & 844 \\ 
\multicolumn{1}{l}{} & Entertainment & 603 & 90 & 71 & 764 \\ 
\multicolumn{1}{l}{} & Law & 603 & 80 & 64 & 747 \\ 
\multicolumn{1}{l}{} & Sports & 602 & 70 & 77 & 749 \\ 
\multicolumn{1}{l}{} & Health & 550 & 72 & 73 & 695 \\ 
\multicolumn{1}{l}{} & Life & 548 & 76 & 79 & 703 \\  
\multicolumn{1}{l}{} & Education & 544 & 69 & 72 & 685 \\  
\multicolumn{1}{l}{} & World & 536 & 62 & 79 & 677 \\ 
\multicolumn{1}{l}{} & Vehicles & 489 & 64 & 52 & 605 \\ \hline
\end{tabular}
\label{tab:Dataset_statistics}
\end{table}

\begin{table}[H]
\centering
\caption{The Model Error Rate Represents the Percentage of the Incorrect Predictions of the Model on the Data per Round in Phase 3. Verify 1 Success Rate Is the Percentage of Successful Confirmations on the Data Samples That the Data Model Predicts Incorrectly in Phase 4.1.}
\begin{tabular}{lcc}
\hline
\textbf{Dataset} & \textbf{Model error rate (\%)} & \textbf{Verify 1 success rate (\%)} \\ \hline
ViA1 (Round1) & 51.65 & 77.59 \\ 
ViA2 (Round2) & 44.93 & 94.27 \\ 
ViA3 (Round3) & 41.98 & 94.71 \\ \hline
ViANLI (Total) & 46.18 & 88.86 \\ \hline
\end{tabular}
\label{tab:ModelErrorRate_Verify1Success}
\end{table}

Furthermore, as outlined in the data-building process for each round, we employed progressively stronger models to evaluate the data generated in each round. The goal was to identify more challenging adversarial patterns by focusing on the samples that the model misclassifies. The model error rate decreases with each round, indicating that the accuracy of the models in Phase 3.2 improves over time as shown in Table \ref{tab:ModelErrorRate_Verify1Success}. This means that the data pairs that mislead the model and are misclassified in later rounds become increasingly difficult and complex.

Additionally, the success rate of Verification 1 increased significantly from 77.59\% in Round 1 to 94.71\% in Round 3. This shows that the number of premise-hypothesis pairs that successfully fool the model and are confirmed as correct by the annotators has risen, reflecting an improvement in the annotators’ hypothesis generation quality over successive rounds. Each data sample for each round is illustrated in Table \ref{tab:Examples} (in \ref{ExamplesfromViANLIDataset}).

\subsection{Length Distribution}
We calculate the sentence lengths of both premise and hypothesis in our dataset using the VnCoreNLP tool \cite{vu2018vncorenlp}, which performs word segmentation for Vietnamese text. After segmentation, we count the number of words in each sentence to determine its length, following the same approach used in the ViNLI dataset. In addition, we compare the sentence lengths in our dataset with those in the ViNLI dataset to analyze differences in sentence complexity and structure.

As shown in Figures \ref{ViA1}, \ref{ViA2}, and \ref{ViA3} in \ref{VisualLengthOfPremiseAndHypothesis} (corresponding to the data from Round 1, Round 2, and Round 3), the length of the hypothesis sentences is distributed within a shorter range compared to the premise sentences. These three charts also demonstrate that the annotators’ writing remains consistent in terms of difficulty, with the hypothesis sentences tending to be\ shorter than the premise sentences.

When comparing ViANLI with ViNLI, Figures \ref{ViANLI} and \ref{ViNLI} in \ref{VisualLengthOfPremiseAndHypothesis} show a common trend: the length of hypothesis sentences is generally shorter than that of premise sentences in both datasets. However, a closer look at these graphs and the data presented in Table \ref{tab:meanlength} reveals that the hypothesis sentences in ViANLI are significantly shorter than those in ViNLI, while the premise sentence lengths are quite similar across both datasets.

\begin{table}[]
\centering
% \resizebox{0.9\textwidth}{!}{
\caption{The Mean Length of Premise and Hypothesis Sentences on ViANLI and ViNLI.}
\label{tab:meanlength}
\begin{tabular}{lcrrrr}
\hline
\multirow{2}{*}{\textbf{Mean word length}} & \multicolumn{5}{c}{\textbf{Dataset}} \\ \cline{2-6} 
 & \multicolumn{1}{c}{\textbf{ViA1}} & \multicolumn{1}{c}{\textbf{ViA2}} & \multicolumn{1}{c}{\textbf{ViA3}} & \multicolumn{1}{c}{\textbf{ViANLI}} & \multicolumn{1}{c}{\textbf{ViNLI}} \\ \hline
Premise & \multicolumn{1}{r}{24.4} & \multicolumn{1}{r}{25.3} & \multicolumn{1}{r}{25.0} & \multicolumn{1}{r}{24.9} & 24.5 \\ 
Hypothesis & \multicolumn{1}{r}{14.3} & \multicolumn{1}{r}{14.4} & \multicolumn{1}{r}{14.6} & \multicolumn{1}{r}{14.4} & 18.1 \\ \hline
\end{tabular}
% }
\end{table}

The authors of MultiNLI \citep{multinli} and ViNLI \citep{huynh-etal-2022-vinli} suggest that shorter hypothesis sentence lengths, such as those found in the SNLI dataset, may lead to more predictable data. While the hypothesis sentences in ViANLI are indeed shorter than those in ViNLI, we have ensured the challenge and difficulty of the dataset through model error rate evaluations in each round, as shown in Table \ref{tab:ModelErrorRate_Verify1Success}. The impact of sentence length on model accuracy is further analyzed in Section \ref{lengthToAcc}.

% \begin{figure}[ht]
%     \centering
%     \begin{subfigure}{0.33\textwidth}
%         \includegraphics[width=\textwidth]{images/lenght_of_P_H_ViANLI.pdf}
%         \caption{ViVQA}
%     \end{subfigure}
%     \begin{subfigure}{0.33\textwidth}
%         \includegraphics[width=\textwidth]{images/lenght_of_P_H_ViNLI.pdf}
%         \caption{OpenViVQA}
%     \end{subfigure}
%     \caption{Comparison of question length among VQA datasets.}
%     \label{fig:datasets-question-length-statistics}
% \end{figure}

\subsection{Word Overlap Rate}
We calculate the word overlap rate (WOR) between the premise and hypothesis sentences in the ViANLI adversarial dataset to investigate the extent to which annotators reuse vocabulary from the premise sentence when constructing the hypothesis sentence. This analysis is crucial for assessing the difficulty of the dataset, as previous studies \cite{huynh-etal-2022-vinli, van2022error} have indicated that word overlap can significantly impact the predictive ability of models. To measure the word overlap, we use the Jaccard metric, which calculates the unordered word overlap between the premise and hypothesis. Before calculating the Jaccard value, we apply the VnCoreNLP toolkit for Vietnamese word segmentation. The Jaccard measure between a pair of premise and hypothesis sentences is calculated by the Equation (\ref{jaccard}).

\begin{equation}
\label{jaccard}
    Jaccard = {\lvert P \cap H \rvert \over \lvert P \rvert + \lvert H \rvert - \lvert P \cap H \rvert}
\end{equation}

\begin{figure}[H]
    \centering
    \begin{subfigure}{1\textwidth}
        \includegraphics[width=\textwidth]{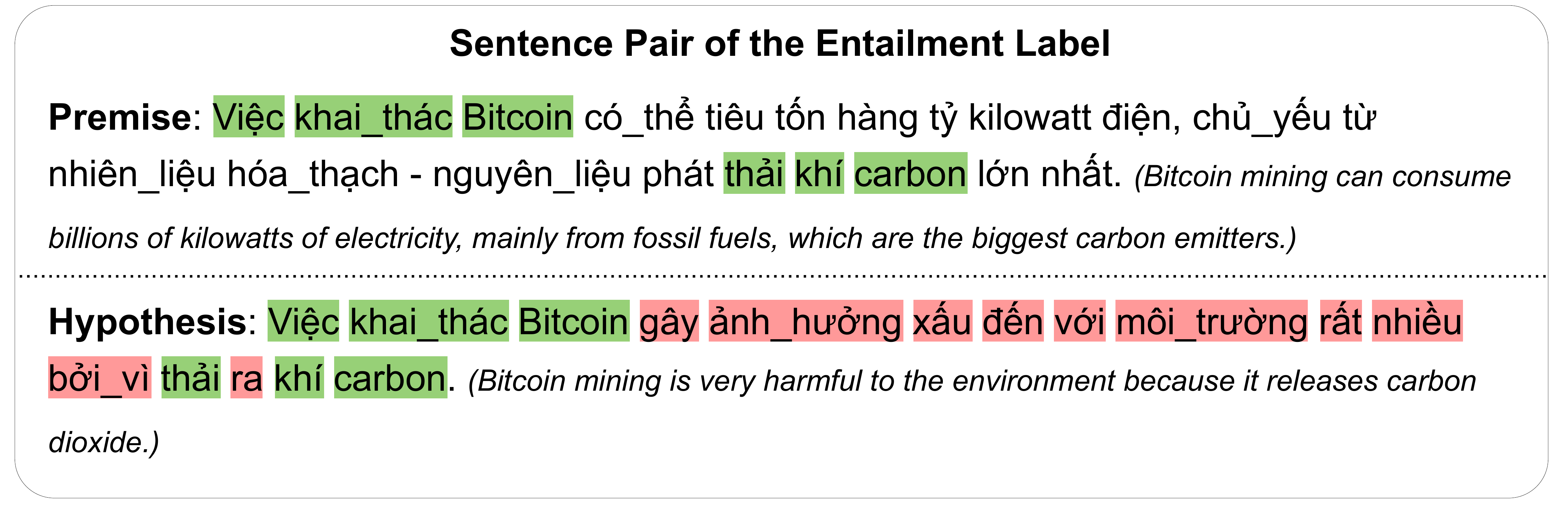}
    \end{subfigure}
    \caption{Example of Word Overlap in the Premise-hypothesis Pair of the Entailment Label. Where the Green Highlight Is the Same Words, the Red Highlight Is the New Words That Only Appear in the Hypothesis Sentence.}
    \label{fig:example_word_overlap}
\end{figure}

 Where, premise and hypothesis sentences are represented in two lists of words, respectively, as follows: P = [W$_{P1}$, W$_{P2}$, W$_{P3}$, ..., W$_{Pn}$], H = [W$_{H1}$, W$_{H2}$, W$_{H3}$, ..., W$_{Hm}$]. $\lvert$P$\cap$H$\rvert$ is the number of words that appear in both premise and hypothesis sentences. $\lvert$P$\rvert$ is the number of words in the sentence premise. $\lvert$H$\rvert$ is the number of words in the sentence hypothesis. $\lvert$P$\cup$H$\rvert$ is the number of words in the premise and hypothesis sentences after the duplicates are removed. An example of word overlap between the premise and hypothesis sentences for the entailment label is shown in Figure \ref{fig:example_word_overlap}.

Table \ref{tab:wordOverlap} presents the average Jaccard value for each dataset after calculating the word overlap for each premise-hypothesis pair. In general, when observing the word overlap rate across each round of data and the entire ViANLI dataset, we found that the overlap rate for the entailment label is the lowest, compared to the higher overlap rates for the contradiction and neutral labels. This suggests that when writing entailment hypotheses, annotators tend to use fewer words directly from the premise sentence. This distribution trend in Table \ref{tab:wordOverlap} overcomes the weaknesses of previous studies, such as ViNLI \cite{van2022error}, which has shown that models tend to predict semantic relationships more accurately when there is a high word overlap rate, particularly for the entailment label. Therefore, to enhance the adversarial nature of ViANLI, annotators were deliberately intentional to limit the reuse of premise words when crafting hypothesis sentences, making it more challenging for models to rely solely on lexical overlap for accurate predictions. To gain a more precise understanding of the impact of word overlap on model accuracy, we provide further analysis in Section \ref{WordOverLapToAcc}.

\begin{table}[h]
\centering
\caption{The Word Overlap Rate in ViANLI Compares to That in ViNLI.}
\label{tab:wordOverlap}
% \resizebox{0.8\textwidth}{!}{
\begin{tabular}{lrrrrr}
\hline
\multicolumn{1}{c}{\multirow{2}{*}{\textbf{Label}}} & \multicolumn{5}{c}{\textbf{Jaccard (\%)}} \\ \cline{2-6} 
\multicolumn{1}{c}{} & \multicolumn{1}{c}{\textbf{ViA1}} & \multicolumn{1}{c}{\textbf{ViA2}} & \multicolumn{1}{c}{\textbf{ViA3}} & \multicolumn{1}{c}{\textbf{ViANLI}} & \multicolumn{1}{c}{\textbf{ViNLI}} \\ \hline
Entailment & \multicolumn{1}{r}{23.72} & \multicolumn{1}{r}{21.36} & \multicolumn{1}{r}{20.62} & \multicolumn{1}{r}{21.85} & 29.88 \\ 
Contradiction & \multicolumn{1}{r}{32.80} & \multicolumn{1}{r}{30.53} & \multicolumn{1}{r}{30.56} & \multicolumn{1}{r}{31.23} & 23.30 \\
Neutral & \multicolumn{1}{r}{30.82} & \multicolumn{1}{r}{31.97} & \multicolumn{1}{r}{30.92} & \multicolumn{1}{r}{31.22} & 20.19 \\ \hline
\end{tabular}
% }
\end{table}

\subsection{New Word Rate}

In contrast to the word overlap rate, we analyze the new word rate in ViANLI. This refers to the proportion of new words in the hypothesis sentence—words that do not appear in the premise sentence. This statistic provides insight into the linguistic diversity of annotators when generating hypothesis sentences. We calculate the ratio of new words for each round and for the entire ViANLI dataset. The VnCoreNLP toolkit is also applied to segment the Vietnamese text before calculating the new word rate. The results are presented in Table \ref{tab:newWordRate}, broken down by the three labels: entailment, contradiction, and neutral.

The new word rate for the entailment hypothesis is the highest, while the lowest rate is observed in the contradiction hypothesis, both for the per-round data and the entire ViANLI dataset. When comparing ViANLI with ViNLI, notable differences are observed. In the ViNLI dataset, the highest new word rate is associated with the neutral label, and the lowest rate with the entailment label. Additionally, except for the fact that the proportion of new words in the entailment label of ViANLI is higher than that of ViNLI, the new word rate for the contradiction and neutral labels in ViANLI is significantly lower than in ViNLI. The new word rate is also an important factor influencing the performance of machine learning models, and its impact is further analyzed in Section \ref{NewWordRateToAcc}.

\begin{table}[]
\centering
\caption{The New Word Rate in ViANLI Compares to That in ViNLI.}
\label{tab:newWordRate}
% \resizebox{0.8\textwidth}{!}{
\begin{tabular}{lrrrrr}
\hline
\multicolumn{1}{c}{\multirow{2}{*}{\textbf{Label}}} & \multicolumn{5}{c}{\textbf{New word rate (\%)}} \\ \cline{2-6} 
\multicolumn{1}{c}{} & \multicolumn{1}{c}{\textbf{ViA1}} & \multicolumn{1}{c}{\textbf{ViA2}} & \multicolumn{1}{c}{\textbf{ViA3}} & \multicolumn{1}{c}{\textbf{ViANLI}} & \multicolumn{1}{c}{\textbf{ViNLI}} \\ \hline
Entailment & \multicolumn{1}{r}{50.95} & \multicolumn{1}{r}{53.44} & \multicolumn{1}{r}{54.31} & \multicolumn{1}{r}{52.95} & 46.59 \\ 
Contradiction & \multicolumn{1}{r}{36.82} & \multicolumn{1}{r}{39.78} & \multicolumn{1}{r}{40.61} & \multicolumn{1}{r}{39.17} & 53.96 \\ 
Neutral & \multicolumn{1}{r}{41.65} & \multicolumn{1}{r}{40.21} & \multicolumn{1}{r}{42.10} & \multicolumn{1}{r}{41.34} & 61.79 \\ \hline
\end{tabular}
% }

\end{table}

% \begin{table}[H]
% \centering
% \resizebox{\columnwidth}{!}{%
% \begin{tabular}{lrrrrrrr}
% \hline
% \multicolumn{1}{c}{\multirow{2}{*}{\textbf{Label}}} & \multicolumn{1}{c}{\multirow{2}{*}{\textbf{\begin{tabular}[c]{@{}c@{}}New word \\ rate (\%)\end{tabular}}}} & \multicolumn{6}{c}{\textbf{Part-Of-Speech (\%)}} \\ \cline{3-8} 
% \multicolumn{1}{c}{} & \multicolumn{1}{c}{} & \multicolumn{1}{c}{\textbf{Noun}} & \multicolumn{1}{c}{\textbf{Verb}} & \multicolumn{1}{c}{\textbf{Adjective}} & \multicolumn{1}{c}{\textbf{Preposition}} & \multicolumn{1}{c}{\textbf{Adjunct}} & \multicolumn{1}{c}{\textbf{Other}} \\ \hline
% Entailment & 53.23 & \multicolumn{1}{r}{32.04} & \multicolumn{1}{r}{26.36} & \multicolumn{1}{r}{8.24} & \multicolumn{1}{r}{8.23} & \multicolumn{1}{r}{9.48} & 15.65 \\ 
% Contradiction & 38.59 & \multicolumn{1}{r}{28.95} & \multicolumn{1}{r}{27.99} & \multicolumn{1}{r}{8.37} & \multicolumn{1}{r}{8.95} & \multicolumn{1}{r}{10.83} & 14.91 \\ 
% Neutral & 40.67 & \multicolumn{1}{r}{27.77} & \multicolumn{1}{r}{27.84} & \multicolumn{1}{r}{8.74} & \multicolumn{1}{r}{9.29} & \multicolumn{1}{r}{11.12} & 15.24 \\ \hline
% \end{tabular}%
% }
% \caption{The new word rate by Part-Of-Speech in ViANLI.}
% \label{tab:Part-Of-Speech}
% \end{table}

\begin{figure}[H]
    \centering
    \begin{subfigure}{1\textwidth}
        \includegraphics[width=\textwidth]{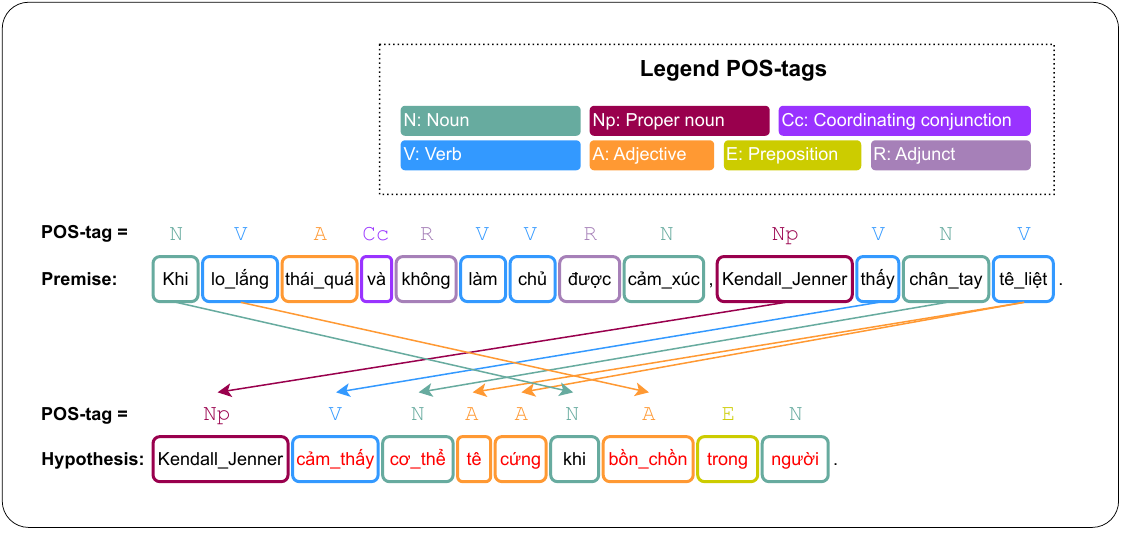}
    \end{subfigure}
    \caption{Example of POS-tag in the Premise-hypothesis Pair of the Entailment Label. The Red Words in the Hypothesis Tree Are New Words, and the Arrows Represent Changes in the Position and Function of the Word.}
    \label{fig:example_POS}
\end{figure}

\begin{figure}[H]
    \centering
    \begin{subfigure}{1\textwidth}
        \includegraphics[width=\textwidth]{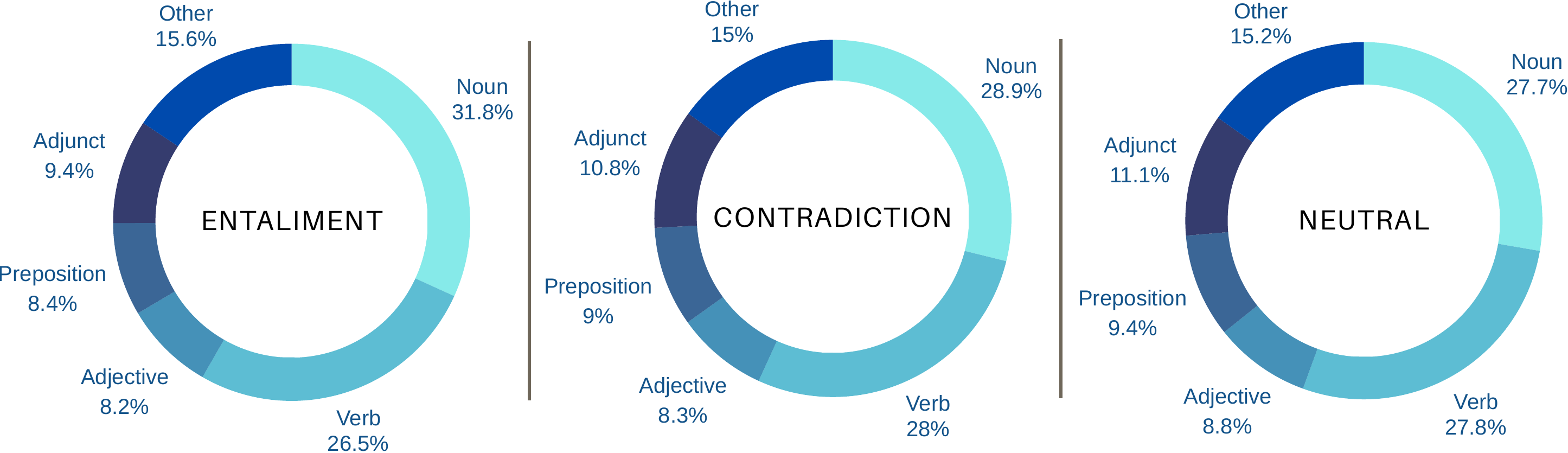}
    \end{subfigure}
    \caption{The New Word Rate by Part-of-Speech in ViANLI.}
    \label{fig:newWR}
\end{figure}

To better understand the trend in word usage by annotators when writing hypothesis sentences, we use PhoNLP \citep{nguyen2021phonlp} to identify the part of speech (POS) of new words in the hypothesis sentences. We then calculate the percentage of each POS category for visualization. PhoNLP recognizes several POS categories, but for clarity, we summarize them into main word types, including Nouns, Verbs, Adjectives, Prepositions, Adjuncts, and Others (combining the remaining types). Figure \ref{fig:newWR} illustrates the new word rate by Part-of-Speech in ViANLI. Statistical results reveal that the usage of words is quite consistent across all three labels, with Nouns and Verbs representing the largest proportions. Furthermore, annotators tend to use more Nouns when writing entailment hypotheses. On the other hand, adjectives are rarely used when writing hypothesis sentences. An illustrative example of the POS-tag analysis on new words in a hypothesis sentence is shown in Figure \ref{fig:example_POS}.

% \begin{figure}[H]
%     \centering
%     \begin{subfigure}{1\textwidth}
%         \includegraphics[width=\textwidth]{Part-Of-Speech.pdf}
%     \end{subfigure}
%     \caption{The new word rate by Part-Of-Speech in ViANLI.}
%     \label{fig:datasets-question-length-statistics}
% \end{figure}

\subsection{Inference Types}
\label{sec:Inference_Types}

To investigate the linguistic characteristics of the ViANLI dataset, we conducted a detailed analysis of the inference types present in the development sets of each round (ViA1, ViA2, and ViA3) as well as the entire ViANLI development set. We categorized the premise-hypothesis pairs into seven distinct inference types. Six of these types were adapted from the inference ontology proposed in the ANLI study \cite{nie-etal-2020-adversarial}, with descriptions tailored to account for the nuances of the Vietnamese language. Additionally, recognizing the unique cultural and contextual features of Vietnamese, we introduced a seventh inference type, cultural and contextual inference, to capture reasoning based on cultural knowledge, such as understanding social hierarchies, traditional practices, or context-specific implicatures, which are prevalent in Vietnamese communication.

The seven inference types used in our analysis are defined as follows:
\begin{itemize}

\item \textbf{Numerical and Quantitative Reasoning (NaQR):} Inferences involving numerical information, such as cardinal and ordinal numbers, dates, ages, or quantitative comparisons, etc. (e.g., "60 minutes is 1 hour", "from February 1 to February 6 lasts 5 days").

\item \textbf{Reference and Names (RaN):} Inferences requiring resolution of coreference (e.g., linking pronouns like "anh ấy - he" or "cô ấy - she" to specific individuals) or understanding attributes associated with proper names, such as gender, etc.

\item \textbf{Standard Logical Inferences (SLI):} Reasoning based on logical structures, including conjunctions (e.g., "và - and"), negations (e.g., "không - not"), cause-and-effect relationships, comparatives, or superlatives, etc.

\item \textbf{Lexical Reasoning (LR):} Inferences derived from lexical relationships, such as synonyms (e.g., "vui vẻ - happy" and "hạnh phúc - delighted"), antonyms (e.g., "lớn - large" vs. "nhỏ - small"), etc.

\item \textbf{Tricky Linguistic Inferences (TLI):} Inferences that involve sophisticated linguistic strategies, such as wordplay, syntactic restructuring, or deducing the speaker’s intent from subtle contextual cues, etc.

\item \textbf{External Knowledge Reasoning (EKR):} Inferences that rely on external or world knowledge not explicitly stated in the premise, such as geographical facts (e.g., "You cannot travel directly from Hanoi to the sea") or general knowledge about Vietnamese society, etc.

\item \textbf{Cultural and Contextual Inference (CaCI):} Inferences grounded in Vietnamese cultural norms, traditions, or implicit meanings unique to the language (e.g., "miền Tây" refers to the Mekong Delta, where people face challenges like salinity intrusion "hạn mặn - salinity drought"), etc.
\end{itemize}

The analysis of inference types utilized by annotators in the ViANLI dataset, as summarized in Table \ref{tab:inference_type}, reveals distinct trends in the creation of adversarial premise-hypothesis pairs across the ViA1, ViA2, and ViA3 rounds, with an overall average across the entire ViANLI Dev Set. Standard Logical Inferences dominate with a consistent average of 72.51\%, indicating a strategic emphasis on basic logical structures to challenge natural language inference (NLI) models. External Knowledge Reasoning remains a stable component, averaging 43.01\% with minor increases (42.35\%–43.63\%), suggesting a reliance on world knowledge to introduce complexity. Notably, Numerical \& Quantitative Reasoning shows a gradual increase from 22.12\% (ViA1) to 27.35\% (ViA3), reflecting an adaptive strategy to incorporate quantitative challenges as models evolve. Conversely, Tricky Linguistic Inferences decline from 14.54\% (ViA1) to 9.71\% (ViA3), hinting at a shift away from intricate linguistic strategies. Cultural \& Contextual Inference, averaging 10.09\% with a peak of 11.21\% (ViA1) and a small dip to 10.09\% (ViA3), underscores the annotators' effort to leverage Vietnam-specific cultural nuances. This trend suggests that annotators initially explore cultural elements but later prioritized logical and quantitative challenges, potentially reflecting model adaptation over the rounds.

\begin{table}[H]
\centering
\caption{Inference Type Distribution (\%) on the Dev Set for Each Round and Averages Across the Entire ViANLI Dev Set.}
\label{tab:inference_type}
\resizebox{0.8\columnwidth}{!}{%
\begin{tabular}{lccccccc}
\hline
\textbf{Round} & \textbf{NaQR} & \textbf{RaN} & \textbf{SLI} & \textbf{LR} & \textbf{TLI} & \textbf{EKR} & \textbf{CaCI} \\ \hline
ViA1           & 22.12                      & 11.21                     & 72.12           & 20.30          & 14.54         & 43.63                     & 11.21                         \\
ViA2           & 24.54                      & 6.06                      & 74.84           & 24.24          & 9.69          & 43.03                     & 8.18                          \\
ViA3           & 27.35                      & 12.64                     & 70.58           & 20.58          & 9.71          & 42.35                     & 10.88                         \\ \hline
ViANLI         & 24.67                      & 9.97                      & 72.51           & 21.71          & 11.31         & 43.01                     & 10.09                         \\ \hline
\end{tabular}%
}
\end{table}

% In addition, annotators may employ one or multiple inference types to write hypothesis sentences that aim to enrich semantic understanding. Combining multiple inference types can make it difficult for the model to recognize the relationship or it can be multiple recognition signals and help the model make better inferences. Figure \ref{fig:numInference} illustrates the proportion of combined inference types in the ViANLI Dev Set. The statistical results show that the combination of two inference types accounts for the highest percentage at 72.6\%, followed by the use of a single inference type at 17.2\%, while the combination of three inference types is the lowest at 10.2\%. The evaluation of the impact of combining inference types on model performance is presented in Section \ref{sec:per_inference_type}.

\begin{figure}[H]
    \centering
    \begin{subfigure}{0.6\textwidth}
        \includegraphics[width=\textwidth]{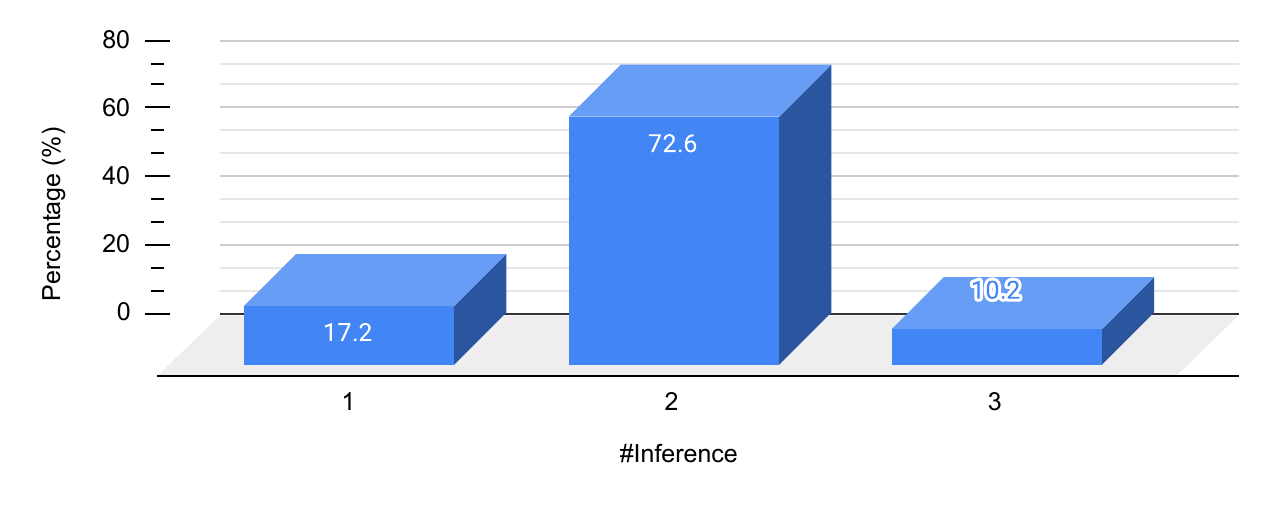}
    \end{subfigure}
    \caption{Analysis of the combination ratio of Inference Types on the ViANLI Dev Set.}
    \label{fig:numInference}
\end{figure}

Annotators were allowed to use one or more inference types when writing hypotheses in order to increase semantic richness and adversarial difficulty. Figure \ref{fig:numInference} shows the distribution of the number of inference types combined in the ViANLI Dev Set. The results indicate that most hypotheses combine two inference types (72.6\%), while a single inference type accounts for 17.2\%, and three inference types account for only 10.2\%. This analysis highlights that annotators often rely on multiple inference strategies rather than a single one. The impact of combining inference types on model performance is further analyzed in Section \ref{sec:per_inference_type}.

\section{Our Proposed Model}
\label{sect:proposed_model}
\begin{figure}[H]
    \centering
    \includegraphics[width=0.9\linewidth]{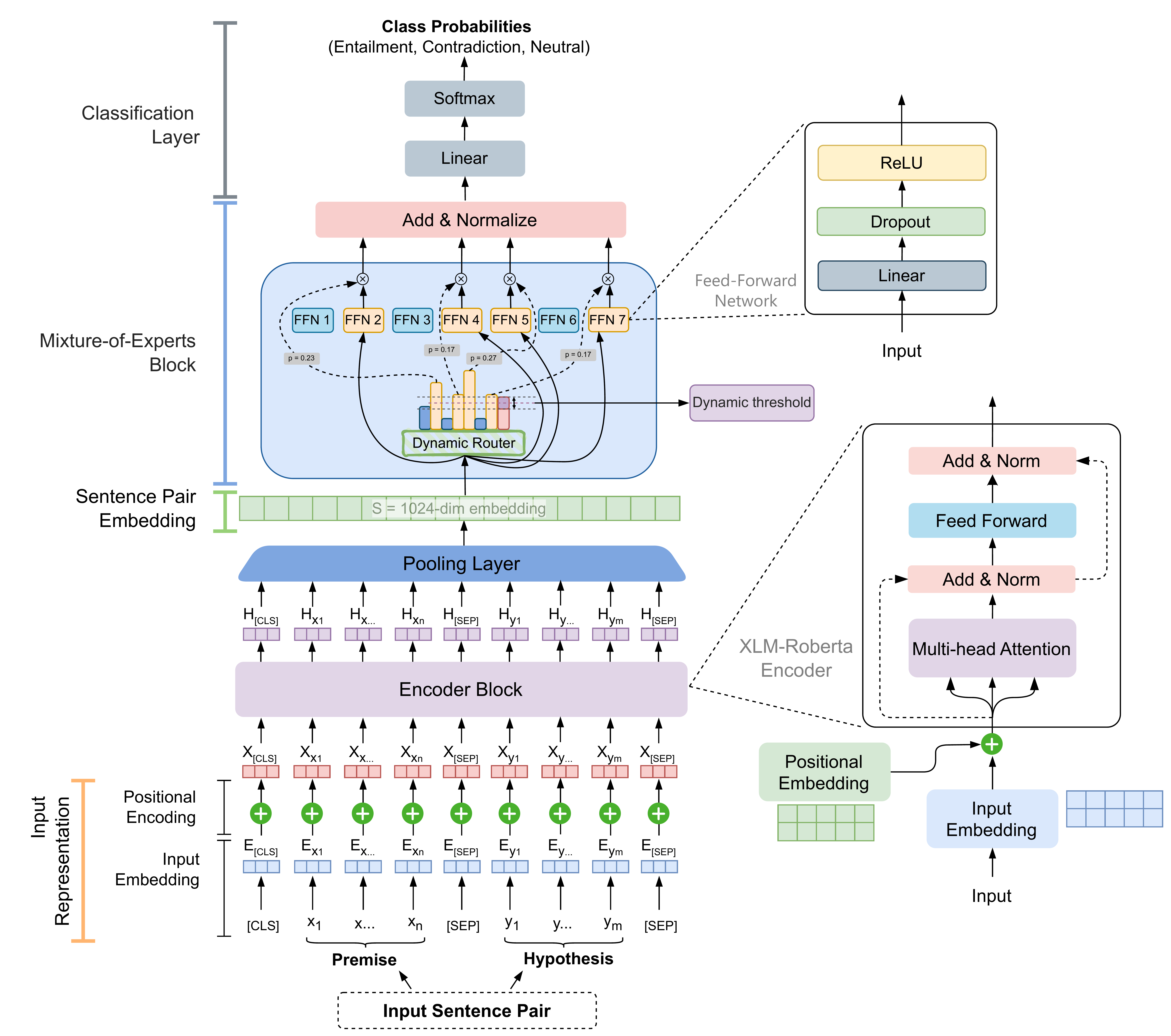}
    \caption{Our Proposed Model NLIMoE - A Mixture of Experts Model for Natural Language Inference Leveraging XLM-RoBERTa Encoder with Dynamic Routing Method.}
    \label{fig:our-model}
\end{figure}

In this section, we present the architecture of our proposed NLIMoE (Mixture-of-Experts language model) for natural language inference, which is designed to effectively handle the complexities and challenges posed by adversarial NLI datasets. This architecture allocates tasks to specialized experts based on the complexity of the input data, enabling the model to focus on different linguistic aspects. By combining the power of XLM-RoBERTa with a dynamic routing mechanism, the model adapts to various linguistic challenges. The architecture of NLIMoE is shown in Figure \ref{fig:our-model}.

Figure \ref{fig:our-model} illustrates the overall architecture of NLIMoE. It consists of several submodules: (i) \textbf{Input Representation}, where the premise and hypothesis are tokenized and embedded, then combined with positional encodings; (ii) \textbf{Encoder Block}, implemented by XLM-RoBERTa, which produces contextualized embeddings; (iii) \textbf{Pooling Layer}, which aggregates token embeddings into a fixed-size sentence pair vector; (iv) \textbf{Mixture-of-Experts Block}, which contains two main components: the \textbf{Dynamic Router} (computing expert selection probabilities and applying a dynamic threshold), the \textbf{Feed-Forward Network Experts} (specialized subnetworks that process the routed inputs); and (v) \textbf{Classification Layer}, which predicts entailment, contradiction, or neutral.

\textbf{Input Representation:} The premise and hypothesis sentences are tokenized into subword units using a pre-trained tokenizer. Let the premise and hypothesis sequences consist of \(n\) and \(m\) tokens, respectively. The tokenized input, including special tokens [CLS], [SEP] (separating premise and hypothesis), and [SEP] (terminating hypothesis), forms a sequence of length \(n + m + 3\). This sequence is embedded into a vector space using the pre-trained XLM-RoBERTa tokenizer, producing a embedding \(\mathbf{E} \in \mathbb{R}^{(n+m+3) \times d}\), where \(d\) is the embedding dimension. Positional encodings \(\mathbf{P} \in \mathbb{R}^{(n+m+3) \times d}\) are added to capture token order, resulting in the combined representation \(\mathbf{X} = \mathbf{E} + \mathbf{P}\), where \(\mathbf{X} \in \mathbb{R}^{(n+m+3) \times d}\) serves as the input to the encoder block.

\textbf{Encoder Block:} The encoder block processes the input representation \(\mathbf{X} \in \mathbb{R}^{(n+m+3) \times d}\) using the XLM-RoBERTa encoder, which consists of multiple transformer layers. Each layer applies multi-head self-attention to capture semantic relationships between the premise and hypothesis tokens, followed by feed-forward networks (FFNs) to transform the attention output into a higher-level representation. The output of the encoder block is a contextualized representation \(\mathbf{H} \in \mathbb{R}^{(n+m+3) \times d}\), where each row corresponds to the contextualized embedding of a token in the input sequence.

\textbf{Pooling Layer:} The pooling layer aggregates the contextualized representation \(\mathbf{H} \in \mathbb{R}^{(n+m+3) \times d}\) from the encoder block into a fixed-size sentence pair embedding. We apply mean pooling to compute the average representation across all tokens in the sequence, including the special tokens [CLS] and [SEP]. Specifically, the sentence pair embedding \(\mathbf{S} \in \mathbb{R}^{d}\) is obtained by averaging all rows of \(\mathbf{H}\) (see Equation (\ref{eqS})).

\begin{equation}
\mathbf{S} = \frac{1}{n+m+3} \sum_{i=1}^{n+m+3} \mathbf{h}_i
\label{eqS}
\end{equation}

where \(\mathbf{h}_i \in \mathbb{R}^{d}\) is the \(i\)-th row of \(\mathbf{H}\). This vector \(\mathbf{S}\) represents a 1024-dimensional embedding that captures the aggregated semantic representation of the premise–hypothesis pair. By integrating information from all tokens, it serves as the input to the Mixture-of-Experts block.

\textbf{Mixture-of-Experts Block:} This block is central to the NLIMoE model, designed to tackle complex linguistic tasks by dynamically assigning inputs to specialized experts.
\begin{itemize}
    \item \textbf{Dynamic Router Mechanism:} The router determines which expert(s) process a given input. We use a softmax function to compute the probability distribution across the available experts based on vector \(\mathbf{S}\), selecting the most appropriate expert based on linguistic features such as syntactic complexity, semantic ambiguity, or adversarial noise. Unlike a static threshold—where the value is predefined and remains fixed throughout the entire training process—we propose a dynamic threshold mechanism consisting of two main steps: (1) computing the expert selection probabilities, and (2) adjusting the Dynamic Threshold based on the complexity of the input.

    To generate gate values and directly normalize them into a probability distribution over experts in a single step using the gate layer $\mathbf{W}_g \in \mathbb{R}^{d \times num\_experts}$ applied to $\mathbf{S}$, followed by the softmax function to obtain the probability distribution over the experts, as calculated in Equation (\ref{probability_distribution}).

    \begin{equation}
    P_i = \frac{\exp\big((\mathbf{W}_g \mathbf{S} + \mathbf{b}_g)_i\big)}{\sum_{j=1}^{num\_experts} \exp\big((\mathbf{W}_g \mathbf{S} + \mathbf{b}_g)_j\big)}
    \label{probability_distribution}
    \end{equation}
    
    where $\mathbf{S} \in \mathbb{R}^d$ is the pooled sentence-pair embedding via Equation (~\ref{eqS}), 
    $\mathbf{W}_g \in \mathbb{R}^{d \times num\_experts}$ and $\mathbf{b}_g \in \mathbb{R}^{num\_experts}$ 
    are the gating parameters, and $P_i$ denotes the probability of selecting expert $i$ for a given input sample.

    To estimate the complexity of the input for dynamic expert selection, we introduce an additional complexity gate layer 
    $\mathbf{W}_c \in \mathbb{R}^{d \times 1}$ with bias $b_c \in \mathbb{R}$. 
    This layer takes the pooled sentence-pair embedding $\mathbf{S} \in \mathbb{R}^d$, 
    which is obtained by averaging the contextualized token representations from the encoder. 
    The complexity score $C$ is then computed by applying a sigmoid activation, as in Equation (\ref{complexity_score}).
    
    \begin{equation}
    C = \sigma(\mathbf{W}_c \mathbf{S} + b_c)
    \label{complexity_score}
    \end{equation}
    
    where $\sigma$ denotes the sigmoid function, which bounds $C$ between 0 and 1.

    The dynamic threshold $\rho_{dynamic}$ is then computed as a combination of a static threshold 
    $\rho_{static}$ and the complexity score $C$ (defined in Equation (\ref{complexity_score})), scaled by a coefficient $\gamma$, 
    as in Equation (\ref{dynamic_threshold}).
    
    \begin{equation}
    \rho_{dynamic} = \rho_{static} + \gamma \cdot C
    \label{dynamic_threshold}
    \end{equation}
    
    where $\rho_{static}$ is a fixed baseline threshold, $\gamma$ is a scaling coefficient, 
    and $C$ is derived from the pooled sentence-pair embedding $\mathbf{S} \in \mathbb{R}^d$ via Equation (\ref{complexity_score}). 
    For inputs with higher complexity (larger $C$), the threshold $\rho_{dynamic}$ increases, thus fewer experts will be activated. This encourages the model to focus on the most specialized experts, thereby improving its ability to handle challenging samples.
    
    To determine which experts are selected, we compare the gate probability $P_i$ 
    (from Equation (\ref{probability_distribution})) against $\rho_{dynamic}$. 
    A binary mask $M_i$ is then defined as Equation (\ref{equa:M_i}).
    
    \begin{equation}
    M_i =
    \begin{cases}
    1, & \text{if } P_i > \rho_{dynamic},\\
    0, & \text{otherwise}.
    \end{cases}
    \label{equa:M_i}
    \end{equation}
    
    Here $M_i$ indicates whether expert $i$ is activated for a given input. 
    The mask is applied across all experts for each sample in the batch, forming a binary selection matrix. 
    Experts with $M_i=1$ are activated, while those with $M_i=0$ remain inactive.

    \item \textbf{Feed-Forward Networks:} Once the router determines the active experts, the pooled sentence-pair embedding $\mathbf{S} \in \mathbb{R}^d$ (defined in Equation (\ref{eqS})) is passed to the corresponding Feed-Forward Networks (FFNs). 
    Each FFN is specialized to capture certain aspects of inference, such as complex linguistic relations. 
    The $i$-th expert applies a Linear transformation, followed by Dropout for regularization, 
    and a ReLU activation to introduce non-linearity. 
    The output of the $i$-th expert is computed as Equation (\ref{FFN}).
    
    \begin{equation}
    \mathbf{FFN}_i(\mathbf{S}) = \text{ReLU}\big(\text{Dropout}(\mathbf{W}_{e_i}\mathbf{S} + \mathbf{b}_{e_i})\big)
    \label{FFN}
    \end{equation}
    
    where $\mathbf{S}$ is the pooled representation from the encoder, $\mathbf{W}_{e_i} \in \mathbb{R}^{d \times d}$ and $\mathbf{b}_{e_i} \in \mathbb{R}^{d}$ are the parameters of the $i$-th expert. The Dropout layer enhances generalization, while the ReLU activation ensures non-linear transformations.

    \item \textbf{Computation in MoE Block:} After computing the outputs of the individual Feed-Forward Networks (FFNs) for the active experts, the Mixture-of-Experts (MoE) block aggregates them to produce a unified representation. The pooled embedding $\mathbf{S} \in \mathbb{R}^d$ serves as the common input to all experts. 
    The final MoE output $O$ is a weighted sum of the expert outputs $\mathbf{FFN}_i(\mathbf{S})$, weighted by the corresponding gate probabilities $P_i$ and masked by $M_i$ as in Equation (\ref{out_moe}).
    
    \begin{equation}
    O = \sum_{i=1}^{num\_experts} M_i \cdot P_i \cdot \mathbf{FFN}_i(\mathbf{S})
    \label{out_moe}
    \end{equation}
    
    where $M_i$ is the binary mask from Equation (\ref{equa:M_i}), $P_i$ is the gate probability from Equation (\ref{probability_distribution}), 
    and $\mathbf{FFN}_i(\mathbf{S})$ is the output of the $i$-th expert defined in Equation (\ref{FFN}). To ensure numerical stability and proper normalization, the weighted sum is divided by the sum of the active gate probabilities, as in Equation (\ref{out_moe_norm}).
    
    \begin{equation}
    O = \frac{\sum_{i=1}^{num\_experts} M_i \cdot P_i \cdot \mathbf{FFN}_i(\mathbf{S})}{\sum_{i=1}^{num\_experts} M_i \cdot P_i}
    \label{out_moe_norm}
    \end{equation}
    
    This normalization ensures that the contribution of each active expert is appropriately scaled, allowing the MoE block to effectively integrate the specialized knowledge from the selected experts into a cohesive output for downstream processing.

\end{itemize}

\textbf{Classification Layer:} After obtaining the unified representation $O$ from the MoE block (Equation (\ref{out_moe_norm})), which is directly passed through a LayerNorm layer to stabilize training and enhance the robustness of the representation. This normalized output is then fed into the classification layer, which applies a linear transformation followed by a softmax activation function to predict the class probabilities for each premise-hypothesis pair. The classification layer computes the probabilities of the three labels: entailment, contradiction, and neutral. The entire process is expressed in Equation (\ref{probabilities}).

\begin{equation}
P(\text{class}) = \text{softmax}\left(\mathbf{W}_{cls} \cdot \text{LayerNorm}(O) + \mathbf{b}_{cls}\right)
\label{probabilities}
\end{equation}

where $O \in \mathbb{R}^{d}$ is the MoE block output for a given input sample, 
$\mathbf{W}_{cls} \in \mathbb{R}^{num\_labels \times d}$ and $\mathbf{b}_{cls} \in \mathbb{R}^{num\_labels}$ 
are the parameters of the classification layer, and $\text{LayerNorm}(O)$ denotes the layer-normalized representation.

\textbf{Loss Computation:} The loss function of the model is designed to optimize the classification performance while ensuring efficient and balanced expert selection in the Mixture-of-Experts (MoE) framework.
\begin{itemize}
    \item \textbf{Cross-Entropy Loss ($\text{Loss}_{ce}$):} This is the primary loss for the sequence classification task, measuring the difference between the predicted logits and the ground truth labels. It is calculated using the cross-entropy loss function to ensure the model accurately predicts class probabilities for premise-hypothesis pairs, as shown in Equation (\ref{CE}).

    \begin{equation}
    \text{Loss}_{\text{ce}} = \text{CrossEntropyLoss}(\text{logits}, \text{labels})
    \label{CE}
    \end{equation}
    
    Where $\text{logits}$ represents the raw output scores from the classification layer before applying softmax (Equation (\ref{probabilities})), and $\text{labels}$ denotes the ground truth labels for for each input pair.

\item \textbf{Dynamic Loss ($\text{Loss}_d$):} 
To encourage the model to select only the minimal necessary set of experts for each premise–hypothesis pair, an entropy-based dynamic loss is applied. This loss minimizes the entropy of the gate probability distribution, promoting a more focused selection of experts similar to the previous study \cite{huang-etal-2024-harder}, 
and is computed as in Equation (\ref{loss_d}).

\begin{equation}
\text{Loss}_d = -\frac{1}{num\_experts} \sum_{i=1}^{num\_experts} P_i \cdot \log(P_i)
\label{loss_d}
\end{equation}

where $P_i$ is the gate probability of the $i$-th expert, and $num\_experts$ denotes the total number of experts in the probability distribution for each input sample. In practice, this entropy is computed for every sample and then averaged across the batch. This penalty reduces routing uncertainty, encouraging the model to activate fewer experts for simpler inputs while still allowing multiple experts for complex cases.

    \item \textbf{Load Balance Loss ($\text{Loss}_b$):} To ensure an even distribution of workload across experts for each premise-hypothesis pair, a load balance loss is incorporated. This loss penalizes imbalances between the fraction of samples assigned to each expert and the average gate probabilities, which was applied in previous studies \citep{huang-etal-2024-harder, fedus2022switch, lepikhin2020gshard}, calculated as in Equation (\ref{loss_b}).

    \begin{equation}
    \text{Loss}_b = {num\_experts} \cdot \sum_{i=1}^{{num\_experts}} f_i \cdot Q_i
    \label{loss_b}
    \end{equation}
    
    where $f_i$ is the mean fraction of samples assigned to the $i$-th expert (computed as the mean of the expert mask across the batch), and $Q_i$ is the mean gate probability for the $i$-th expert across the batch. specifically, $f_i$ and $Q_i$ are calculated as in Equations (\ref{fi}) and (\ref{qi}).
    
    \begin{equation}
    f_i = \frac{1}{B} \sum_{j=1}^{B} M_{ji}
    \label{fi}
    \end{equation}
    \begin{equation}
    Q_i = \frac{1}{B} \sum_{j=1}^{B} P_{ji}
    \label{qi}
    \end{equation}

where $B$ is the batch size. 
Here, $M_{ji}$ corresponds to the binary expert mask introduced in Equation (\ref{equa:M_i}), indicating whether expert $i$ is activated for the $j$-th sample, and $P_{ji}$ denotes the gating probability defined in Equation (\ref{probability_distribution}) for expert $i$ given the $j$-th input.

    \item \textbf{Final Loss (Loss):} The final loss is a weighted combination of the cross-entropy loss, dynamic loss, and load balance loss (see Equation \ref{eq:totalloss}), allowing the model to balance classification accuracy with expert selection efficiency.
    \begin{equation}
    \text{Loss} = \text{Loss}_{\text{ce}} + \alpha \cdot \text{Loss}_d + \beta \cdot \text{Loss}_b
    \label{eq:totalloss}
    \end{equation}
    where $\alpha$ and $\beta$ are hyperparameters (set to $1e-3$ and $1e-2$ respectively in the code) that control the contribution of the dynamic and load balance losses. This composite loss ensures the model learns to classify accurately while optimizing the dynamic routing mechanism and maintaining load balance among experts.
    
    \end{itemize}

\section{Experiments and Results}
\label{sect:experiment}
In this section, we present the design of the main experiments to address the following questions: whether our dataset is more complex than previous datasets, whether it poses challenges to current models, and whether our proposed NLIMoE model outperforms other models. Detailed information regarding the datasets used, the selection and configuration of baseline models, and the evaluation metrics for performance measurement are also provided.

\subsection{Benchmark Datasets}
We combine the ViANLI dataset with other datasets to train machine learning models. These trained models are then used to evaluate our adversarial data, as well as several other datasets. This allows us to assess both the challenge posed by the adversarial data we have created and the effectiveness of using such data against attacks in the other datasets. The datasets combined with the ViANLI data include ViNLI \citep{huynh-etal-2022-vinli}, XNLI \citep{conneau2018xnli}, VLSP2021-NLI \citep{JCSCE}, and VnNewsNLI \citep{nguyen2022building}. Due to the varying nature and structure of these datasets, we reorganized and subdivided them to fit our experimental design. Specifically, XNLI and VLSP2021-NLI are multilingual inference datasets. Therefore, we extracted only Vietnamese samples for inclusion in the training data. Additionally, the authors of the VnNewsNLI and XNLI datasets developed only the development and test sets. Consequently, we used 90\% of the dev set for training and the remaining 10\% for model tuning. Furthermore, we allocated 10\% of the training set from VLSP2021-NLI for model tuning during the training process. Details of the sample sizes in each dataset used for the experiment are presented in Table \ref{tab:datasetInMainEx}.

\begin{table}[H]
\centering
\caption{Quantity Statistics of the Datasets in Our Main Experiment. ViANLI Dataset Is Aggregated Data from ViA1, ViA2, and ViA3.}
\label{tab:datasetInMainEx}
\begin{tabular}{lrrr}
\hline
\multicolumn{1}{c}{\multirow{2}{*}{\textbf{Dataset}}} & \multicolumn{3}{c}{\textbf{Quantity}} \\ \cline{2-4} 
\multicolumn{1}{c}{} & \multicolumn{1}{c}{\textbf{Train}} & \multicolumn{1}{c}{\textbf{Dev}} & \multicolumn{1}{c}{\textbf{Test}} \\ \hline
ViA1 & \multicolumn{1}{r}{2,609} & \multicolumn{1}{r}{330} & 330 \\ 
ViA2 & \multicolumn{1}{r}{2,672} & \multicolumn{1}{r}{330} & 330 \\ 
ViA3 & \multicolumn{1}{r}{2,731} & \multicolumn{1}{r}{340} & 340 \\ 
ViANLI & \multicolumn{1}{r}{8,012} & \multicolumn{1}{r}{1,000} & 1,000 \\ \hline 
ViNLI & \multicolumn{1}{r}{24,376} & \multicolumn{1}{r}{3,009} & 2,991 \\ 
XNLI & \multicolumn{1}{r}{6,750} & \multicolumn{1}{r}{750} & \multicolumn{1}{r}{-} \\ 
VLSP2021-NLI & \multicolumn{1}{r}{7,816} & \multicolumn{1}{r}{869} & 2,118 \\ 
VnNewsNLI & \multicolumn{1}{r}{18,221} & \multicolumn{1}{r}{2,025} & 11,866 \\ \hline
\end{tabular}
\end{table}

\subsection{Baseline Models}

In our experiments, we focus on using state-of-the-art transformer models, including mBERT \citep{devlin-etal-2019-bert}, XLM-R \citep{conneau-etal-2020-unsupervised}, CafeBERT \citep{do-etal-2024-vlue}, and PhoBERT \citep{nguyen2020phobert}. The mBERT, XLM-R, and CafeBERT models are powerful multilingual transformer models trained on large document corpora, while PhoBERT is a monolingual transformer model developed specifically for Vietnamese. Since Vietnamese uses both single words (one token) and compound words (multiple tokens), we use the VnCoreNLP tool \citep{vu2018vncorenlp} to tokenize the Vietnamese text, ensuring proper word segmentation for PhoBERT input.

Additionally, the hyperparameters selected for mBERT, XLM-R, CafeBERT, and PhoBERT during model training are designed to optimize performance for most models after evaluation. The basic parameters are as follows: max\_length=256, learning\_rate=1e-05, eval\_frequency=400, batch\_size=16, weight\_decay=0.0, adam\_epsilon=1e-08, dropout=0.4. Furthermore, we set epochs=7 since this yielded the best accuracy for the models. Reducing the number of epochs resulted in suboptimal performance, while increasing the epochs led to either saturation or a decline in model accuracy.

For our proposed NLIMoE model, in addition to the above hyperparameters, we introduce specific configurations for the Mixture-of-Experts (MoE) block. We set the number of experts to 7 and employ a dynamic routing mechanism to select an appropriate number of experts for each input. The dynamic threshold ($\rho_{static}$) is defined by combining a static threshold $\rho_{static} = 0.1$ with an input complexity score scaled by $\gamma = 0.1$. The coefficients of the additional loss terms are set to $\alpha$ = 1e-3 for the dynamic loss and $\beta$ = 1e-2 for the load balance loss, which provided the best trade-off between accuracy and expert utilization. This configuration enables NLIMoE to leverage conditional computation while maintaining robustness against adversarial inputs.

In addition, to compare NLIMoE with dynamic routing, we also implement a Top-K routing variant, where the top $K$ experts with the highest probabilities are selected. The values of $N$ (total experts) and $K$ (selected experts) are dataset-dependent and tuned separately for each case. Specifically, we set $N=5$ and $K=4$ for experiments on ViANLI, while $N=13$ and $K=9$ were used for ViNLI.

\subsection{Evaluation Metrics}

We employ both accuracy and F1-score (macro average) as evaluation metrics to assess the performance of the models, aligning with the approach used in the ViNLI evaluation \citep{huynh-etal-2022-vinli}. These metrics provide a comprehensive analysis of model effectiveness, with accuracy measuring the proportion of correctly predicted samples and F1-score offering a balanced assessment of precision and recall across label categories. The accuracy metric is formally defined in Equation (\ref{Accuracy}), while the F1-score is detailed in Equation (\ref{F1Score}).

\begin{equation}
\label{Accuracy}
    Accuracy = {C \over N}
\end{equation}

where \(N\) is the total number of samples evaluated, and \(C\) is the number of samples that the model correctly predicted.

\begin{equation}
\label{F1Score}
    F1_{\text{macro}} = \frac{1}{L} \sum_{i=1}^{L} F1_l
\end{equation}

where \( L \) is the total number of labels, and \( F1_l \) is the F1-Score for each label \( l \), calculated as in Equation (\ref{F1-label}).

\begin{equation}
\label{F1-label}
    F1_l = 2 \cdot \frac{\text{Precision}_l \cdot \text{Recall}_l}{\text{Precision}_l + \text{Recall}_l}
\end{equation}

where \( \text{Precision}_l = \frac{TP_l}{TP_l + FP_l} \) is the precision for label \( l \) (with \( TP_l \) as true positives and \( FP_l \) as false positives), and \( \text{Recall}_l = \frac{TP_l}{TP_l + FN_l} \) is the recall for label \( l \) (with \( FN_l \) as false negatives).

\subsection{Experimental Results} 
\label{sect:main_result}

\textbf{NLIMoE$_{Dynamic}$ model performance is better than baseline models, and ViANLI dataset is more challenging than previous dataset.} Table \ref{tab:mainresult} presents the results of our main experiment. The objective is to compare and evaluate the performance of our proposed NLIMoE$_{Dynamic}$ model against baseline models such as mBERT, PhoBERT, CafeBERT, and NLIMoE$_{TopK}$ on the ViANLI dataset, which we developed, as well as on the ViNLI dataset. This comparison highlights the challenges posed by our dataset compared to previous datasets and demonstrates the versatility of our model across adversarial and traditionally constructed datasets like ViNLI.

In this experiment, we evaluated two different input representations for the model: word-based and syllable-based. For the word-based model, we used PhoBERT, while the other models were based on syllable-based representations. To preprocess the Vietnamese text for the PhoBERT model, we used the VnCoreNLP tool \citep{vu2018vncorenlp} for word segmentation.

The results in Table \ref{tab:mainresult} indicate that the ViANLI dataset presents more challenges to the models compared to ViNLI, as the accuracy of all models on the ViANLI dataset is significantly lower than on ViNLI. Notably, when comparing our model with the others, on the development set of ViANLI, CafeBERT achieved 48.2\%, NLIMoE$_{TopK}$ achieved 48.3\%, which is lower than the 49.2\% of our model. On the test set, this discrepancy becomes even more pronounced, with our model reaching 47.3\%, nearly 2\% higher than the highest-performing baseline model, XLM-R$_{Large}$.

In addition, the NLIMoE$_{Dynamic}$ model not only performs better on the ViANLI dataset but also outperforms other models on ViNLI. On the development set, NLIMoE$_{Dynamic}$ achieved 83.59\%, surpassing XLM-R$_{Large}$ and CafeBERT by 0.57\% and 1.42\%, respectively. On the test set, NLIMoE$_{Dynamic}$ achieved 82.82\%, outperforming XLM-R$_{Large}$ by 1.46\%. In general, when comparing the performance of the same model on both datasets, the ViANLI dataset proves to be more challenging for the models to predict semantic relationships accurately compared to ViNLI. Our model has shown greater effectiveness in handling the more complex cases presented by our dataset.

\begin{table}[H]
\renewcommand{\arraystretch}{1.5}
\centering
\caption{Model Performance on Two Benchmark Datasets including ViANLI and ViNLI. Results marked with \textsuperscript{*} are extracted from the study of \cite{huynh-etal-2022-vinli}. Dif$_{Acc}$ is the Abbreviation for the Accuracy Difference Between the ViNLI and ViANLI Datasets.}
\label{tab:mainresult}
\resizebox{\columnwidth}{!}{%
\begin{tabular}{llrrrrr|rrrrr}
\hline
\multicolumn{2}{c}{\multirow{3}{*}{\textbf{Models}}}                   & \multicolumn{5}{c|}{\textbf{Dev}}                                                                                                                                                                               & \multicolumn{5}{c}{\textbf{Test}}                                                                                                                                                                              \\ \cline{3-12} 
\multicolumn{2}{c}{}                                                   & \multicolumn{2}{c}{\textbf{ViNLI}}                                        & \multicolumn{2}{c}{\textbf{ViANLI}}                                       & \multicolumn{1}{c|}{\multirow{2}{*}{\textbf{Dif$_{Acc}$}}} & \multicolumn{2}{c}{\textbf{ViNLI}}                                        & \multicolumn{2}{c}{\textbf{ViANLI}}                                       & \multicolumn{1}{c}{\multirow{2}{*}{\textbf{Dif$_{Acc}$}}} \\ \cline{3-6} \cline{8-11}
\multicolumn{2}{c}{}                                                   & \multicolumn{1}{c}{\textbf{Acc}} & \multicolumn{1}{c}{\textbf{F1-score}} & \multicolumn{1}{c}{\textbf{Acc}} & \multicolumn{1}{c}{\textbf{F1-score}} & \multicolumn{1}{c|}{}                                 & \multicolumn{1}{c}{\textbf{Acc}} & \multicolumn{1}{c}{\textbf{F1-score}} & \multicolumn{1}{c}{\textbf{Acc}} & \multicolumn{1}{c}{\textbf{F1-score}} & \multicolumn{1}{c}{}                                 \\ \hline
\multicolumn{1}{l}{\multirow{2}{*}{Word}}     & PhoBERT$_{Base}$            & \multicolumn{1}{r}{75.07*}       & \multicolumn{1}{r}{75.08*}            & \multicolumn{1}{r}{45.20}        & \multicolumn{1}{r}{43.12}             & 29.87                                                 & \multicolumn{1}{r}{72.87*}       & \multicolumn{1}{r}{72.79*}            & \multicolumn{1}{r}{44.30}        & \multicolumn{1}{r}{41.92}             & 28.57                                                 \\ 
\multicolumn{1}{l}{}                          & PhoBERT$_{Large}$           & \multicolumn{1}{r}{77.33*}       & \multicolumn{1}{r}{77.34*}            & \multicolumn{1}{r}{46.80}        & \multicolumn{1}{r}{44.27}             & 30.63                                                 & \multicolumn{1}{r}{75.93*}       & \multicolumn{1}{r}{75.87*}            & \multicolumn{1}{r}{44.90}        & \multicolumn{1}{r}{41.51}             & 31.03                                                 \\ \hline
\multicolumn{1}{l}{\multirow{6}{*}{Syllable}} & mBERT                  & \multicolumn{1}{r}{67.41*}       & \multicolumn{1}{r}{67.46*}            & \multicolumn{1}{r}{44.00}        & \multicolumn{1}{r}{41.99}             & 23.41                                                 & \multicolumn{1}{r}{64.84*}       & \multicolumn{1}{r}{64.83*}            & \multicolumn{1}{r}{42.50}        & \multicolumn{1}{r}{39.41}             & 22.34                                                 \\  
\multicolumn{1}{l}{}                          & XLM-R$_{Base}$              & \multicolumn{1}{r}{72.02*}       & \multicolumn{1}{r}{71.99*}            & \multicolumn{1}{r}{46.90}        & \multicolumn{1}{r}{42.54}             & 25.12                                                 & \multicolumn{1}{r}{71.59*}       & \multicolumn{1}{r}{71.51*}            & \multicolumn{1}{r}{44.80}        & \multicolumn{1}{r}{39.88}             & 26.79                                                 \\ 
\multicolumn{1}{l}{}                          & XLM-R$_{Large}$             & \multicolumn{1}{r}{83.02*}       & \multicolumn{1}{r}{82.98*}            & \multicolumn{1}{r}{47.00}        & \multicolumn{1}{r}{45.75}             & 36.02                                                 & \multicolumn{1}{r}{81.36*}       & \multicolumn{1}{r}{81.31*}            & \multicolumn{1}{r}{45.50}        & \multicolumn{1}{r}{44.18}             & 35.86                                                 \\ 
\multicolumn{1}{l}{}                          & CafeBERT               & \multicolumn{1}{r}{82.17}        & \multicolumn{1}{r}{82.15}             & \multicolumn{1}{r}{48.20}        & \multicolumn{1}{r}{\textbf{47.93}}             & 33.97                                                 & \multicolumn{1}{r}{82.51}        & \multicolumn{1}{r}{82.49}             & \multicolumn{1}{r}{45.10}        & \multicolumn{1}{r}{44.98}             & 37.41                                                 \\ \cline{2-12} 
\multicolumn{1}{l}{}                          & NLIMoE$_{TopK}$             & \multicolumn{1}{r}{83.54}             & \multicolumn{1}{r}{83.54}                  & \multicolumn{1}{r}{48.30}             & \multicolumn{1}{r}{47.80}                  &  35.24                                                     & \multicolumn{1}{r}{82.55}             & \multicolumn{1}{r}{82.56}                  & \multicolumn{1}{r}{46.70}             & \multicolumn{1}{r}{\textbf{46.10}}                  &     35.85                                                  \\  
\multicolumn{1}{l}{}                          & \textbf{NLIMoE$_{Dynamic}$} & \multicolumn{1}{r}{\textbf{83.59}}    & \multicolumn{1}{r}{\textbf{83.60}}         & \multicolumn{1}{r}{\textbf{49.20}}    & \multicolumn{1}{r}{47.24}         & 34.39                                             & \multicolumn{1}{r}{\textbf{82.82}}    & \multicolumn{1}{r}{\textbf{82.83}}         & \multicolumn{1}{r}{\textbf{47.30}}    & \multicolumn{1}{r}{44.92}         & 35.52                                             \\ \hline
\end{tabular}}
\end{table}

\vspace{9pt}

\textbf{Continuously augmenting training data with ViANLI improves model performance on other datasets.} Table \ref{tab:result1} (in \ref{ModelAccuracyonOtherTestSets}) presents the results of another experiment we conducted to evaluate the ability of combining ViANLI with other datasets to enhance reasoning capabilities and recognize semantic relationships across datasets during the training process. The experimental results reveal many interesting findings about the performance of the models and the characteristics of the adversarial data we have created.

The models selected for evaluation are also the ones involved in the data generation process, including mBERT, PhoBERT, XLM-R, and our proposed model NLIMoE$_{Dynamic}$. As we gradually increased the training data by sequentially adding ViA1, ViA2, and ViA3, the performance of the models improved incrementally with each addition. This indicates that each additional round of adversarial data made the models more robust and better equipped to handle data attacks, especially on the test sets of ViANLI and VLSP2021-NLI.

%Kiet: Chưa rõ ý chính muốn nói gì trong đoạn này.
Moreover, when combining the entire ViANLI dataset with all other datasets (ViNLI, XNLI, VLSP2021-NLI, VnNewsNLI) to create a large training set, models including mBERT, PhoBERT, XLM-R, and NLIMoE$_{Dynamic}$ showed significant performance improvements on datasets such as VLSP2021-NLI and VnNewsNLI. However, this combination inadvertently introduced noise, which reduced the effectiveness of the models when evaluated on the ViANLI dataset. Overall, ViANLI still demonstrates that incorporating this dataset into the training process helps the model learn more complex semantic relationships, thereby improving its ability to handle and reason over other datasets. However, a model that achieves high accuracy on these datasets may still fail when evaluated on ViANLI.

% \begin{table}[]
% \renewcommand{\arraystretch}{1.5}
% \centering
% \resizebox{\columnwidth}{!}{%
% \begin{tabular}{llllllllll}
% \hline
% \multicolumn{1}{c}{\multirow{2}{*}{\textbf{Model}}} & \multicolumn{1}{c}{\multirow{2}{*}{\textbf{Training data}}} & \multicolumn{2}{c}{\textbf{ViA1}} & \multicolumn{2}{c}{\textbf{ViA2}} & \multicolumn{2}{c}{\textbf{ViA3}} & \multicolumn{2}{c}{\textbf{ViANLI}} \\ \cline{3-10} 
% \multicolumn{1}{c}{} & \multicolumn{1}{c}{} & \multicolumn{1}{c}{\textbf{Dev}} & \multicolumn{1}{c}{\textbf{Test}} & \multicolumn{1}{c}{\textbf{Dev}} & \multicolumn{1}{c}{\textbf{Test}} & \multicolumn{1}{c}{\textbf{Dev}} & \multicolumn{1}{c}{\textbf{Test}} & \multicolumn{1}{c}{\textbf{Dev}} & \multicolumn{1}{c}{\textbf{Test}} \\ \hline
% \multirow{3}{*}{NLIMoE} & ViA1 & \multicolumn{1}{l}{47.27} & 45.15 & \multicolumn{1}{l}{40.00} & 42.12 & \multicolumn{1}{l}{39.17} & 40.00 & \multicolumn{1}{l}{44.30} & 42.10 \\ 
%  & ViA1+ViA2 & \multicolumn{1}{l}{49.09} & 49.39 & \multicolumn{1}{l}{45.75} & 47.78 & \multicolumn{1}{l}{43.23} & 45.00 & \multicolumn{1}{l}{46.00} & 43.90 \\ 
%  & ViA1+ViA2+ViA3 & \multicolumn{1}{l}{53.93} & 50.60 & \multicolumn{1}{l}{47.57} & 48.78 & \multicolumn{1}{l}{43.52} & 43.23 & \multicolumn{1}{l}{48.70} & 47.30 \\ \hline
% \end{tabular}%
% }
% \caption{Accuracy (\%) of the Proposed NLIMoE Model on Round-by-round Data, Trained with Incremental Data.}
% \label{tab:result2}
% \end{table}

\vspace{9pt}

\begin{figure}[]
    \centering
    \begin{subfigure}{0.45\textwidth}
        \includegraphics[width=\textwidth]{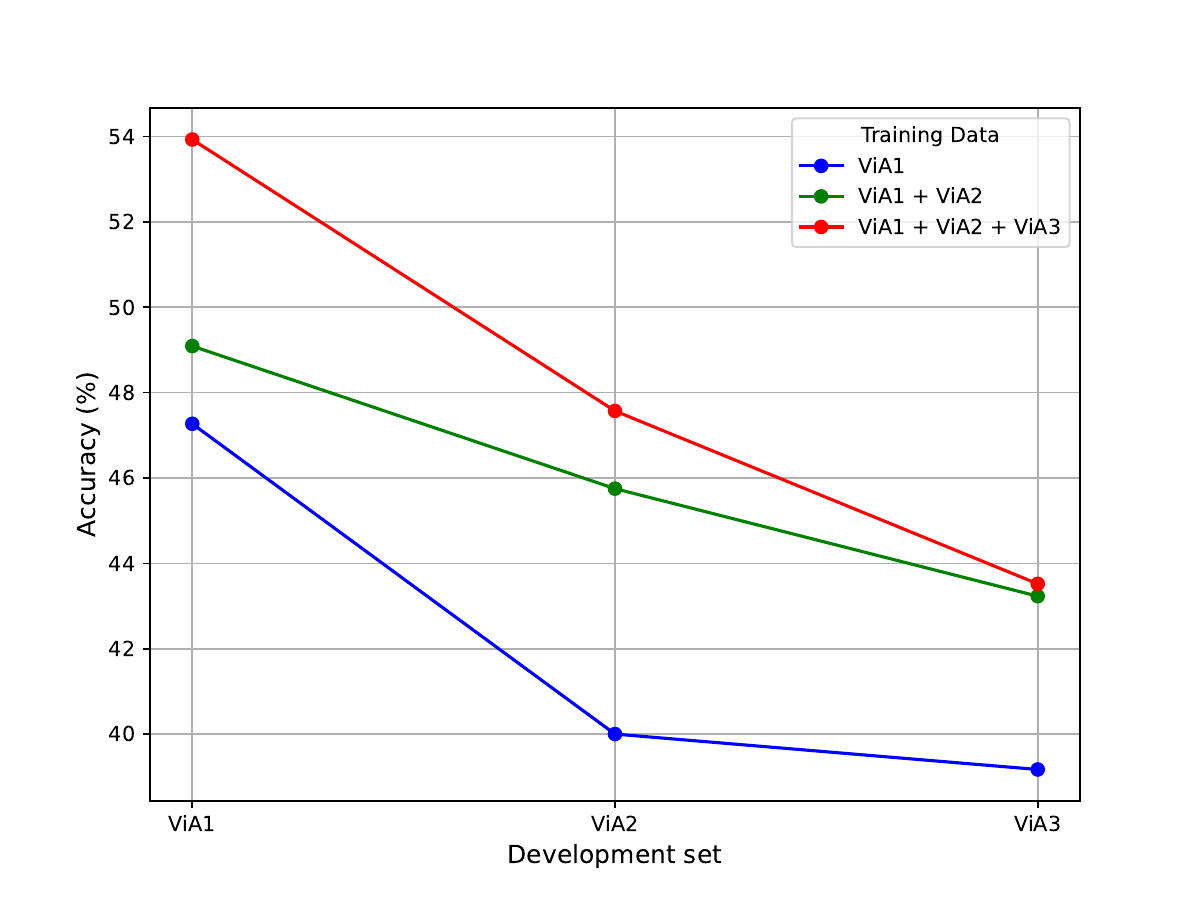}
        \caption{Dev}
        \label{Dev}
    \end{subfigure}
    \begin{subfigure}{0.45\textwidth}
        \includegraphics[width=\textwidth]{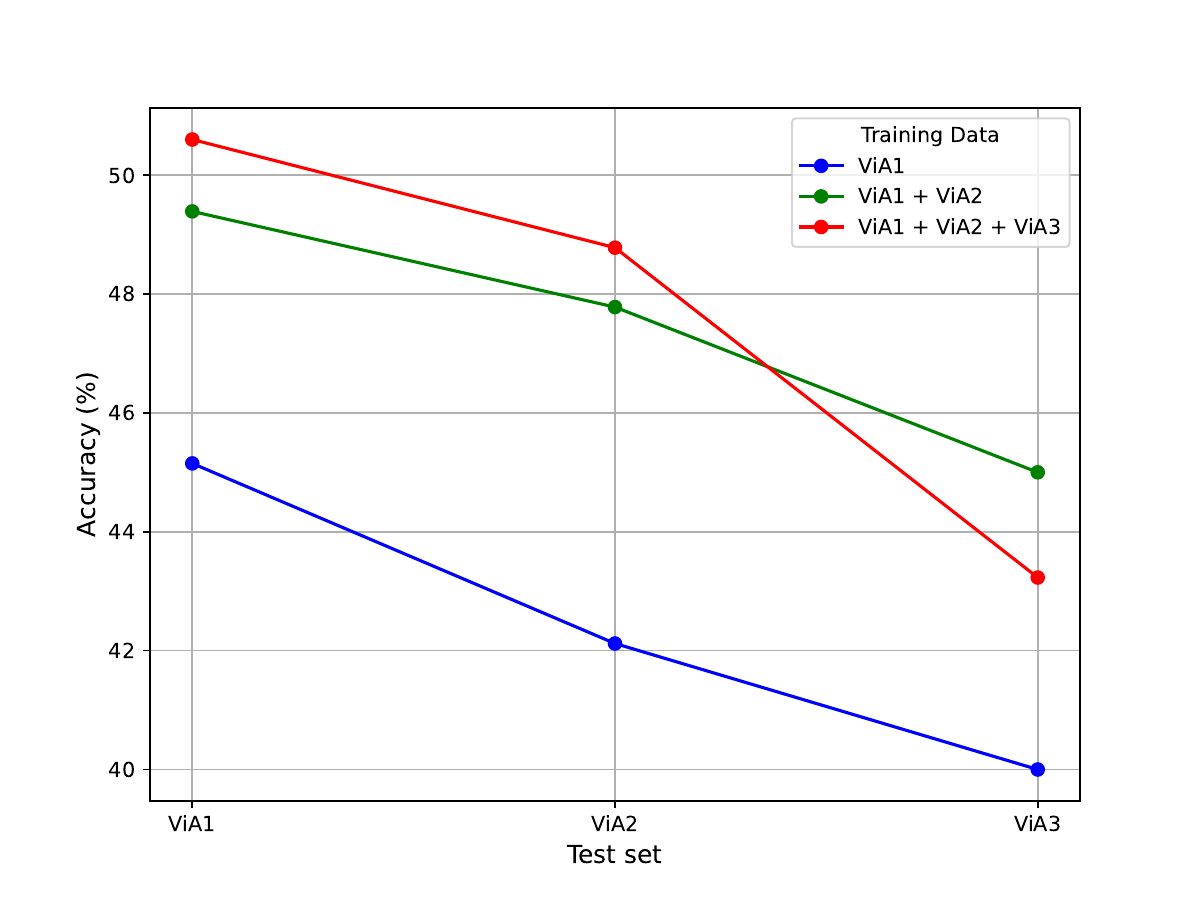}
        \caption{Test}
        \label{Test}
    \end{subfigure}
    \caption{Accuracy (\%) of the Proposed NLIMoE Model on the dev and test sets of ViANLI in each round (ViA1, ViA2, ViA3). NLIMoE$_{Dynamic}$ Model is Trained with Incremental Training Data.}
    \label{fig:acc-r-by-r}
\end{figure}

\textbf{Data in rounds become increasingly more challenging.} We train NLIMoE$_{Dynamic}$ on an increasing amount of data from different rounds and evaluate the performance of the model on the data from each round, as well as on the entire ViANLI dataset. This experiment, presented in Figure \ref{fig:acc-r-by-r}, helps us specifically assess the impact of the amount of training data on the predictive power of our model and the challenge of the data built in each round.

The results of this experiment show that as the model is trained on an increasing amount of adversarial data, the performance of the NLIMoE$_{Dynamic}$ model strengthens, with accuracy generally improving on the dev and test sets of ViA1, ViA2, ViA3, and even ViANLI. This indicates that as the model is exposed to more adversarial data, it improves its ability to handle attacks from such data.Besides, we observe that for all three training cases of the NLIMoE$_{Dynamic}$ model, the accuracy of the model gradually decreases on both the Dev and Test sets through each round. This reinforces the fact that the difficulty of the data is increasing. If this data-building process continues for future rounds, it generates increasingly high-quality and challenging data. This is of great significance in maintaining the challenge and attractiveness of the NLI task, thereby encouraging more future research rather than relying on a static data-building process.

\section{Results Analysis and Discussion}
\label{sect:result_analysis}
% In this section, we present a comprehensive analysis of the experimental results to highlight characteristics ViANLI dataset and the performance of the proposed NLIMoE$_{Dynamic}$ model compared to other NLI models. Our evaluation focuses on several key aspects of model design and data complexity, addressing critical questions including how the dynamic routing mechanism with dynamic thresholding that we proposed is more effective than the static thresholding mechanism and also the previous TopK routing mechanism? How sensitive is NLIMoE$_{Dynamic}$ performance to dropout and learning rate variations? Does sentence length expose weaknesses of NLI models, and how does NLIMoE stability compare to other models on ViANLI? Does word overlap between premise and hypothesis simplify NLI tasks or highlight NLIMoE$_{Dynamic}$ advantages on ViANLI? Does the new word rate challenge NLI models, and how does NLIMoE perform in such scenarios? Do annotator artifacts in hypotheses make ViANLI less challenging than other datasets? In particular, we evaluate the performance of NLIMoE$_{Dynamic}$ according to the linguistic features that appear in ViANLI through different types of inference. Finally, we analyze error patterns in model predictions to identify areas for future improvement, providing insights into enhancing NLIMoE robustness and performance on complex Vietnamese NLI data.

In this section, we present a comprehensive analysis of the experimental results to highlight the characteristics of the ViANLI dataset and evaluate the performance of the proposed NLIMoE$_{Dynamic}$ model in comparison with other NLI baselines. Our evaluation addresses several key aspects of model design and data complexity. Specifically, we examine whether the proposed dynamic routing with a dynamic threshold outperforms both static thresholding and traditional Top-K routing. We also investigate the sensitivity of NLIMoE$_{Dynamic}$ to ablation studies, and assess the impact of sentence length on model stability relative to other NLI models. Furthermore, we analyze whether word overlap between premise and hypothesis simplifies the NLI task or highlights the advantages of NLIMoE$_{Dynamic}$, and whether a high rate of new words presents challenges that NLIMoE can better handle. We also consider the influence of potential annotator artifacts in hypotheses on dataset difficulty. In particular, we evaluate the effectiveness of NLIMoE$_{Dynamic}$ across diverse linguistic features through inference types present in ViANLI. Finally, we analyze model error patterns to identify areas for improvement, providing insights into enhancing robustness and performance on complex Vietnamese NLI tasks.

\subsection{Ablation Studies}

To evaluate the effectiveness of each component in our proposed NLIMoE model, we conduct a comprehensive set of ablation studies on ViANLI and ViNLI. The primary objective of these experiments is to quantify the contribution of the auxiliary loss terms—namely the dynamic loss ($\alpha$) and the load-balance loss ($\beta$)—to the overall model performance. We further investigate the impact of the routing regularization coefficient ($\gamma$), which interacts with input complexity to determine the dynamic selection threshold for experts. In addition, we examine the role of dropout regularization in controlling overfitting within the expert networks, and assess the direct impact of the Mixture-of-Experts (MoE) block itself compared to a standard transformer encoder.

We systematically vary the weighting coefficients of the auxiliary losses, including $\alpha$ for the dynamic loss, $\beta$ for the load-balance loss, and $\gamma$ for the dynamic router mechanism. We also adjust the dropout rate within the expert networks and compare model variants trained with and without the MoE block. Accuracy on both datasets is showed in Table~\ref{tab:ablation-test}.

For the dynamic loss coefficient $\alpha$, the best performance is obtained at $\alpha=1e{-}3$ (49.20\% on ViANLI and 83.59\% on ViNLI). Both smaller and larger values reduce accuracy, indicating that a moderate level of entropy regularization is crucial for stable and diverse expert routing. The load-balance loss coefficient $\beta$ achieves the best results at $\beta=1e{-}2$, which balances the workload across experts while still allowing sufficient specialization. Increasing or decreasing $\beta$ away from this value leads to load imbalance and performance drops. For the routing regularization coefficient $\gamma$, moderate values with $1e{-}1$ produce the most consistent improvements by dynamically adjusting the threshold to activate an appropriate number of experts for each input. In contrast, higher values of $\gamma$ raise the threshold excessively, resulting in fewer selected experts and reduced flexibility, which significantly degrades performance.

Dropout also plays a critical role, especially when a rate of 0.4 consistently yields the strongest results across both datasets, effectively balancing overfitting and underfitting within the expert networks. Furthermore, the comparison between variants with and without the MoE block highlights the substantial contribution of the expert mechanism itself. Incorporating MoE leads to clear performance gains, with improvements of +1.4\% on ViANLI and +1.1\% on ViNLI. These results confirm that dynamic expert routing introduces complementary modeling capacity beyond the base transformer encoder, enabling the model to specialize in diverse inference patterns and handle adversarially challenging inputs more effectively.

\begin{table}[H]
\renewcommand{\arraystretch}{1.2}
\centering
\caption{Ablation study results of the proposed NLIMoE model on ViANLI and ViNLI datasets across different hyperparameters (Alpha - $\alpha$, Beta - $\beta$, Gamma - $\gamma$, Dropout) and model configurations (with/without MoE).}
\label{tab:ablation-test}
\begin{tabular}{lcccccc}
\hline
\multicolumn{1}{c}{\textbf{Dataset}} &
  \multicolumn{6}{c}{\textbf{Ablation Studies}} \\ \hline
\multicolumn{7}{c}{\textbf{Alpha ($\alpha$)}} \\ \hline
\multicolumn{1}{l}{} &
  \multicolumn{1}{c}{\textbf{w/o}} &
  \multicolumn{1}{c}{\textbf{5e-4}} &
  \multicolumn{1}{c}{\textbf{1e-3}} &
  \multicolumn{1}{c}{\textbf{2e-3}} &
  \multicolumn{1}{c}{\textbf{5e-3}} &
  \textbf{1e-2} \\ \hline
\multicolumn{1}{l}{ViANLI} &
  \multicolumn{1}{c}{46.80} &
  \multicolumn{1}{c}{46.90} &
  \multicolumn{1}{c}{\textbf{49.20}} &
  \multicolumn{1}{c}{47.30} &
  \multicolumn{1}{c}{46.40} &
  47.00 \\ 
\multicolumn{1}{l}{ViNLI} &
  \multicolumn{1}{c}{82.61} &
  \multicolumn{1}{c}{83.32} &
  \multicolumn{1}{c}{\textbf{83.59}} &
  \multicolumn{1}{c}{82.66} &
  \multicolumn{1}{c}{82.44} &
  82.31 \\ \hline
\multicolumn{7}{c}{\textbf{Beta ($\beta$)}} \\ \hline
\multicolumn{1}{l}{} &
  \multicolumn{1}{c}{\textbf{w/o}} &
  \multicolumn{1}{c}{\textbf{2e-3}} &
  \multicolumn{1}{c}{\textbf{5e-3}} &
  \multicolumn{1}{c}{\textbf{1e-2}} &
  \multicolumn{1}{c}{\textbf{2e-2}} &
  \textbf{5e-2} \\ \hline
\multicolumn{1}{l}{ViANLI} &
  \multicolumn{1}{c}{44.40} &
  \multicolumn{1}{c}{46.90} &
  \multicolumn{1}{c}{45.60} &
  \multicolumn{1}{c}{\textbf{49.20}} &
  \multicolumn{1}{c}{46.10} &
  40.60 \\ 
\multicolumn{1}{l}{ViNLI} &
  \multicolumn{1}{c}{82.00} &
  \multicolumn{1}{c}{82.83} &
  \multicolumn{1}{l}{83.14} &
  \multicolumn{1}{c}{\textbf{83.59}} &
  \multicolumn{1}{l}{82.88} &
  \multicolumn{1}{l}{82.17} \\ \hline
\multicolumn{7}{c}{\textbf{Gamma ($\gamma$)}} \\ \hline
\multicolumn{1}{l}{} &
  \multicolumn{1}{l}{\textbf{w/o}} &
  \multicolumn{1}{r}{\textbf{1e-2}} &
  \multicolumn{1}{r}{\textbf{5e-2}} &
  \multicolumn{1}{r}{\textbf{1e-1}} &
  \multicolumn{1}{r}{\textbf{1.5e-1}} &
  \multicolumn{1}{r}{\textbf{2e-1}} \\ \hline
\multicolumn{1}{l}{ViANLI} &
  \multicolumn{1}{c}{47.10} &
  \multicolumn{1}{c}{47.70} &
  \multicolumn{1}{c}{47.10} &
  \multicolumn{1}{c}{\textbf{49.20}} &
  \multicolumn{1}{c}{46.70} &
  44.5 \\ 
\multicolumn{1}{l}{ViNLI} &
  \multicolumn{1}{l}{82.26} &
  \multicolumn{1}{l}{82.48} &
  \multicolumn{1}{l}{83.23} &
  \multicolumn{1}{c}{\textbf{83.59}} &
  \multicolumn{1}{l}{81.59} &
  \multicolumn{1}{l}{80.33} \\ \hline
\multicolumn{7}{c}{\textbf{Dropout}} \\ \hline
\multicolumn{1}{l}{} &
  \multicolumn{1}{c}{\textbf{w/o}} &
  \multicolumn{1}{c}{\textbf{0.1}} &
  \multicolumn{1}{c}{\textbf{0.2}} &
  \multicolumn{1}{c}{\textbf{0.3}} &
  \multicolumn{1}{c}{\textbf{0.4}} &
  \textbf{0.5} \\ \hline
\multicolumn{1}{l}{ViANLI} &
  \multicolumn{1}{c}{46.90} &
  \multicolumn{1}{c}{48.10} &
  \multicolumn{1}{c}{47.10} &
  \multicolumn{1}{c}{47.80} &
  \multicolumn{1}{c}{\textbf{49.20}} &
  47.60 \\ 
\multicolumn{1}{l}{ViNLI} &
  \multicolumn{1}{c}{82.66} &
  \multicolumn{1}{c}{82.88} &
  \multicolumn{1}{c}{83.10} &
  \multicolumn{1}{c}{83.06} &
  \multicolumn{1}{c}{\textbf{83.59}} &
  83.14 \\ \hline
\multicolumn{7}{c}{\textbf{Mixture-of-Experts Block}} \\ \hline
\multicolumn{1}{l}{} &
  \multicolumn{3}{c}{\textbf{w/o MoE}} &
  \multicolumn{3}{c}{\textbf{MoE}} \\ \hline
\multicolumn{1}{l}{ViANLI} &
  \multicolumn{3}{c}{47.80} &
  \multicolumn{3}{c}{\textbf{49.20 (\textcolor{blue}{$\uparrow$1.40})}} \\ 
\multicolumn{1}{l}{ViNLI} &
  \multicolumn{3}{c}{82.48} &
  \multicolumn{3}{c}{\textbf{83.59 (\textcolor{blue}{$\uparrow$1.11})}} \\ \hline
\end{tabular}
\end{table}

\subsection{Comparing NLIMoE with Dynamic and Top-K Routing}

We conducted comprehensive experiments to evaluate the effectiveness of our proposed NLIMoE with dynamic routing compared to Top-K routing. The design of expert quantity and Top-K selection are crucial factors in optimizing Mixture-of-Experts models, as they directly affect the ability of the model to specialize and capture complex inference relations. To explore the best configuration, we varied both the number of experts (N) and the Top-K value on the ViANLI and ViNLI datasets, and systematically compared the results with our dynamic routing approach. In addition, we explicitly included Top-1 and Top-2 routing, which correspond to the strategies used in Switch Transformer \cite{fedus2022switch} and GShard \cite{lepikhin2020gshard}, respectively, to contextualize our method against prior advances in MoE architectures.

Table \ref{tab:topk_dynamic} presents the performance of the models on both the development and test sets of ViANLI and ViNLI. On the ViANLI dataset, the optimal configuration was obtained with 5 experts and Top-K = 4, achieving 48.30\% on the development set and 46.70\% on the test set. Reducing the number of experts to 3 or increasing it to 7, 9, 11, or 13 resulted in suboptimal accuracy. This suggests that ViANLI benefits from a moderate number of experts combined with a carefully chosen Top-K value. In contrast, for the ViNLI dataset, the best results with Top-K routing were achieved with 13 experts and Top-K = 9, yielding 83.54\% on the development set and 82.55\% on the test set. Lower expert counts, such as 11, 9, 7, 5, or 3, consistently led to reduced performance. This indicates that ViNLI requires a larger number of experts and a higher Top-K value to fully leverage the advantages of MoE routing.

Another important observation is that sparse Top-K routing generally outperformed dense routing, where all experts are activated simultaneously. This finding demonstrates the importance of careful and selective expert activation rather than relying on full expert utilization, which can lead to computational inefficiency without corresponding performance gains.

When compared with our NLIMoE using dynamic routing, the proposed model consistently outperformed nearly all Top-K routing configurations on both ViANLI and ViNLI. In particular, our approach achieved higher accuracy than Top-1 and Top-2 routing, which replicate the strategies of Switch Transformer and GShard. With 7 experts, NLIMoE with dynamic routing demonstrated superior effectiveness, showing a stronger ability to capture diverse linguistic inference patterns. These results confirm that our proposed dynamic routing mechanism provides more robust and adaptive expert utilization compared to recent MoE architectures, thereby establishing the innovation and contribution of NLIMoE in advancing Mixture-of-Experts approaches for natural language inference.

\begin{table}[H]
\renewcommand{\arraystretch}{1.1}
\centering
\caption{Impact of Expert Quantity (\#Expert) and Top-k Expert Selection of Top-K Routing on Model Accuracy and Computational Efficiency on ViANLI and ViNLI Datasets compared to Dynamic Routing. Top-1\textbf{\textsuperscript{†}} corresponds to the approach of Switch Transformer \cite{fedus2022switch}, while Top-2\textbf{\bf\textsuperscript{§}} corresponds to the approach of GShard \cite{lepikhin2020gshard}.}
\label{tab:topk_dynamic}
\begin{tabular}{lccccccc}
\hline
\multicolumn{4}{c}{\textbf{NLIMoE}} &
  \multicolumn{2}{c}{\textbf{ViANLI}} &
  \multicolumn{2}{c}{\textbf{ViNLI}} \\ \hline
\multicolumn{1}{c}{\textbf{Routing Method}} &
  \multicolumn{1}{c}{\textbf{\#Expert}} &
  \multicolumn{2}{c}{\textbf{\#TopK}} &
  \multicolumn{1}{c}{\textbf{Dev}} &
  \textbf{Test} &
  \multicolumn{1}{c}{\textbf{Dev}} &
  \textbf{Test} \\ \hline
\multicolumn{1}{l}{\multirow{34}{*}{\textbf{Top-K Routing}}} &
  \multicolumn{1}{c}{\multirow{3}{*}{3}} &
  \multicolumn{1}{c}{\multirow{2}{*}{Sparse}} &
  1\bf\textsuperscript{†} &
  \multicolumn{1}{c}{36.60} &
  35.20 &
  \multicolumn{1}{c}{83.05} &
  82.26 \\ 
\multicolumn{1}{l}{} &
  \multicolumn{1}{c}{} &
  \multicolumn{1}{c}{} &
  2\bf\textsuperscript{§} &
  \multicolumn{1}{c}{48.20} &
  45.40 &
  \multicolumn{1}{c}{83.28} &
  82.38 \\ \cline{3-8} 
\multicolumn{1}{l}{} &
  \multicolumn{1}{c}{} &
  \multicolumn{1}{c}{Dense} &
  3 &
  \multicolumn{1}{c}{47.40} &
  44.70 &
  \multicolumn{1}{c}{83.19} &
  82.46 \\ \cline{2-8} 
\multicolumn{1}{l}{} &
  \multicolumn{1}{c}{\multirow{5}{*}{5}} &
  \multicolumn{1}{c}{\multirow{4}{*}{Sparse}} &
  1\bf\textsuperscript{†} &
  \multicolumn{1}{c}{48.20} &
  46.70 &
  \multicolumn{1}{c}{82.79} &
  81.71 \\ 
\multicolumn{1}{l}{} &
  \multicolumn{1}{c}{} &
  \multicolumn{1}{c}{} &
  2\bf\textsuperscript{§} &
  \multicolumn{1}{c}{47.70} &
  45.30 &
  \multicolumn{1}{c}{82.97} &
  82.28 \\ 
\multicolumn{1}{l}{} &
  \multicolumn{1}{c}{} &
  \multicolumn{1}{c}{} &
  3 &
  \multicolumn{1}{c}{47.00} &
  46.30 &
  \multicolumn{1}{c}{83.06} &
  81.97 \\ 
\multicolumn{1}{l}{} &
  \multicolumn{1}{c}{} &
  \multicolumn{1}{c}{} &
  4 &
  \multicolumn{1}{c}{\textbf{48.30}} &
  \textbf{46.70} &
  \multicolumn{1}{c}{82.57} &
  82.03 \\ \cline{3-8} 
\multicolumn{1}{l}{} &
  \multicolumn{1}{c}{} &
  \multicolumn{1}{c}{Dense} &
  5 &
  \multicolumn{1}{c}{48.20} &
  44.40 &
  \multicolumn{1}{c}{83.15} &
  81.97 \\ \cline{2-8} 
\multicolumn{1}{l}{} &
  \multicolumn{1}{c}{\multirow{5}{*}{7}} &
  \multicolumn{1}{c}{\multirow{4}{*}{Sparse}} &
  1\bf\textsuperscript{†} &
  \multicolumn{1}{c}{47.80} &
  46.10 &
  \multicolumn{1}{c}{82.53} &
  82.11 \\ 
\multicolumn{1}{l}{} &
  \multicolumn{1}{c}{} &
  \multicolumn{1}{c}{} &
  2\bf\textsuperscript{§} &
  \multicolumn{1}{c}{47.30} & 47.40
   &
  \multicolumn{1}{c}{83.23} & 82.06
   \\ 
\multicolumn{1}{l}{} &
  \multicolumn{1}{c}{} &
  \multicolumn{1}{c}{} &
  3 &
  \multicolumn{1}{c}{47.20} &
  46.20 &
  \multicolumn{1}{c}{82.97} &
  82.38 \\  
\multicolumn{1}{l}{} &
  \multicolumn{1}{c}{} &
  \multicolumn{1}{c}{} &
  5 &
  \multicolumn{1}{c}{48.30} &
  45.10 &
  \multicolumn{1}{c}{82.84} &
  81.80 \\ \cline{3-8} 
\multicolumn{1}{l}{} &
  \multicolumn{1}{c}{} &
  \multicolumn{1}{c}{Dense} &
  7 &
  \multicolumn{1}{c}{47.80} &
  45.10 &
  \multicolumn{1}{c}{82.79} &
  82.46 \\ \cline{2-8} 
\multicolumn{1}{l}{} &
  \multicolumn{1}{c}{\multirow{6}{*}{9}} &
  \multicolumn{1}{c}{\multirow{5}{*}{Sparse}} &
  1\bf\textsuperscript{†} &
  \multicolumn{1}{c}{48.00} &
  46.40 &
  \multicolumn{1}{c}{83.02} &
  82.41 \\ 
\multicolumn{1}{l}{} &
  \multicolumn{1}{c}{} &
  \multicolumn{1}{c}{} &
  2\bf\textsuperscript{§} &
  \multicolumn{1}{c}{46.60} & 46.20
   &
  \multicolumn{1}{c}{82.88} & 81.66
   \\ 
\multicolumn{1}{l}{} &
  \multicolumn{1}{c}{} &
  \multicolumn{1}{c}{} &
  3 &
  \multicolumn{1}{c}{46.60} &
  42.60 &
  \multicolumn{1}{c}{82.66} &
  82.37 \\ 
\multicolumn{1}{l}{} &
  \multicolumn{1}{c}{} &
  \multicolumn{1}{c}{} &
  5 &
  \multicolumn{1}{c}{48.30} &
  45.10 &
  \multicolumn{1}{c}{82.84} &
  81.40 \\ 
\multicolumn{1}{l}{} &
  \multicolumn{1}{c}{} &
  \multicolumn{1}{c}{} &
  7 &
  \multicolumn{1}{c}{47.90} &
  43.90 &
  \multicolumn{1}{c}{83.19} &
  81.41 \\ \cline{3-8} 
\multicolumn{1}{l}{} &
  \multicolumn{1}{c}{} &
  \multicolumn{1}{c}{Dense} &
  9 &
  \multicolumn{1}{c}{47.90} &
  45.30 &
  \multicolumn{1}{c}{83.45} &
  82.39 \\ \cline{2-8} 
\multicolumn{1}{l}{} &
  \multicolumn{1}{c}{\multirow{7}{*}{11}} &
  \multicolumn{1}{c}{\multirow{6}{*}{Sparse}} &
  1\bf\textsuperscript{†} &
  \multicolumn{1}{c}{47.40} &
  46.00 &
  \multicolumn{1}{c}{83.10} &
  81.45 \\ 
\multicolumn{1}{l}{} &
  \multicolumn{1}{c}{} &
  \multicolumn{1}{c}{} &
  2\bf\textsuperscript{§} &
  \multicolumn{1}{c}{47.60} & 46.60 
   &
  \multicolumn{1}{c}{82.83} & 82.32
   \\ 
\multicolumn{1}{l}{} &
  \multicolumn{1}{c}{} &
  \multicolumn{1}{c}{} &
  3 &
  \multicolumn{1}{c}{47.10} &
  44.70 &
  \multicolumn{1}{c}{83.15} &
  81.71 \\ 
\multicolumn{1}{l}{} &
  \multicolumn{1}{c}{} &
  \multicolumn{1}{c}{} &
  5 &
  \multicolumn{1}{c}{48.10} &
  46.20 &
  \multicolumn{1}{c}{82.48} &
  82.14 \\ 
\multicolumn{1}{l}{} &
  \multicolumn{1}{c}{} &
  \multicolumn{1}{c}{} &
  7 &
  \multicolumn{1}{c}{46.10} &
  45.70 &
  \multicolumn{1}{c}{83.19} &
  82.33 \\ 
\multicolumn{1}{l}{} &
  \multicolumn{1}{c}{} &
  \multicolumn{1}{c}{} &
  9 &
  \multicolumn{1}{c}{48.00} &
  45.70 &
  \multicolumn{1}{c}{82.93} &
  82.37 \\ \cline{3-8} 
\multicolumn{1}{l}{} &
  \multicolumn{1}{c}{} &
  \multicolumn{1}{c}{Dense} &
  11 &
  \multicolumn{1}{c}{48.20} &
  44.10 &
  \multicolumn{1}{c}{83.31} &
  82.24 \\ \cline{2-8} 
\multicolumn{1}{l}{} &
  \multicolumn{1}{c}{\multirow{8}{*}{13}} &
  \multicolumn{1}{c}{\multirow{7}{*}{Sparse}} &
  1\textsuperscript{†} &
  \multicolumn{1}{c}{48.30} &
  45.80 &
  \multicolumn{1}{c}{83.37} &
  82.46 \\
\multicolumn{1}{l}{} &
  \multicolumn{1}{c}{} &
  \multicolumn{1}{c}{} &
  2\textsuperscript{§} &
  \multicolumn{1}{c}{48.10} & 45.20
   &
  \multicolumn{1}{c}{83.23} & 82.44
   \\ 
\multicolumn{1}{l}{} &
  \multicolumn{1}{c}{} &
  \multicolumn{1}{c}{} &
  3 &
  \multicolumn{1}{c}{47.70} &
  46.30 &
  \multicolumn{1}{c}{83.15} &
  82.45 \\
\multicolumn{1}{l}{} &
  \multicolumn{1}{c}{} &
  \multicolumn{1}{c}{} &
  5 &
  \multicolumn{1}{c}{47.30} &
  46.00 &
  \multicolumn{1}{c}{83.46} &
  82.15 \\ 
\multicolumn{1}{l}{} &
  \multicolumn{1}{c}{} &
  \multicolumn{1}{c}{} &
  7 &
  \multicolumn{1}{c}{48.00} &
  44.90 &
  \multicolumn{1}{c}{83.49} &
  82.50 \\
\multicolumn{1}{l}{} &
  \multicolumn{1}{c}{} &
  \multicolumn{1}{c}{} &
  9 &
  \multicolumn{1}{c}{47.60} &
  45.80 &
  \multicolumn{1}{c}{\textbf{83.54}} &
  \textbf{82.55} \\ 
\multicolumn{1}{l}{} &
  \multicolumn{1}{c}{} &
  \multicolumn{1}{c}{} &
  11 &
  \multicolumn{1}{c}{47.30} &
  46.70 &
  \multicolumn{1}{c}{83.37} &
  82.37 \\ \cline{3-8} 
\multicolumn{1}{l}{} &
  \multicolumn{1}{c}{} &
  \multicolumn{1}{c}{Dense} &
  13 &
  \multicolumn{1}{c}{47.90} &
  45.60 &
  \multicolumn{1}{c}{83.02} &
  79.95 \\ \hline
\multicolumn{4}{l}{\textbf{Dynamic Routing (Ours)}} &
  \multicolumn{1}{c}{\textbf{49.20}} &
  \textbf{47.24} &
  \multicolumn{1}{c}{\textbf{83.59}} &
  \textbf{82.82} \\ \hline
\end{tabular}
\end{table}

\subsection{Do the Lengths of Premise and Hypothesis Affect the Model Performances?}
\label{lengthToAcc}
\begin{figure}[H]
    \centering
    \begin{subfigure}{0.6\textwidth}
        \includegraphics[width=\textwidth]{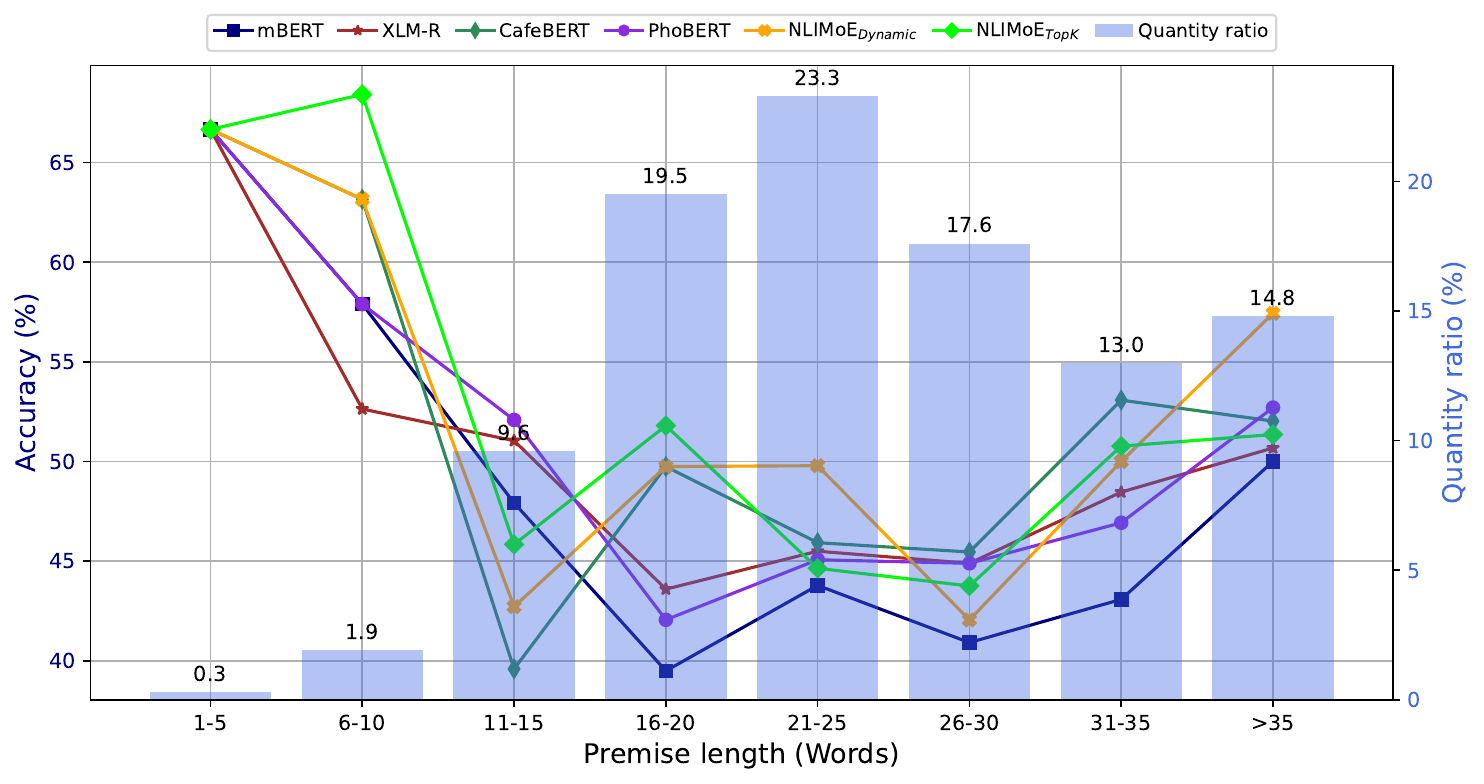}
    \end{subfigure}
    \caption{Premise Sentence Length Affects the Accuracy of Models.}
    \label{fig:P_lengthToAcc}
\end{figure}

Figure \ref{fig:P_lengthToAcc} clearly illustrates the effect of premise sentence length on the accuracy of different models, including our NLIMoE$_{Dynamic}$ model, in the ViANLI dataset. From the chart, it is evident that shorter premise sentences (1-5 words) generally lead to higher accuracy for most models, including mBERT, XLM-R, CafeBERT, and PhoBERT, with our model (NLIMoE$_{Dynamic}$) also performing better at shorter lengths. However, the performance of all models shows a sharp decline in accuracy as the premise length increases, especially beyond 15 words, indicating a significant challenge with longer sentences. This drop in accuracy highlights how the increasing complexity of longer premises introduces more difficult inferences with potential adversarial samples. specific, models such as mBERT, PhoBERT, and XLM-R experience a sharp decline in accuracy when the premise length in range of 16-20 words. As the length continues to increase beyond 20 words, the accuracy of the models slightly increases. In general, it is clear that as the premise sentence length increases, the data becomes more complex, making it harder for the models to handle.

Surprisingly, when comparing NLIMoE$_{Dynamic}$ with other models, our model demonstrates superior stability compared to all other models with an accuracy of about 57\% when the length of the premise sentence is at its highest with over 35 words. This demonstrates that NLIMoE$_{Dynamic}$ is more robust to high sentence lengths, which may be attributed to its Mixture-of-Experts (MoE) architecture, enabling it to handle sentence complexity more effectively.

\begin{figure}[H]
    \centering
    \begin{subfigure}{0.6\textwidth}
        \includegraphics[width=\textwidth]{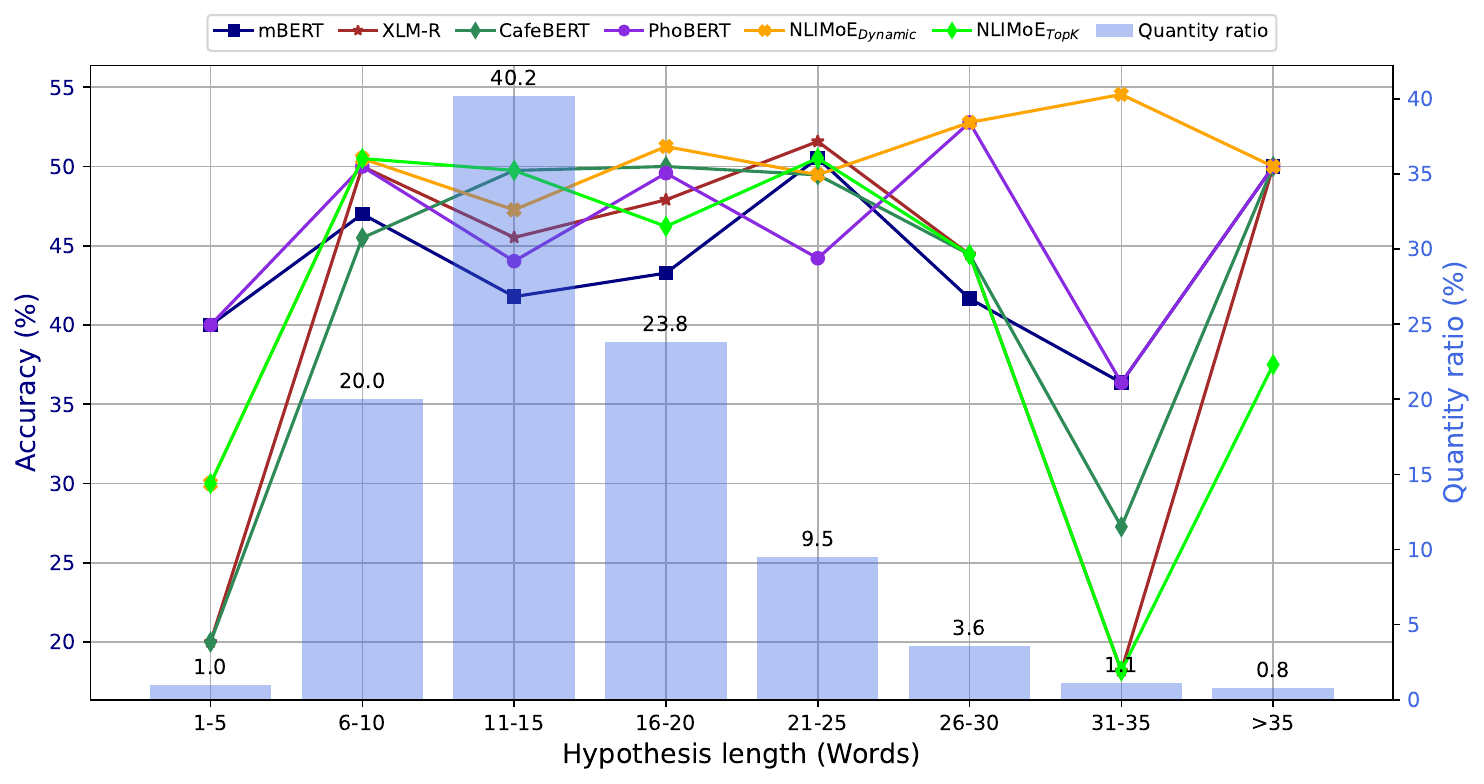}
    \end{subfigure}
    \caption{Hypothesis Sentence Length Affects the Accuracy of Models.}
    \label{fig:H_lengthToAcc}
\end{figure}

Figure \ref{fig:H_lengthToAcc} illustrates the relationship between hypothesis sentence length and model accuracy. When the hypothesis is short (1–5 words), most models perform poorly. As the hypothesis length increases from 6 up to more than 35 words, the accuracy of all baseline models fluctuates considerably, with the exception of NLIMoE$_{Dynamic}$. Thanks to its flexible expert selection mechanism, NLIMoE${Dynamic}$ maintains stable performance as sentence length grows. In particular, for longer hypotheses (26–35 words), our model achieves the highest accuracy among all compared models. Notably, in the 31–35 word range, mBERT, PhoBERT, XLM-R, CafeBERT, and NLIMoE$_{TopK}$ all suffer a sharp drop in accuracy, while NLIMoE$_{Dynamic}$ continues to improve. These results demonstrate that hypothesis length strongly influences task difficulty, as longer sentences introduce greater structural and semantic complexity that many models struggle to process. However, our Mixture-of-Experts architecture with dynamic routing effectively mitigates this challenge, highlighting its advantage in handling long, complex, and adversarial data in Vietnamese NLI.

%\subsection{Does Word Overlap Simplify NLI or Highlight NLIMoE Advantages on ViANLI?}

\subsection{Does Word Overlap Affect the Model Performances?}
\label{WordOverLapToAcc}

In the process of constructing the NLI dataset, one important issue is that annotators may reuse words from the premise sentence when writing the hypothesis. This reuse of words can serve as a straightforward indicator, helping the model easily identify the semantic relationship between the premise and hypothesis, as the shared words in both sentences create a clear similarity. However, this also reduces the difficulty and complexity of the data, as the model may rely on the word overlap to determine the relationship without needing to perform deeper inference. Therefore, we conducted a performance analysis of various models based on the word overlap ratio to comprehensively assess the difficulty of the ViANLI dataset.

\begin{figure}[H]
    \centering
    \begin{subfigure}{0.6\textwidth}
        \includegraphics[width=\textwidth]{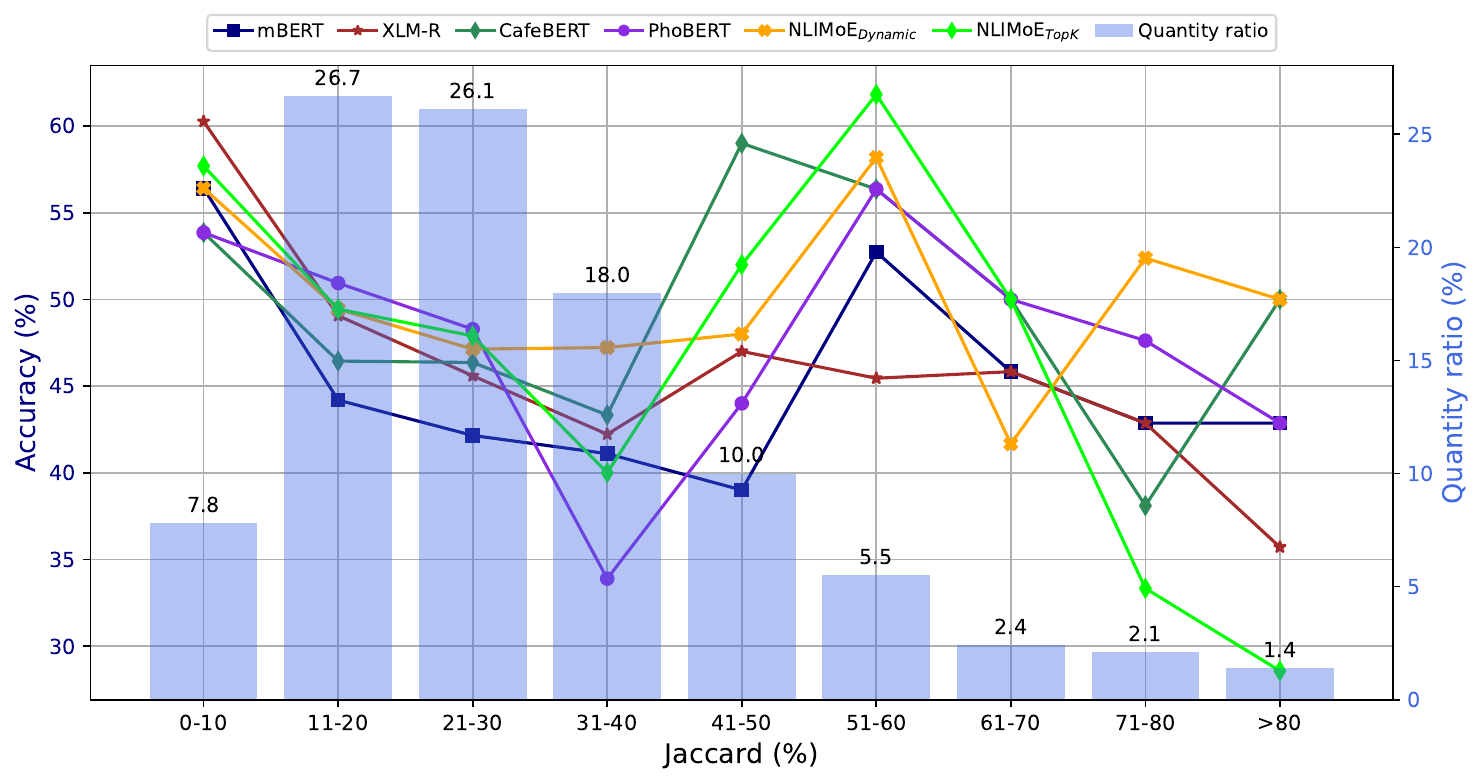}
    \end{subfigure}
    \caption{Word Overlap Rate Affects the Accuracy of Models.}
    \label{fig:WordOverlapToAcc}
\end{figure}

In Figure \ref{fig:WordOverlapToAcc}, when the word overlap ratio between the premise and hypothesis is low (below 10\%), the accuracy of all models remains relatively high. However, as the overlap ratio increases from 11\% to 40\%, the performance of models such as mBERT, PhoBERT, XLM-R, CafeBERT, NLIMoE$_{TopK}$, and even our model declines significantly. Although most models show some recovery when the overlap ratio reaches 51–60\%, their accuracy quickly drops again once the overlap increases to 61–70\%. Notably, when the overlap becomes very high (71–80\%), most of the models fail to effectively exploit the high word overlap—especially NLIMoE$_{TopK}$, whose accuracy falls to around 25\%. In contrast, NLIMoE$_{Dynamic}$ demonstrates improved performance in this range and achieves the highest accuracy among all models. Overall, NLIMoE$_{Dynamic}$ exhibits greater stability at low overlap levels and superior performance under high-overlap conditions, highlighting its robustness compared to other approaches.

In comparison to the previous ViNLI study \cite{huynh-etal-2022-vinli}, where the word overlap ratio between the premise and hypothesis was considered a factor that helps identify the inferential relationship between the two sentences, our ViANLI dataset offers a different perspective. Although reusing words in the hypothesis can assist the model in easily recognizing the relationship, our dataset maintains a high level of difficulty because we design the hypothesis sentences in a way that requires more complex reasoning, not just relying on word overlap. This result further highlights the distinction of our adversarial NLI dataset, where even when there is word reuse from the premise, the model still faces significant challenges in identifying and reasoning about the relationship between the premise and hypothesis. Consequently, our NLIMoE$_{Dynamic}$ model shows a distinct advantage, as it effectively leverages these lexical patterns while still capturing the deeper semantic reasoning needed for accurate inference.

\subsection{Does New Word Rate Challenge NLI Models?}
\label{NewWordRateToAcc}

\begin{figure}[H]
    \centering
    \begin{subfigure}{0.6\textwidth}
        \includegraphics[width=\textwidth]{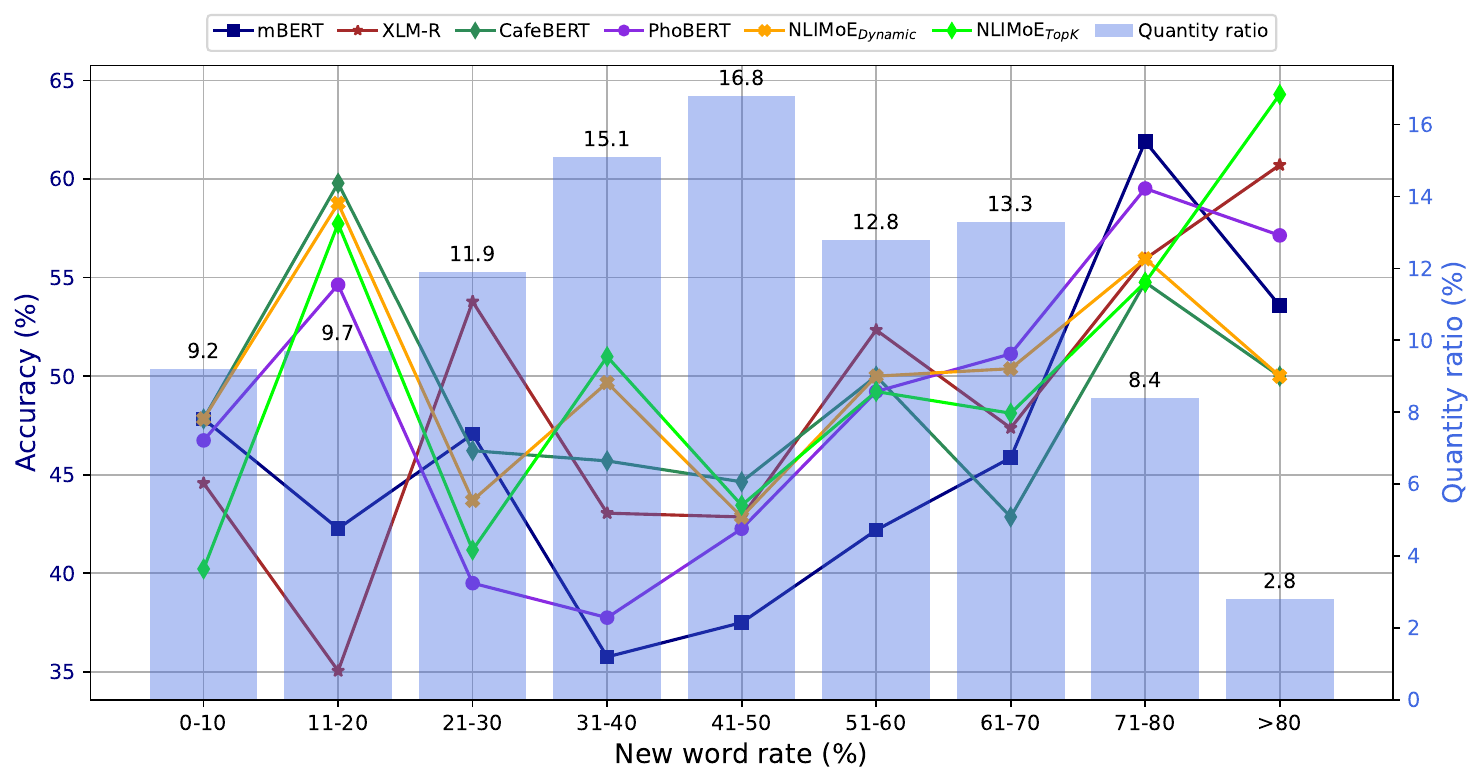}
    \end{subfigure}
    \caption{New Word Rate Affects the Accuracy of Models.}
    \label{fig:NewWordToAcc}
\end{figure}

In contrast to the word overlap rate, the new word rate measures the proportion of words used by the annotator to write the hypothesis sentence that do not appear in the premise sentence. This is also considered a factor that may affect the performance of machine learning models, as the more new words in the hypothesis sentence, the more challenging it may be for the models to infer. To evaluate this, we also measure the accuracy of the models in terms of new word rates in the hypothesis sentence on the development set (see Figure \ref{fig:NewWordToAcc}).

Figure \ref{fig:NewWordToAcc} show that the performance of most models fluctuates and declines as the new word rate increases from 0\% to 30\%, with the exception of XLM-R. When the rate rises from 31\% to 40\%, models such as mBERT, PhoBERT, XLM-R, and CafeBERT experience a drop in accuracy, while NLIMoE$_{Dynamic}$ and NLIMoE$_{TopK}$ show improvements. As the new word rate continues to increase from 41\% to 80\%, the performance of all models improves significantly. Although, our proposed NLIMoE$_{Dynamic}$ does not achieve the highest accuracy in any specific range, it demonstrates the greatest stability across different levels of new word occurrence, maintaining consistent performance without the large fluctuations observed in other models.

Overall, once again, we found the opposite of the results on the ViNLI dataset. The analysis results on ViNLI show that as the rate of new words increases, it causes the performance of the model to gradually decrease. There is a clear dependence of accuracy on the new word rate, whereas this dependence is less evident in our data. However, the new word rate in ViANLI is not the only factor determining the complexity of the data. This shows that the data generated from our process warrants a high level of difficulty and challenges for modern machine learning models.

\subsection{Do Annotator Artifacts in Hypotheses Make ViANLI Less Challenging Than Other Datasets?}
In natural language inference, annotator artifacts refer to unintended patterns or biases introduced by human annotators during the data creation process, such as repetitive word choices, sentence structures, or stylistic preferences in hypothesis sentences. These artifacts can allow models to make predictions based on superficial cues rather than true semantic understanding, thereby reducing the challenge of the dataset. To investigate whether such artifacts exist in the hypothesis sentences of ViANLI, we evaluate the accuracy of models when they are trained and tested solely on hypothesis sentences. This analysis helps assess the presence of annotator artifacts and their impact on the predictive ability of models during training, as well as the overall difficulty of the dataset. The results of evaluating the models on data from each round and the full ViANLI dataset are summarized in Table \ref{tab:hypothesis_only_result}.

\begin{table}[H]
\renewcommand{\arraystretch}{1.4}
\centering
\caption{Model Accuracy (\%) That is Trained on Hypothesis-only Data.}
\label{tab:hypothesis_only_result}
%\resizebox{1\columnwidth}{!}{%
\begin{tabular}{lrrrrrrrr}
\hline
\multicolumn{1}{c}{\multirow{2}{*}{\textbf{Model}}} &
  \multicolumn{2}{c}{\textbf{ViA1}} &
  \multicolumn{2}{c}{\textbf{ViA2}} &
  \multicolumn{2}{c}{\textbf{ViA3}} &
  \multicolumn{2}{c}{\textbf{ViANLI}} \\ \cline{2-9} 
\multicolumn{1}{c}{} &
  \multicolumn{1}{c}{\textbf{Dev}} &
  \multicolumn{1}{c}{\textbf{Test}} &
  \multicolumn{1}{c}{\textbf{Dev}} &
  \multicolumn{1}{c}{\textbf{Test}} &
  \multicolumn{1}{c}{\textbf{Dev}} &
  \multicolumn{1}{c}{\textbf{Test}} &
  \multicolumn{1}{c}{\textbf{Dev}} &
  \multicolumn{1}{c}{\textbf{Test}} \\ \hline
mBERT &
  \multicolumn{1}{r}{39.15} &
  40.00 &
  \multicolumn{1}{r}{35.75} &
  37.87 &
  \multicolumn{1}{r}{38.52} &
  32.35 &
  \multicolumn{1}{r}{37.80} &
  36.70 \\
XLM-R &
  \multicolumn{1}{r}{40.60} &
  40.50 &
  \multicolumn{1}{r}{38.18} &
  36.66 &
  \multicolumn{1}{r}{40.58} &
  38.53 &
  \multicolumn{1}{r}{39.80} &
  38.60 \\
PhoBERT &
  \multicolumn{1}{r}{40.30} &
  40.90 &
  \multicolumn{1}{r}{40.90} &
  40.00 &
  \multicolumn{1}{r}{39.11} &
  38.52 &
  \multicolumn{1}{r}{40.10} &
  39.80 \\
CafeBERT &
  \multicolumn{1}{r}{42.42} &
  42.42 &
  \multicolumn{1}{r}{\textbf{41.51}} &
  36.96 &
  \multicolumn{1}{r}{38.23} &
  40.58 &
  \multicolumn{1}{r}{40.70} &
  40.00 \\     
NLIMoE$_{TopK}$ &
  \multicolumn{1}{r}{38.78} &
  41.51 &
  \multicolumn{1}{r}{41.21} &
  41.51 &
  \multicolumn{1}{r}{\textbf{41.76}} &
   \textbf{41.17}&
  \multicolumn{1}{r}{40.70} &
  41.40 \\ \hline
  NLIMoE$_{Dynamic}$ &
  \multicolumn{1}{r}{\textbf{42.42}} &
  \textbf{45.75} &
  \multicolumn{1}{r}{40.00} &
  \textbf{42.12} &
  \multicolumn{1}{r}{40.88} &
  39.11 &
  \multicolumn{1}{r}{\textbf{40.80}} &
  \textbf{42.20} \\ \hline
\end{tabular}%
%}
\end{table}

Overall, the performance of the models on the dev and test sets of ViANLI is quite low when the models are trained on hypothesis sentences only, which suggests that the artifacts in the hypothesis sentence do not seem clear enough to help the models make accurate predictions. Our model, NLIMoE$_{Dynamic}$, achieves the highest accuracy on the dev set at 40.8\%. This accuracy is significantly lower compared to other datasets such as OCNLI (~66\%) \citep{hu2020ocnli}, IndoNLI (~60\%) \citep{mahendra-etal-2021-indonli}, SNLI (~69\%) \citep{poliak2018hypothesis}, MultiNLI (~62\%) \citep{bowman2020new}, and ViNLI (~58\%) \citep{huynh-etal-2022-vinli}. This indicates that during the construction of these datasets, annotators intentionally limited the inclusion of cues in the hypothesis sentences to hinder the model in making accurate predictions. In contrast to our data, the rigorously designed data construction process avoids hypothetical artifacts, which helps ensure the difficulty and challenge of our dataset. To investigate the artifacts left by annotators when writing hypothesis sentences, we calculated the PMI (Pointwise Mutual Information) score on the hypothesis sentences of our dataset to observe which words in the hypothesis might distinguish the labels from one another in a manner similar to the ViNLI, OCNLI, and IndoNLI datasets.

The results of the PMI scores are presented in Table \ref{tab:PMI}, show that datasets such as IndoNLI and OCNLI, built using conventional methods, contain clear linguistic cues in hypothesis sentences that enable models to easily identify semantic relationships with premise sentences. Notably, for the contradiction label, negation words like “no” and “not” rank among the top three words with the highest PMI scores, serving as strong indicators for this label. In contrast, ViNLI exhibits a distinct pattern: negation words appear more frequently in hypothesis sentences for the entailment label. The authors of ViNLI explain that annotators were instructed to limit the use of negation words when crafting contradiction hypotheses, inadvertently leading to their increased presence in entailment hypotheses. Conversely, our dataset, ViANLI, shows no prominent lexical cues in hypothesis sentences that would allow models to easily distinguish semantic labels. The top five words with the highest PMI scores across all three labels (entailment, contradiction, and neutral) are relatively similar, indicating a lack of distinctive lexical patterns. This design makes ViANLI significantly more challenging than other datasets, as models cannot rely solely on superficial lexical features to accurately predict semantic labels. Instead, they must consider the deeper semantic intent targeted by the annotators, enhancing the difficulty and robustness of the dataset.

\begin{CJK}{UTF8}{gbsn}
\begin{table}[H]
\renewcommand{\arraystretch}{1.2}
\centering
\caption{Comparison of PMI Scores to Identify Representative Words for Labels to Distinguish from Other Labels. We select the Top 3 (word, label) pairs from ViANLI for comparison with IndoNLI and OCNLI, and the Top 5 (word, label) pairs from ViANLI for comparison with ViNLI. The results marked with \textsuperscript{*} are extracted from the ViNLI dataset \citep{van2022error}, those marked with \textsuperscript{†} are from the IndoNLI dataset \citep{mahendra-etal-2021-indonli}, and those marked with \textsuperscript{§} are from the SINGLE subset of OCNLI dataset \cite{hu2020ocnli}.}
\resizebox{\columnwidth}{!}{%
\begin{tabular}{llllcccccc}
\hline
\multicolumn{1}{c}{\multirow{2}{*}{\textbf{\#Top}}} & \multicolumn{4}{c}{\textbf{Word}} & \multirow{2}{*}{\textbf{Label}} & \multicolumn{4}{c}{\textbf{PMI}} \\ \cline{2-5} \cline{7-10} 
\multicolumn{1}{c}{} & \multicolumn{1}{c}{\textbf{ViANLI}} & \multicolumn{1}{c}{\textbf{ViNLI\textsuperscript{*}}} & \multicolumn{1}{c}{\textbf{IndoNLI\textsuperscript{†}}} & \textbf{OCNLI\textsuperscript{§}} &  & \multicolumn{1}{c}{\textbf{ViANLI}} & \multicolumn{1}{c}{\textbf{ViNLI\textsuperscript{*}}} & \multicolumn{1}{c}{\textbf{IndoNLI\textsuperscript{†}}} & \textbf{OCNLI\textsuperscript{§}} \\ \hline
\#1 & \multicolumn{1}{l}{được - tobe + verb (past participle)} & \multicolumn{1}{l}{và - and} & \multicolumn{1}{l}{salah - wrong} & - & E & \multicolumn{1}{c}{0.32} & \multicolumn{1}{c}{0.18} & \multicolumn{1}{c}{0.72} & - \\ 
\#2 & \multicolumn{1}{l}{các - some/several} & \multicolumn{1}{l}{các - some/several} & \multicolumn{1}{l}{sekitar - around} & - & E & \multicolumn{1}{c}{0.31} & \multicolumn{1}{c}{0.22} & \multicolumn{1}{c}{0.72} & - \\ 
\#3 & \multicolumn{1}{l}{của - of/object’s} & \multicolumn{1}{l}{của - of/object’s} & \multicolumn{1}{l}{suatu - something} & - & E & \multicolumn{1}{c}{0.31} & \multicolumn{1}{c}{0.28} & \multicolumn{1}{c}{0.58} & - \\ 
\#4 & \multicolumn{1}{l}{là - tobe} & \multicolumn{1}{l}{trong - in/inside} & \multicolumn{1}{c}{-} & - & E & \multicolumn{1}{c}{0.25} & \multicolumn{1}{c}{0.35} & \multicolumn{1}{c}{-} & - \\ 
\#5 & \multicolumn{1}{l}{và - and} & \multicolumn{1}{l}{\textbf{không - no/not}} & \multicolumn{1}{c}{-} & - & E & \multicolumn{1}{c}{0.24} & \multicolumn{1}{c}{0.35} & \multicolumn{1}{c}{-} & - \\ \hline
\#1 & \multicolumn{1}{l}{và - and} & \multicolumn{1}{l}{của - of/object’s} & \multicolumn{1}{l}{\textbf{bukan - no}} & \multicolumn{1}{l}{任何 - any} & C & \multicolumn{1}{c}{0.33} & \multicolumn{1}{c}{0.19} & \multicolumn{1}{c}{1.28} & 0.89 \\ 
\#2 & \multicolumn{1}{l}{của - of/object’s} & \multicolumn{1}{l}{là - tobe} & \multicolumn{1}{l}{\textbf{tidak - no}} & \multicolumn{1}{l}{\textbf{没有 - no}} & C & \multicolumn{1}{c}{0.26} & \multicolumn{1}{c}{0.21} & \multicolumn{1}{c}{1.24} & 0.83 \\ 
\#3 & \multicolumn{1}{l}{được - tobe + verb (past participle)} & \multicolumn{1}{l}{trong - in/inside} & \multicolumn{1}{l}{apapun - anything} & \multicolumn{1}{l}{\textbf{无关 - not related}} & C & \multicolumn{1}{c}{0.22} & \multicolumn{1}{c}{0.22} & \multicolumn{1}{c}{1.20} & 0.72 \\ 
\#4 & \multicolumn{1}{l}{có- has/have} & \multicolumn{1}{l}{và - and} & \multicolumn{1}{c}{-} & - & C & \multicolumn{1}{c}{0.21} & \multicolumn{1}{c}{0.23} & \multicolumn{1}{c}{-} & - \\ 
\#5 & \multicolumn{1}{l}{đã - was/were/verb(past/ed)} & \multicolumn{1}{l}{các - some/several} & \multicolumn{1}{c}{-} & - & C & \multicolumn{1}{c}{0.21} & \multicolumn{1}{c}{0.30} & \multicolumn{1}{c}{-} & - \\ \hline
\#1 & \multicolumn{1}{l}{là - tobe} & \multicolumn{1}{l}{một - one/a/an} & \multicolumn{1}{l}{selain - aside from} & - & N & \multicolumn{1}{c}{0.25} & \multicolumn{1}{c}{0.27} & \multicolumn{1}{c}{1.08} & - \\ 
\#2 & \multicolumn{1}{l}{có - has/have} & \multicolumn{1}{l}{là - tobe} & \multicolumn{1}{l}{juga - also} & - & N & \multicolumn{1}{c}{0.25} & \multicolumn{1}{c}{0.31} & \multicolumn{1}{c}{1.05} & - \\ 
\#3 & \multicolumn{1}{l}{được - tobe + verb (past participle)} & \multicolumn{1}{l}{và - and} & \multicolumn{1}{l}{banyak - a lot} & - & N & \multicolumn{1}{c}{0.24} & \multicolumn{1}{c}{0.32} & \multicolumn{1}{c}{0.94} & - \\ 
\#4 & \multicolumn{1}{l}{của - of/object’s} & \multicolumn{1}{l}{có - has/have} & \multicolumn{1}{c}{-} & - & N & \multicolumn{1}{c}{0.24} & \multicolumn{1}{c}{0.33} & \multicolumn{1}{c}{-} & - \\ 
\#5 & \multicolumn{1}{l}{là - tobe} & \multicolumn{1}{l}{trong - in/inside} & \multicolumn{1}{c}{-} & - & N & \multicolumn{1}{c}{0.23} & \multicolumn{1}{c}{0.33} & \multicolumn{1}{c}{-} & - \\ \hline
\end{tabular}%
}
\label{tab:PMI}
\end{table}

\end{CJK}

\subsection{Comparison of NLIMoE Reasoning Ability with Other Models by Inference Types}
\label{sec:per_inference_type}

In this section, we evaluate the reasoning capabilities of the NLIMoE model compared to baseline models (CafeBERT, XLM-R, PhoBERT, and mBERT) on the ViANLI Dev Set, with a focus on analyzing performance across different inference types. The primary objective is to conduct a linguistic analysis that uncovers how these models handle diverse reasoning challenges inherent in Vietnamese natural language inference.

Table \ref{tab:per_inference} provides a detailed comparison of NLIMoE$_{Dynamic}$’s reasoning ability with other models across seven distinct inference types: numerical \& quantitative reasoning, reference \& names, standard logical inferences, lexical reasoning, tricky linguistic inference, external knowledge reasoning, and cultural \& contextual inference. NLIMoE$_{Dynamic}$ demonstrates superior performance in most categories, achieving the highest scores in numerical \& quantitative reasoning (51.42\%), reference \& names (60.00\%), standard (49.79\%), lexical (52.55\%), and external knowledge (49.13\%). NLIMoE$_{Dynamic}$ effectively captures complex linguistic phenomena such as numerical relationships, proper noun references, lexical ambiguities, and supplementary information from external sources. However, the performance of NLIMoE$_{TopK}$ is highest in cultural \& contextual inference (48.51\%), indicating that NLIMoE$_{Dynamic}$ still faces challenges in reasoning over phenomena tied to Vietnamese culture and context. Overall, NLIMoE$_{Dynamic}$ demonstrates strong and consistent reasoning capabilities across most inference types compared to other models, particularly in handling numerical \& quantitative reasoning, resolving reference-based ambiguities, and addressing lexical variations, which are often difficult for multilingual pre-trained models on low-resource languages like Vietnamese.

\begin{table}[H]
\centering
\caption{Comparison of NLIMoE$_{Dynamic}$ Reasoning Ability (Accuracy - \%) with Other Models on the ViANLI Dev Set by Different Inference Types.}
\renewcommand{\arraystretch}{1.4}
\resizebox{0.8\columnwidth}{!}{%
\label{tab:per_inference}
\begin{tabular}{lcccccccc}
\hline
\textbf{Models} &
  \multicolumn{1}{l}{\textbf{NaQR}} &
  \multicolumn{1}{l}{\textbf{RaN}} &
  \multicolumn{1}{l}{\textbf{SLI}} &
  \multicolumn{1}{l}{\textbf{LR}} &
  \multicolumn{1}{l}{\textbf{TLI}} &
  \multicolumn{1}{l}{\textbf{EKR}} &
  \multicolumn{1}{l}{\textbf{CaCI}} &
  \multicolumn{1}{l}{\textbf{Average}} \\ \hline 
mBERT         & 43.32          & 40.00          & 44.00          & 45.62          & 38.05          & 48.83          & 42.57          & 43.19 \\ 
XLM-R         & 47.77          & 52.00          & 47.03          & 51.61          & \textbf{46.91} & 44.41          & 47.52          & 48.17 \\ 
PhoBERT       & 46.96          & 56.00          & 47.17          & 49.76          & 33.62          & 48.37          & 38.61          & 45.78 \\ 
CafeBERT      & 50.62          & 52.00          & 49.11          & 50.23          & 44.24          & 47.44          & 39.61          & 47.61 \\ 
NLIMoE$_{TopK}$    & 47.77          & 57.00          & 48.55          & 49.76          & 40.71          & 48.61          & \textbf{48.51} & 48.69 \\ 
NLIMoE$_{Dynamic}$ & \textbf{51.42} & \textbf{60.00} & \textbf{49.79} & \textbf{52.55} & 43.36          & \textbf{49.13} & 43.56          & 49.97 \\ 
\hline
Average       & 47.97          & 52.83          & 47.61          & 49.92          & 41.14          & 47.79          & 43.39          & -     \\ \hline
\end{tabular}}
\end{table}

Table \ref{tab:per_num_inference} compares the reasoning ability of NLIMoE$_{Dynamic}$ with baseline models when handling combinations of multiple inference types on the ViANLI Dev Set. Results show that NLIMoE$_{Dynamic}$ delivers competitive performance on premise–hypothesis pairs involving single-type inference (50.00\%), slightly behind CafeBERT (50.61\%) but outperforming most other baselines. In pairs requiring two-type inference, NLIMoE$_{Dynamic}$ maintains stable accuracy (47.52\%), although CafeBERT achieves the highest score (52.00\%).

Notably, in the most challenging cases involving three-type inference where models must simultaneously reason over multiple intertwined linguistic phenomena, NLIMoE$_{Dynamic}$ achieves a significant performance advantage, reaching 59.80\%, the highest among all models and surpassing the second-best (XLM-R, 54.90\%). This highlights NLIMoE$_{Dynamic}$’s superior capability to integrate diverse reasoning skills and effectively tackle complex, multi-faceted inference scenarios in Vietnamese NLI, benefiting from its adaptive Mixture-of-Experts mechanism.

\begin{table}[H]
\centering
\caption{Comparison of NLIMoE$_{Dynamic}$ Reasoning Ability with Other Models on the ViANLI Dev Set by combining multiple types of inference.}
\label{tab:per_num_inference}
\resizebox{0.45\columnwidth}{!}{%
\begin{tabular}{lccc}
\hline
\multicolumn{1}{c}{\multirow{2}{*}{\textbf{Models}}} & \multicolumn{3}{c}{\textbf{\#Inference Type}}                                                  \\ \cline{2-4} 
\multicolumn{1}{c}{}                                 & \multicolumn{1}{c}{\textbf{1 type}} & \multicolumn{1}{c}{\textbf{2 types}} & \textbf{3 types} \\ \hline
mBERT       & \multicolumn{1}{c}{42.44}          & \multicolumn{1}{c}{43.53}          & 50.00         \\ 
XLM-R       & \multicolumn{1}{c}{48.25}          & \multicolumn{1}{c}{45.59}          & 54.90          \\ 
PhoBERT     & \multicolumn{1}{c}{47.67}          & \multicolumn{1}{c}{46.14}          & 50.00         \\ 
CafeBERT    & \multicolumn{1}{c}{\textbf{50.61}} & \multicolumn{1}{c}{\textbf{52.00}}          & 49.1          \\
NLIMoE$_{TopK}$ & \multicolumn{1}{c}{47.09}          & \multicolumn{1}{c}{48.21} & 50.98         \\
NLIMoE$_{Dynamic}$   & \multicolumn{1}{c}{50.00}          & \multicolumn{1}{c}{47.52}          & \textbf{59.80} \\
\hline
\end{tabular}}
\end{table}

\subsection{Error Analysis}
Confusion matrix analysis serves as a vital tool in evaluating the performance of machine learning models, enabling precise identification of error types encountered during label classification. The purpose of this analysis is to gain deeper insights into the ability to differentiate between the labels including contradiction, entailment, and neutral, thereby uncovering model weaknesses, especially in complex or adversarial data scenarios. This process not only aids in enhancing model accuracy but also highlights the semantic factors models must learn to achieve more precise predictions. Results from the confusion matrices of the proposed NLIMoE$_{Dynamic}$ and other models, including mBERT, PhoBERT, XLM-R, CafeBERT, and NLIMoE$_{TopK}$ on the development set of the ViANLI dataset are presented in Figure \ref{fig:confusion_matrix} (in \ref{ConfusionMatrix}).

The confusion matrices reveal distinct challenges posed by the Adversarial NLI dataset, highlighting varying limitations among the compared models. mBERT shows a clear bias toward Neutral predictions, often misclassifying both contradiction and entailment as neutral, suggesting difficulty in capturing the adversarial nuances requiring fine-grained semantic understanding. PhoBERT and XLM-R outperform mBERT in entailment predictions, demonstrating a better grasp of entailment relationships, but they encounter consistency issues elsewhere, PhoBERT struggles to maintain accuracy on contradiction, frequently misclassifying it, while XLM-R has difficulty sustaining performance on neutral, often confusing it with other labels. CafeBERT appears more stable across all labels compared to the other models, yet it still exhibits a notable weakness, with a substantial number of contradiction samples being mispredicted as neutral, indicating sensitivity to the dataset adversarial design. NLIMoE$_{TopK}$ performs well on the entailment label but misclassifies a substantial number of patterns belonging to the contradiction and neutral labels. In contrast, NLIMoE$_{Dynamic}$ demonstrates greater stability than the other models, showing particularly strong performance on entailment and neutral labels, though it still misclassifies a notable portion of contradiction samples, likely due to the adversarial nature of the dataset. These findings suggest that the Mixture-of-Experts architecture of NLIMoE$_{Dynamic}$ offers a significant advantage in handling diverse inference scenarios, representing a clear improvement over the more biased or inconsistent tendencies observed in the other models.

Table \ref{tab:errorsample} (in \ref{ComparisonofPredictedLabelsbyNLIMoE})provides examples of incorrect predictions by models, offering valuable insight into challenges posed by the ViANLI dataset. The dataset tests models with subtle semantic ambiguities, complex contextual dependencies, and adversarial perturbations, as seen in premise-hypothesis pairs. Examples highlight difficulties in detecting contradictions from implicit cues (e.g., gender specificity in testicular torsion), recognizing temporal discrepancies (e.g., 18/5 to 20/5 spans 2 days, not 5), handling paraphrased entailments (e.g., the President of Palau meeting Taiwan’s leader implies both leaders participated in an event), and managing numerical inconsistencies (e.g., currency conversion in LG pricing). These characteristics exploit weaknesses in semantic understanding, leading to frequent misclassifications, particularly for contradiction, across mBERT, PhoBERT, XLM-R, and CafeBERT, underscoring the nuanced complexity of the dataset. While NLIMoE$_{Dynamic}$  also errors in cases like the first two examples alongside other models, it correctly predicts labels in scenarios involving paraphrased entailments and numerical comparisons, as seen in the last two examples, demonstrating enhanced capability in handling complex semantic relationships.

To further analyze this behavior, we investigate the distribution of experts activated across different inference types in the ViANLI dev set as in Figure \ref{fig:active-expert} (\ref{active_expert}). The analysis reveals that Expert 3 consistently emerges as the dominant contributor, being frequently activated across nearly all reasoning categories. Specifically, for numerical and quantitative reasoning, Expert 3 plays a pivotal role, while Experts 1, 5, and 6 serve as complementary contributors. For example, in Case 4 of Table \ref{tab:errorsample}, NLIMoE successfully activated three out of these four experts, enabling accurate numerical inference. Conversely, for standard logical inferences (e.g., Case 3), Experts 1 and 6 demonstrate stronger specialization, and NLIMoE successfully activated these two experts and effectively leveraged both to arrive at the correct prediction. These observations highlight that the Mixture-of-Experts architecture in NLIMoE$_{Dynamic}$ not only partitions inference tasks among specialized experts but also dynamically selects the most relevant subset. This adaptive routing provides a distinct advantage, allowing the model to better address the diverse challenges of ViANLI and achieve superior accuracy through both specialization and expert collaboration.

\section{Conclusion and Future Work}
\label{sect:conclusion}

In this article, we introduced ViANLI, the first adversarial dataset for Vietnamese natural language inference, including over 10,000 premise-hypothesis pairs generated by the human-and-machine-in-the-loop approach with human-machine verification. ViANLI exhibits greater complexity compared to previous Vietnamese NLI datasets, significantly challenging state-of-the-art models and setting a new benchmark for future Vietnamese and multilingual NLI research. We also proposed NLIMoE$_{Dynamic}$ with a dynamic routing mechanism capable of choosing the number of experts dynamically based on the complexity of the input, this Mixture-of-Experts model tailored for adversarial NLI, which compares to all baselines with an accuracy of 47.3\% on the ViANLI test set, compared to 45.5\% for XLM-R. Notably, NLIMoE$_{Dynamic}$ demonstrates superior inference capabilities, particularly in handling long sentences, high new word rates in hypotheses, and effectively resolves entailment ambiguity cases such as numerical \& quantitative, reference \& names, standard, lexical, and external knowledge reasoning, as shown in our analyses. Furthermore, augmenting Vietnamese NLI datasets such as ViNLI, VLSP2021-NLI, and VnNewsNLI with ViANLI data enhances the performance of NLI models like mBERT, PhoBERT, XLM-R, and NLIMoE highlighting the value of ViANLI as a key resource for advancing NLI research in Vietnamese.

In future work, our dataset creation approach of ViANLI can be used with a wider range of text genres, especially focusing on those reasoning types that machine models commonly fail, to further enhance model knowledge. This approach provides a replicable paradigm for NLI research in other low-resource languages, driving global AI technology toward more inclusive and balanced development. Indeed, while large language models demonstrate exceptional potential across industries, previous research \cite{guo2025surge} has also emphasized that the emergence of such models exacerbates global resource inequality. In addition, other state-of-the-art generative models such as MT5 \citep{xue-etal-2021-mt5}, ViT5 \citep{phan-etal-2022-vit5}, mBART \citep{liu-etal-2020-multilingual-denoising}, and BARTpho \citep{bartpho} are explored on NLIMoE$_{Dynamic}$. Furthermore, we aim to refine our NLIMoE$_{Dynamic}$ model, making it more adept at handling high-complexity data patterns when combined with large language models (LLMs). Furthermore, we also combine ViANLI with other NLI datasets to enhance effectiveness in tasks such as machine reading comprehension and text summarization.

\section*{Acknowledgement}
This research is funded by Vietnam National University HoChiMinh City (VNU-HCM) under grant number NCM2025-26-02.

\section*{Declarations of Interest}
The authors declare that they have no conflict of interest.

\section*{Data Availability}

Data will be made available on reasonable request.

\section*{Author Contribution}
Tin Van Huynh: Conceptualization; Data curation; Formal analysis; Investigation; Methodology; Validation; Visualization; Writing - original draft.

Kiet Van Nguyen: Conceptualization; Data curation; Formal analysis; Investigation; Methodology; Validation; Visualization; Writing - review\&editing.

Ngan Luu-Thuy Nguyen: Conceptualization; Formal analysis; Investigation; Methodology; Supervision; Validation; Writing - review\&editing.

% \bibliographystyle{elsarticle-harv} 
%\bibliographystyle{apalike}
%\biboptions{authoryear}

\bibliography{cas-refs}

\appendix

\section{Examples from ViANLI Dataset} 
\label{ExamplesfromViANLIDataset}

\begin{table*}[]
\centering
\caption{Examples from ViANLI Dataset. Where Pred. Is the Predictive Label of the Model, Vali.1 Is the Three Confirmation Labels of the Three Verifiers in the Phase 4.1. C, E, and N Are the Abbreviations for Contradiction, Entailment, and Neutral.}
\label{tab:Examples}
\resizebox{0.7\textwidth}{!}{\begin{tabular}{p{4.5cm}p{2.5cm}p{4cm}p{1cm}p{2.3cm}}
\hline
 \begin{tabular}[t]{p{4.5cm}}\bf Premise \end{tabular}                & 
 \begin{tabular}[t]{p{2.5cm}}\bf Hypothesis \end{tabular}  & 
 \begin{tabular}[t]{p{4cm}}\bf Reason \end{tabular}& 
 \begin{tabular}[t]{p{1cm}}\bf Round \end{tabular}
 &\begin{tabular}[t]{p{2.3cm}}\textbf{Gold label}\\ \bf Pred. label\\ \bf Vali.1 labels\end{tabular}                                                                                                                                                                                       \\ \hline
 
\begin{tabular}[t]{p{4.5cm}}Các giả thuyết trước đây cho rằng Hadrosaurid đã di cư từ Bắc Mỹ tới châu Á nhưng khám phá mới lại cho thấy điều ngược lại. \textit{(Previous theories suggested that the Hadrosaurid migrated from North America to Asia, but the new discovery suggests otherwise.)}\end{tabular}                 & \begin{tabular}[t]{p{2.5cm}}Các phát hiện mới đã phủ định những phỏng đoán trước đây, Hadrosaurid không hề có sự thay đổi nơi sinh sống. \textit{(The new findings have denied previous conjectures, that the Hadrosaurid did not change its habitat.)}\end{tabular}  & \begin{tabular}[t]{p{4cm}}Dùng nhiều từ ngữ khác với câu ban đầu nên mô hình có thể đoán sai. \textit{(Using many different words from the original sentence, the model may guess wrong.)}\end{tabular} & \begin{tabular}[t]{p{1cm}}ViA1\end{tabular}& \begin{tabular}[t]{p{2.3cm}}C\\ E\\ C C C\end{tabular}                

\\ \hline
\begin{tabular}[t]{p{4.5cm}}Hơn 30 phút sau, đám cháy được khống chế, tàu hàng 2371 tiếp tục hành trình theo hướng Bắc Nam. \textit{(More than 30 minutes later, the fire was under control, cargo ship 2371 continued its journey in the north-south direction.)}\end{tabular}                 & \begin{tabular}[t]{p{2.5cm}}Đám cháy được khống chế sau hơn nửa giờ đồng hồ. \textit{(The fire was brought under control after more than half an hour.)}\end{tabular}  & \begin{tabular}[t]{p{4cm}}Mô hình khó khăn trong việc nhận biết nửa giờ đồng hồ với 30 phút là giống nhau, do đó có thể dự đoán sai thành Contradiction. \textit{(The difficulty model recognizes that half an hour and 30 minutes are the same, so it can be wrong to predict Contradiction.)}\end{tabular} & \begin{tabular}[t]{p{1cm}}ViA2\end{tabular}& \begin{tabular}[t]{p{2.3cm}}E\\ C\\ E E E\end{tabular} 

\\ \hline
\begin{tabular}[t]{p{4.5cm}}Việt Trinh hái mít, xoài, dứa... trĩu quả khi vào mùa ở vườn nhà rộng 3.000 m2. \textit{(Viet Trinh picks jackfruit, mango, pineapple... laden with fruit in season in a 3,000 m2 home garden.)}\end{tabular}                 & \begin{tabular}[t]{p{2.5cm}}Nhà Việt Trinh rộng 3.000 m2. \textit{(Viet Trinh’s house is 3,000 m2 wide.)}\end{tabular}  & \begin{tabular}[t]{p{4cm}}Tôi lược bỏ chữ "vườn" để tạo câu neutral và model có thể nhầm thành entailment. \textit{I omitted the word "garden" to create a neutral sentence and model could be mistaken for entailment.)}\end{tabular} & \begin{tabular}[t]{p{1cm}}ViA3\end{tabular}& \begin{tabular}[t]{p{2.3cm}}N\\ C\\ N N E\end{tabular} 

\\ \hline
\begin{tabular}[t]{p{4.5cm}}Chính phủ vừa đồng ý cho các thương nhân nước ngoài được nhập cảnh vào Việt Nam thu mua vải thiều Bắc Giang. \textit{(The Government has just agreed to allow foreign traders to enter Vietnam to buy Bac Giang lychees.)}\end{tabular}                 & \begin{tabular}[t]{p{2.5cm}}Vải thều Bắc Giang chỉ được trồng ở Bắc Giang. \textit{(Bac Giang lychee is only grown in Bac Giang.)}\end{tabular}  & \begin{tabular}[t]{p{4cm}}Tôi dùng kiến thức thực tế nhưng sai nên có thể máy sẽ đoán thành neutral. \textit{(I used practical knowledge but it was wrong, so maybe the machine will guess neutral.)}\end{tabular} & \begin{tabular}[t]{p{1cm}}ViA2\end{tabular}& \begin{tabular}[t]{p{2.3cm}}N\\ C\\ C N N\end{tabular} 

\\ \hline
\begin{tabular}[t]{p{4.5cm}}Đại diện EVN cho biết, tập đoàn không có hoạt động cho vay tín chấp và không có đơn vị trực thuộc hay doanh nghiệp liên kết có tên Ngân hàng Điện lực Việt Nam. \textit{(A representative of EVN said that the group has no unsecured lending activities and has no affiliated units or affiliated enterprises named Electricity of Vietnam Bank.)}\end{tabular}                 & \begin{tabular}[t]{p{2.5cm}}EVN không có bất kì hoạt động cho vay tín chấp nào trong thời gian gần đây. \textit{(EVN has not had any unsecured lending activities recently.)}\end{tabular}  & \begin{tabular}[t]{p{4cm}}Nêu thêm thời gian nên mô hình sẽ đoán sai. \textit{(Given more time, the model will guess wrong.)}\end{tabular} & \begin{tabular}[t]{p{1cm}}ViA3\end{tabular}& \begin{tabular}[t]{p{2.3cm}}E\\ C\\ E E C\end{tabular} 
\\ \hline
\end{tabular}}
\end{table*}

\newpage
\section{Visual Length of Premise and Hypothesis of ViANLI}
\label{VisualLengthOfPremiseAndHypothesis}

\begin{figure}[H]
    \centering
    \begin{subfigure}{0.3\textwidth}
        \includegraphics[width=\textwidth]{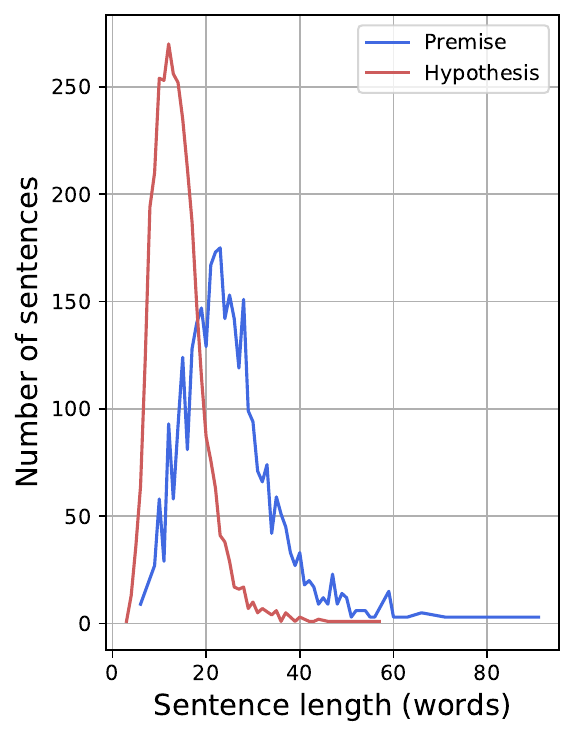}
        \caption{ViA1}
        \label{ViA1}
    \end{subfigure}
    \begin{subfigure}{0.3\textwidth}
        \includegraphics[width=\textwidth]{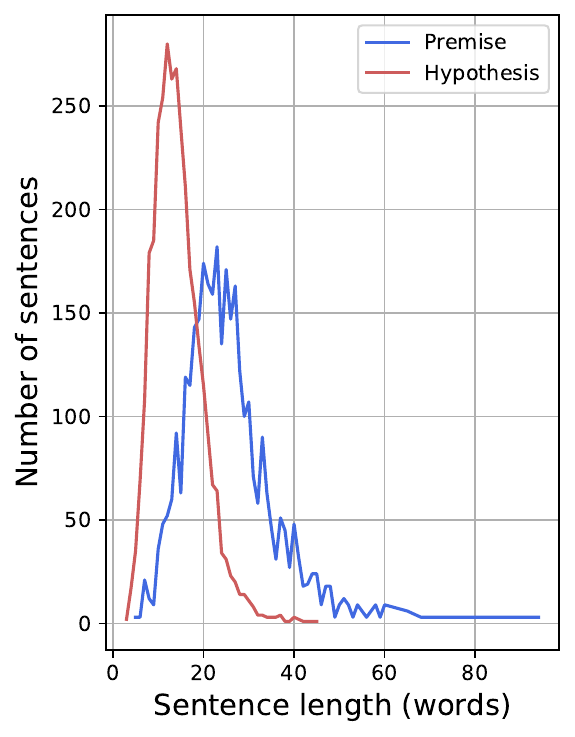}
        \caption{ViA2}
        \label{ViA2}
    \end{subfigure}
    \begin{subfigure}{0.3\textwidth}
        \includegraphics[width=\textwidth]{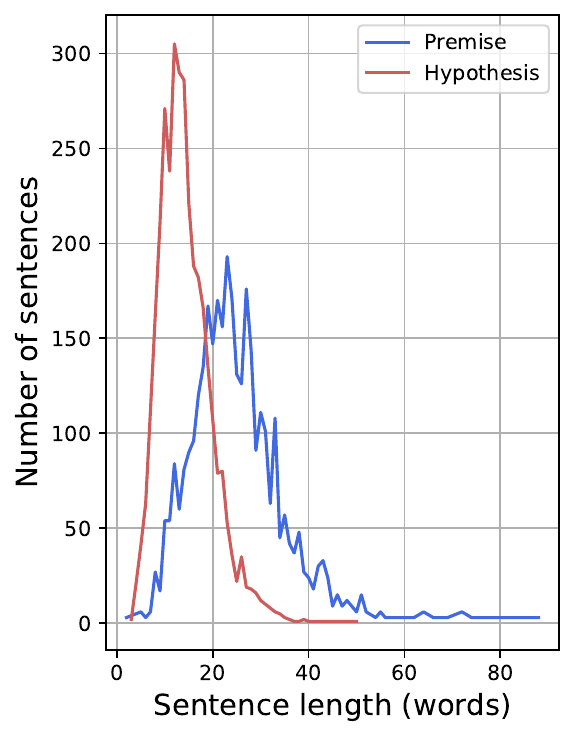}
        \caption{ViA3}
        \label{ViA3}
    \end{subfigure}
    \begin{subfigure}{0.3\textwidth}
        \includegraphics[width=\textwidth]{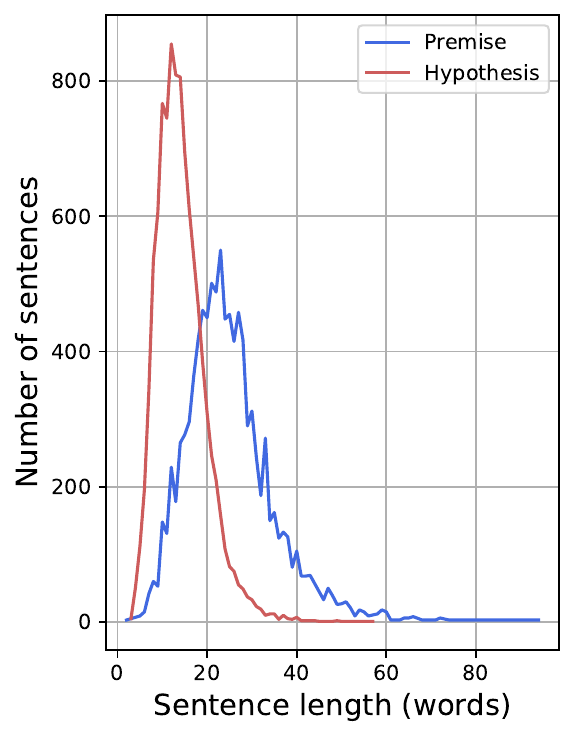}
        \caption{ViANLI}
        \label{ViANLI}
    \end{subfigure}
    \begin{subfigure}{0.305\textwidth}
        \includegraphics[width=\textwidth]{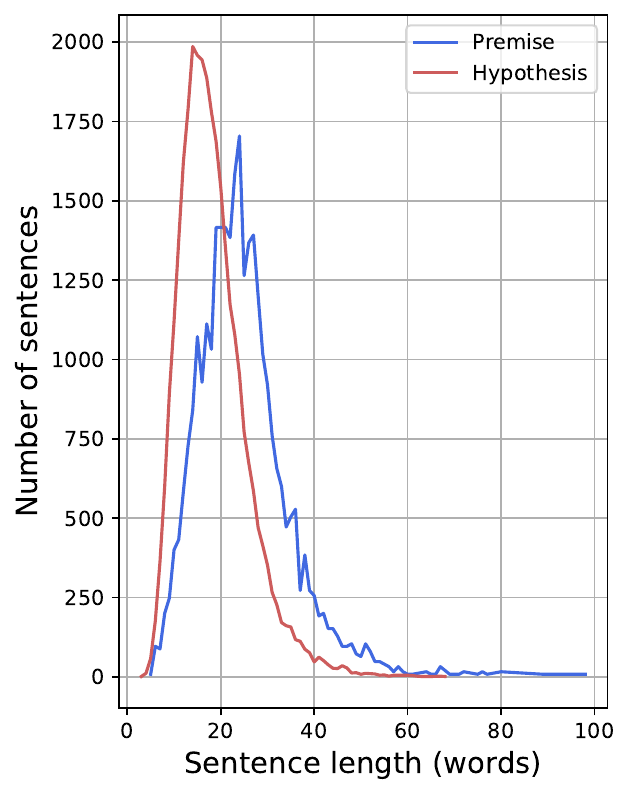}
        \caption{ViNLI}
        \label{ViNLI}
    \end{subfigure}
    \caption{Comparison of Length Distribution Between ViANLI and ViNLI.}
    \label{fig:datasets-length-distribution}
\end{figure}

\section{Model Accuracy on Other Test Sets When Training Data is Combined with Vietnamese Adversarial Data}
\label{ModelAccuracyonOtherTestSets}

\begin{table}[]
\renewcommand{\arraystretch}{1.5}
\centering
\caption{Model Accuracy (\%) on Other Test Sets When Training Data is Combined with Vietnamese Adversarial Data. ViA1, ViA2, ViA3 Are the Data of Round 1, Round 2 and Round 3, Respectively. ViANLI Refers to ViA1+ViA2+ViA3. VL Refers to the VLSP2021-NLI Dataset, and ViNe Refers to the VnNewsNLI Dataset.}
\label{tab:result1}
\resizebox{\columnwidth}{!}{%
\begin{tabular}{llrrrr}
\hline
\textbf{Model} & \textbf{Training data}                       & \textbf{ViANLI}       & \textbf{ViNLI}        & \textbf{VLSP2021-NLI}    & \textbf{VnNewsNLI}    \\ \hline
\multirow{5}{*}{mBERT}        & ViNLI, XNLI (Vi)                             & 27.50 & 66.21 & 70.16 & 72.87 \\ 
                              & + ViA1                                       & 28.80 & 66.82 & 70.06 & 62.11 \\ 
                              & + ViA1 + ViA2                                & 31.90 & 65.85 & 70.20 & 58.60 \\ 
                              & + ViA1 + ViA2 + ViA3                         & 33.70 & 66.47 & 71.43 & 58.16 \\ 
                              & ViNLI, XNLI (Vi), ViA1, ViA2, ViA3, VL, ViNe & 34.00 & 66.61 & 78.71 & 96.58 \\ \hline
\multirow{6}{*}{PhoBERT$_{Large}$} & ViNLI, XNLI (Vi)                             & 30.80 & 75.57 & 79.46 & 73.93 \\ 
                              & + ViNe                                       & 31.80 & 75.66 & 79.32 & 97.18 \\ 
                              & + ViNe + ViA1                                & 34.70 & 75.57 & 80.35 & 97.58 \\ 
                              & + ViNe + ViA1 + ViA2                         & 36.80 & 75.39 & 80.55 & 97.81 \\ 
                              & + ViNe + ViA1 + ViA2 + ViA3                  & 39.20 & 75.13 & 80.73 & 97.78 \\ 
                              & ViNLI, XNLI (Vi), ViA1, ViA2, ViA3, VL, ViNe & 41.00 & 75.22 & 87.72 & 97.51 \\ \hline
\multirow{5}{*}{XLM-R$_{Large}$}   & ViNLI, XNLI (Vi)                             & 30.80 & 82.12 & 82.11 & 80.93 \\  
                              & + VL + ViNe                                & 31.70 & 81.18 & 88.71 & 97.06 \\ 
                              & + VL + ViNe + ViA1                           & 34.00 & 82.59 & 89.14 & 96.68 \\ 
                              & + VL + ViNe + ViA1 + ViA2                    & 38.70 & 82.11 & 90.36 & 97.21 \\ 
                              & ViNLI, XNLI (Vi), ViA1, ViA2, ViA3, VL, ViNe & 40.10 & 81.49 & 89.84 & 97.31 \\ \hline
\multirow{5}{*}{NLIMoE$_{Dynamic}$}   & ViNLI, XNLI (Vi)                             &30.90  &82.33  &83.56  & 80.21 \\  
                              & + VL + ViNe                                &30.30  &82.37  &89.14  &97.19  \\ 
                              & + VL + ViNe + ViA1                           &34.90 &82.99  &89.66   & 97.59  \\ 
                              & + VL + ViNe + ViA1 + ViA2                    &38.30  &82.55  &89.75  &97.53  \\ 
                              & ViNLI, XNLI (Vi), ViA1, ViA2, ViA3, VL, ViNe & 40.60 & 82.59 & 90.58 &	97.72 \\ \hline
\end{tabular}%
}
\end{table}

\newpage
\section{Confusion Matrix of Pre-trained Language Models on the Development Set}
\label{ConfusionMatrix}

\begin{figure}[H]
    \centering
    \begin{subfigure}{0.45\textwidth}
        \includegraphics[width=\textwidth]{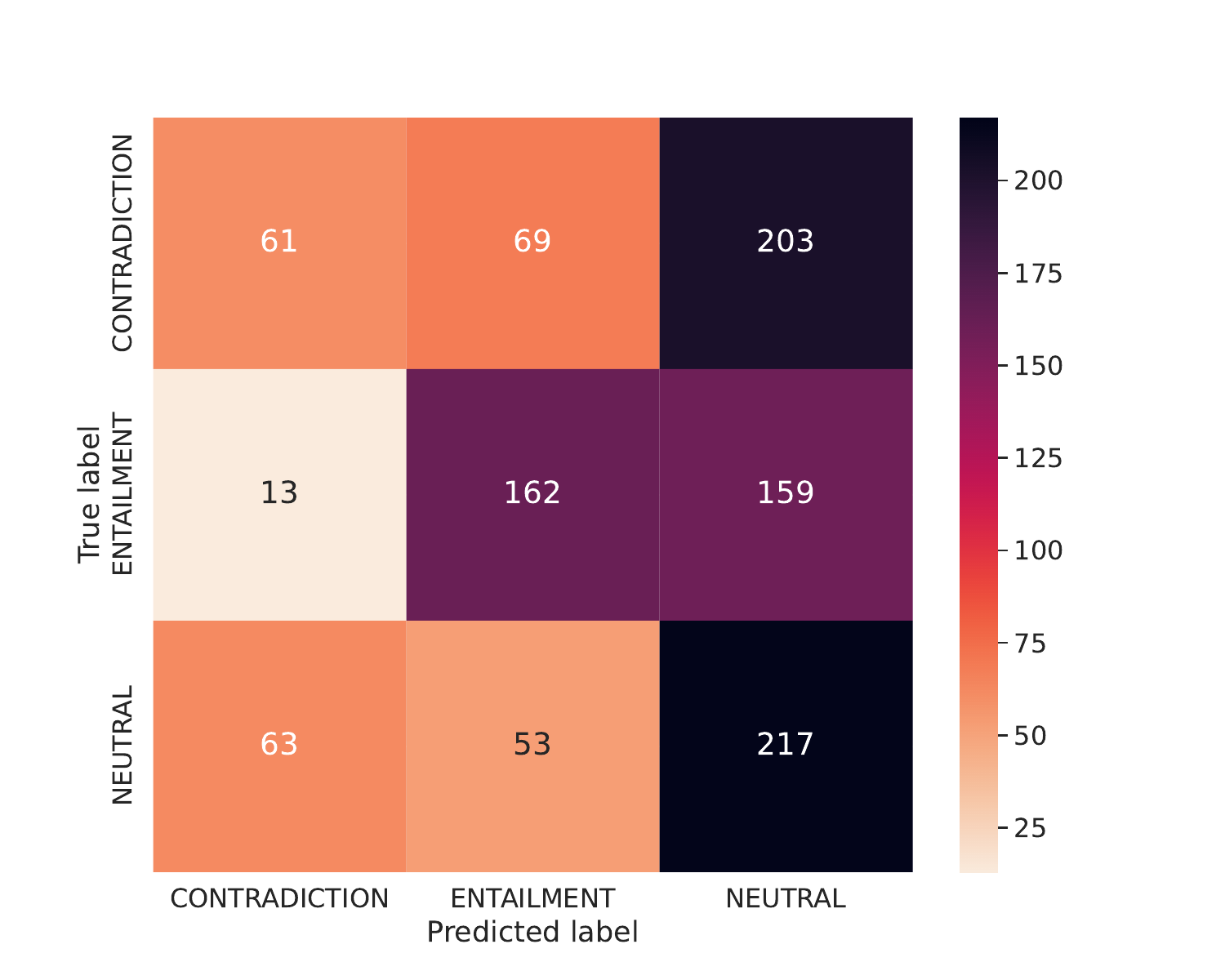}
        \caption{mBERT}
    \end{subfigure}
    \begin{subfigure}{0.45\textwidth}
        \includegraphics[width=\textwidth]{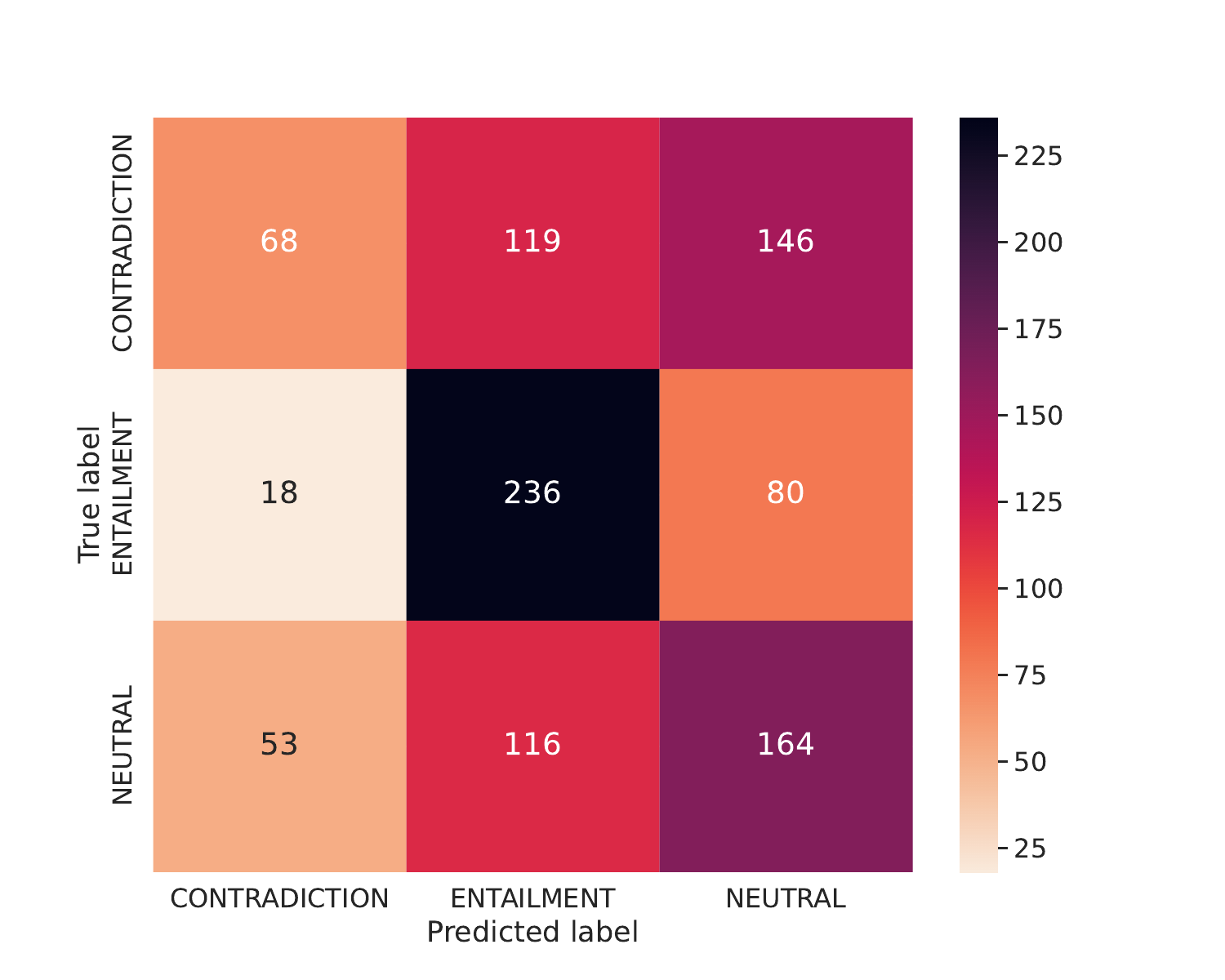}
        \caption{PhoBERT}
    \end{subfigure}
    \begin{subfigure}{0.45\textwidth}
        \includegraphics[width=\textwidth]{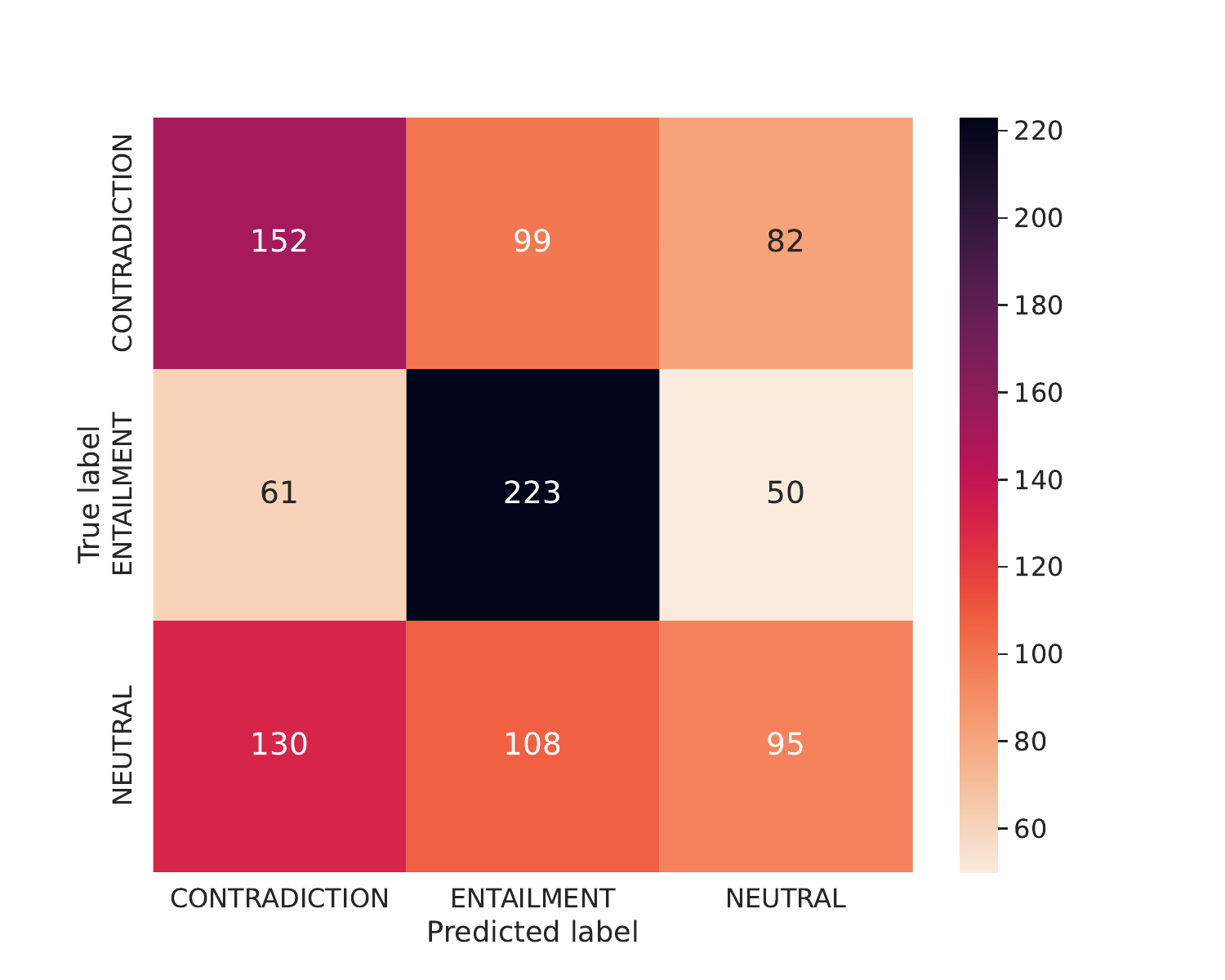}
        \caption{XLM-R}
    \end{subfigure}
    \begin{subfigure}{0.45\textwidth}
        \includegraphics[width=\textwidth]{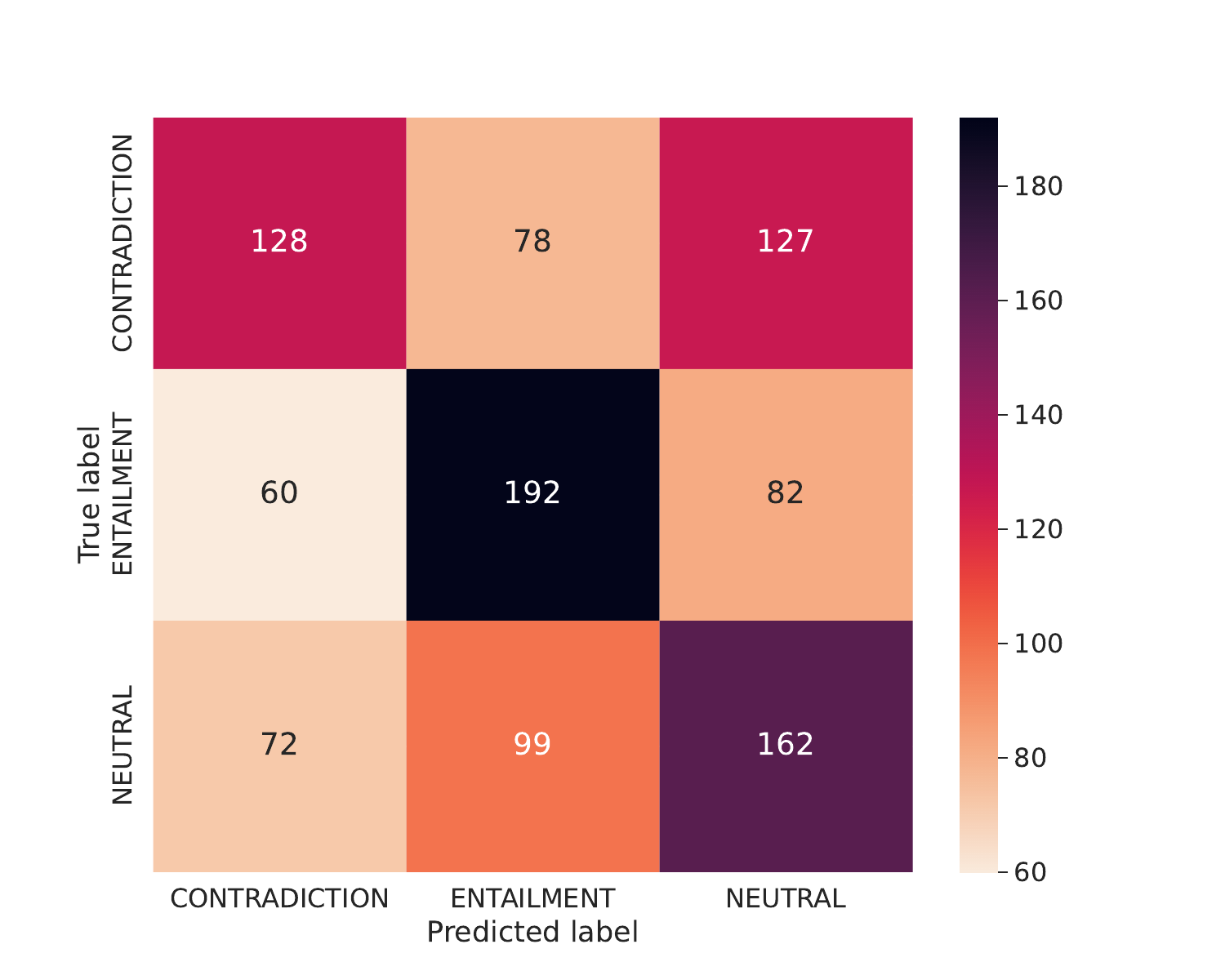}
        \caption{CafeBERT}
    \end{subfigure}
    \begin{subfigure}{0.45\textwidth}
        \includegraphics[width=\textwidth]{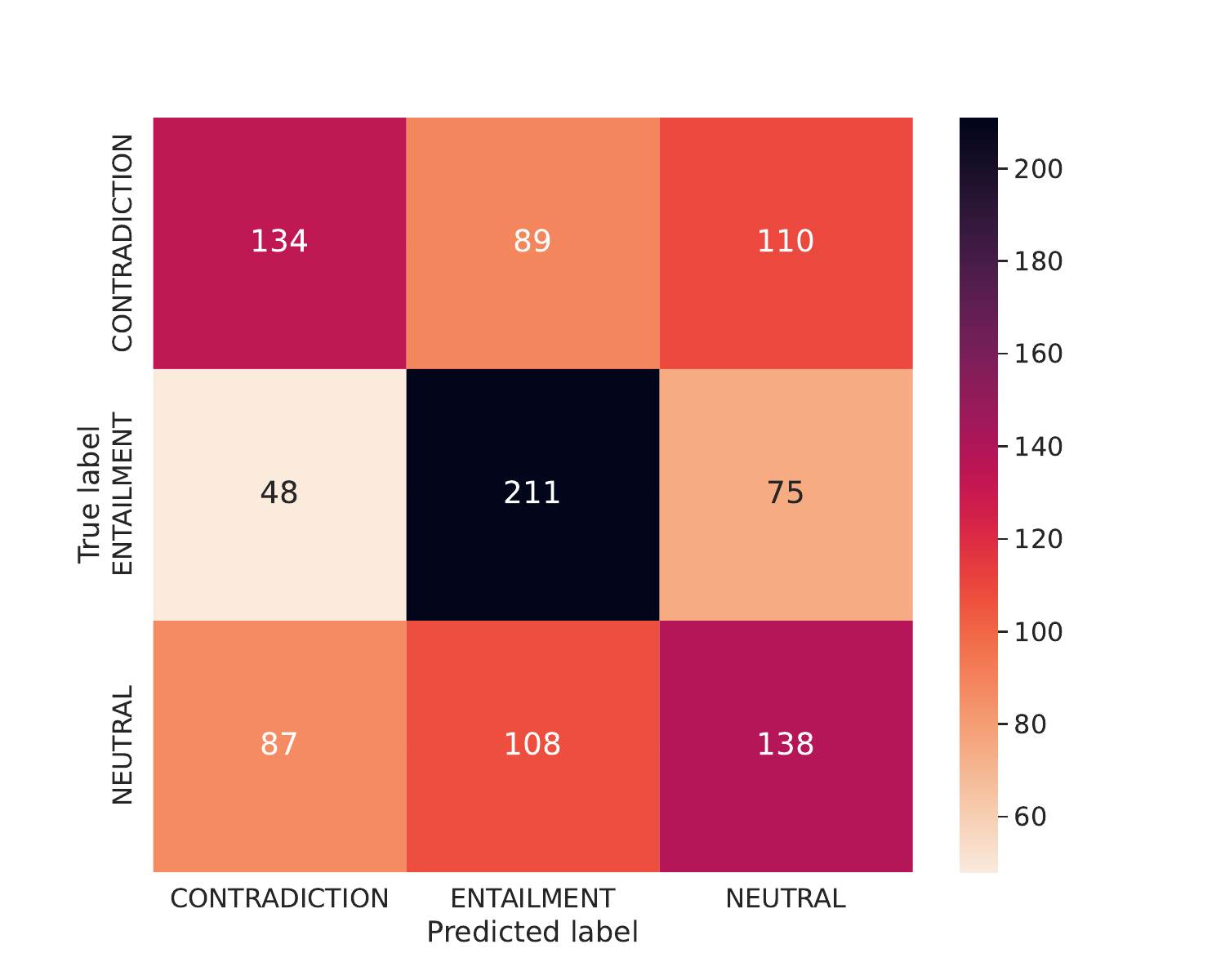}
        \caption{NLIMoE$_{TopK}$}
    \end{subfigure}
    \begin{subfigure}{0.45\textwidth}
        \includegraphics[width=\textwidth]{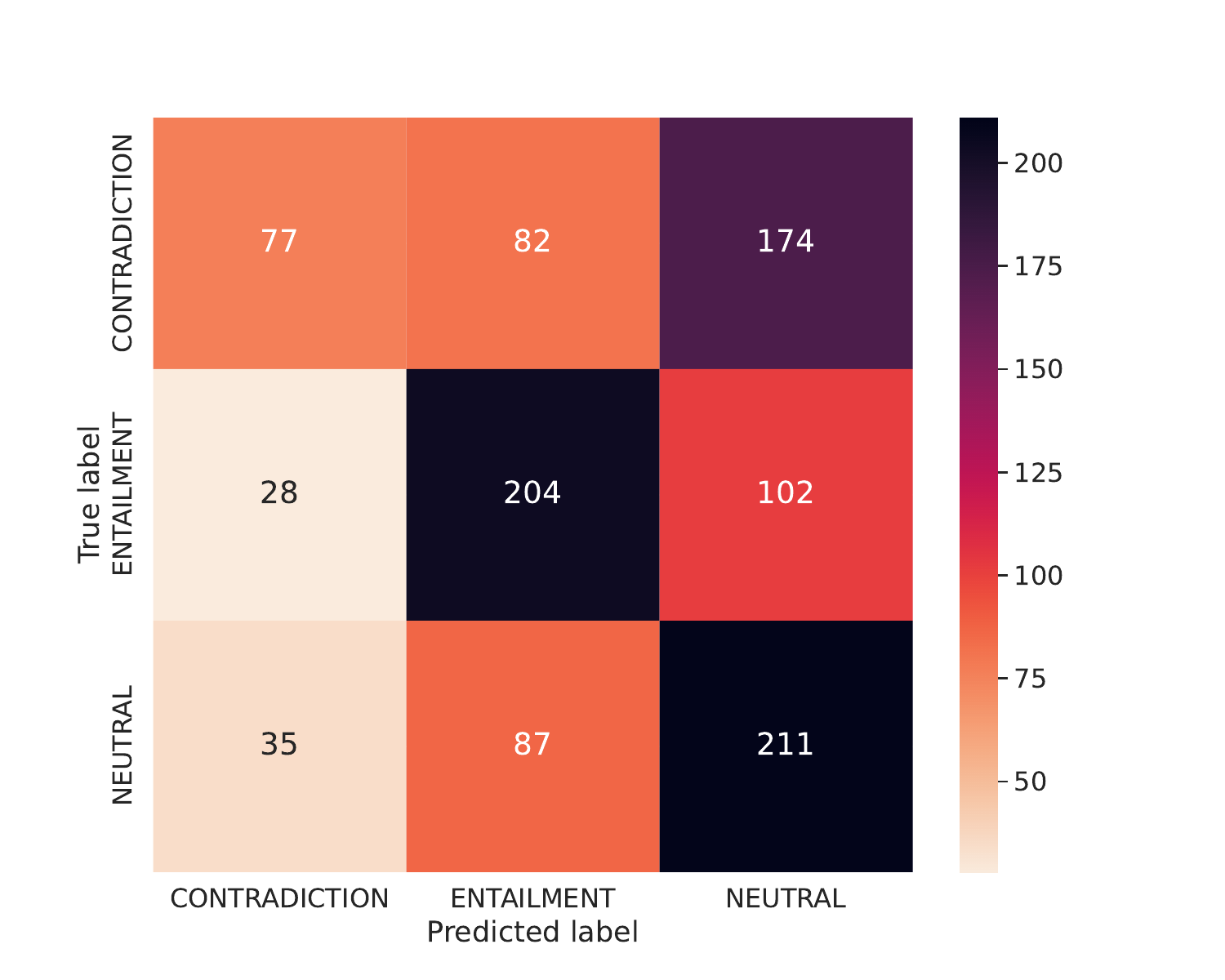}
        \caption{NLIMoE$_{Dynamic}$}
    \end{subfigure}

    \caption{Confusion Matrix of Pre-trained Language Models on the Development Set.}
    \label{fig:confusion_matrix}
\end{figure}

\newpage
\section{Comparison of Predicted Labels by NLIMoE$_{Dynamic}$ with Other Models}
\label{ComparisonofPredictedLabelsbyNLIMoE}

\begin{table}[H]
\centering
\caption{Comparison of Predicted Labels by NLIMoE$_{Dynamic}$ with Other Models (E = Entailment, C = Contradiction, N = Neutral).}
\label{tab:errorsample}
\resizebox{\columnwidth}{!}{%
\begin{tabular}{p{4cm}p{3.5cm}cccc}
\hline
\multicolumn{1}{c}{\multirow{2}{*}{\textbf{Premise}}} &
  \multicolumn{1}{c}{\multirow{2}{*}{\textbf{Hypothesis}}} &
  \multirow{2}{*}{\textbf{\begin{tabular}[c]{@{}c@{}}Gold \\ Label\end{tabular}}} &
  \multicolumn{2}{c}{\textbf{Predicted Label}} &
  \multirow{2}{*}{\textbf{\begin{tabular}[c]{@{}c@{}}Activated \\ Expert\end{tabular}}} \\ \cline{4-5}
\multicolumn{1}{c}{} &
  \multicolumn{1}{c}{} &
   &
  \multicolumn{1}{c|}{\textbf{mBERT / PhoBERT / XLM-R / CafeBERT}} &
  \textbf{NLIMoE} &
   \\ \hline
\begin{tabular}[c]{p{4cm}}Xoắn tinh hoàn xảy ra khi tinh hoàn di động quá mức quanh thừng tinh, dẫn đến tắc nghẽn hệ thống mạch máu cấp tính. \\ \textit{(Torsion of the testicle occurs when the testicle moves excessively around the spermatic cord, leading to acute vascular obstruction.)}\end{tabular} &
  \begin{tabular}[c]{p{3.5cm}}Xoắn tinh hoàn chỉ xảy ra ở nam giới, không xảy ra ở phụ nữ.\\ \textit{(Testicular torsion only occurs in males, not in females.)}\end{tabular} &
  E &
  \multicolumn{1}{c}{N / C / C / N} &
  N &
  {[}3{]} \\ \hline
\begin{tabular}[c]{p{4cm}}Bệnh nhân được các giáo sư đầu ngành Hội chẩn Quốc gia ngày 18/5, tuy nhiên tình trạng bệnh không giảm, tử vong ngày 20/5.\\ \textit{(The patient was consulted by leading professors at the National Consultation on May 18, however, the condition did not improve, and the patient passed away on May 20.)}\end{tabular} &
  \begin{tabular}[c]{p{3.5cm}}Bệnh nhân tử vong sau 5 ngày hội chẩn.\\ \textit{(The patient died 5 days after the consultation.)}\end{tabular} &
  C &
  \multicolumn{1}{c}{N / E / E / E} &
  E &
  {[}3, 5, 7{]} \\ \hline
\begin{tabular}[c]{p{4cm}}Tổng thống Palau gặp mặt lãnh đạo Đài Loan Thái Anh Văn, cùng tham gia sự kiện quảng bá du lịch Palau cũng như tham quan các công ty đóng tàu và nuôi trồng thủy sản.\\ \textit{(The President of Palau met with Taiwan’s leader, Tsai Ing-wen, to attend an event promoting Palau’s tourism and to visit shipbuilding and aquaculture companies.)}\end{tabular} &
  \begin{tabular}[c]{p{3.5cm}}Lãnh đạo 2 nước Palau và Đài Loan cùng nhau tham gia sự kiện quảng bá du lịch Palau, đồng thời tham quan các công ty đóng tàu và nuôi trồng thuỷ sản.\\ \textit{(The leaders of Palau and Taiwan participated in an event promoting Palau’s tourism while also visiting shipbuilding and aquaculture companies.)}\end{tabular} &
  E &
  \multicolumn{1}{c}{N / N / C / N} &
  \textbf{E} &
  {[}1, 3{]} \\ \hline
\begin{tabular}[c]{p{4cm}}Ban đầu LG kỳ vọng sẽ tìm được khách hàng mua lại nhà máy Hải Phòng với giá 100 tỷ won (2.064 tỷ đồng).\\ \textit{(Initially, LG expected to find a buyer for the Hai Phong factory at a price of 100 billion won (2.064 trillion VND).)}\end{tabular} &
  \begin{tabular}[c]{p{3.5cm}}LG định giá nhà máy tại Hải Phòng với giá dưới 2 nghìn tỷ đồng.\\ \textit{(LG priced the Hai Phong factory at under 2 thousand billion VND.)}\end{tabular} &
  C &
  \multicolumn{1}{c}{N / E / E / E} &
  \textbf{C} &
  {[}3, 5, 6, 7{]} \\ \hline
\end{tabular}}
\end{table}

\newpage
\section{Expert Activation Distribution across Inference Types}
\label{active_expert}

\begin{figure}[H]
    \centering
    \begin{subfigure}{1\textwidth}
        \includegraphics[width=\textwidth]{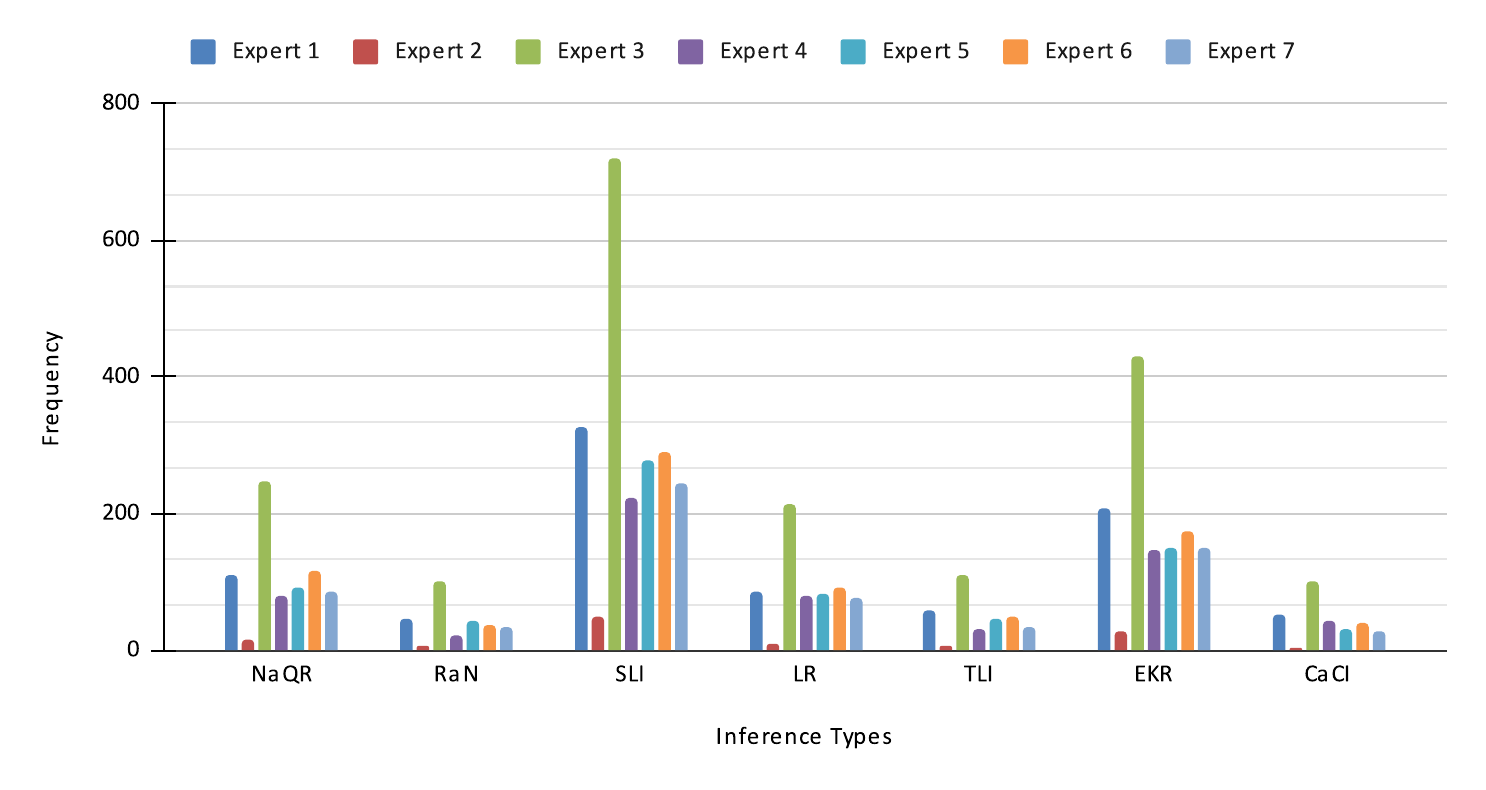}
    \end{subfigure}
    \caption{Distribution of activation frequencies of experts in the proposed NLIMoE$_{Dynamic}$ model across the seven reasoning types on the ViANLI development set.}
    \label{fig:active-expert}
\end{figure}

\end{document}